\pdfoutput=1
\documentclass{article}
\usepackage[final,dblblindworkshop]{neurips_2026}
\workshoptitle{BabyVLM: Toward Developmentally Plausible Multimodal Systems}
\usepackage[utf8]{inputenc}
\usepackage[T1]{fontenc}
\usepackage{hyperref}
\usepackage{url}
\usepackage{booktabs}
\usepackage{amsfonts,amsmath}
\usepackage[expansion=false]{microtype}
\usepackage{xcolor}
\usepackage{graphicx}
\usepackage{multirow}
\usepackage{array}
\usepackage{subcaption}
\usepackage[capitalise,noabbrev]{cleveref}
\usepackage{enumitem}
\setlist{nosep,leftmargin=1.2em}
\newcommand{\nModels}{9}
\newcommand{\nArms}{7}
\newcommand{\nDot}{272}
\newcommand{\nFsc}{200}
\newcommand{\nPixmo}{180}
\newcommand{\nImages}{652}
\newcommand{\totalCostUSD}{153}
\newcommand{\totalCostUSDexact}{153.13}

\newcommand{\dupGemProTagFsc}{1.8}

\newcommand{\omitGemProTagFsc}{28.7}

\newcommand{\multiGemProTagAll}{46.7}

\newcommand{\cardGemProTagDot}{100.0}

\newcommand{\cardGptMiniTagDot}{66.9}

\newcommand{\maeGemLiteTilesDot}{0.8}

\newcommand{\accGemLiteTilesFsc}{3.0}

\newcommand{\phantomGemLiteListFsc}{61.8}

\newcommand{\cardGemLiteTagDot}{93.0}

\newcommand{\multiGemLiteTagAll}{17.2}

\newcommand{\dupGemLiteTagFsc}{4.4}
\newcommand{\phantomGemLiteTagFsc}{24.8}
\newcommand{\omitGemLiteTagFsc}{38.1}

\newcommand{\accQwenThirtyTwoDirectDot}{60.7}
\newcommand{\maeQwenThirtyTwoDirectDot}{1.6}

\newcommand{\accQwenThirtyTwoTagDot}{68.0}
\newcommand{\maeQwenThirtyTwoTagDot}{17.5}

\newcommand{\nCommon}{240}

\newcommand{\acccmGemProDirectPix}{92.2}

\newcommand{\acccmQwenEightDirectFsc}{11.0}

\newcommand{\acccmQwenEightDirectPix}{62.2}

\newcommand{\acccmGemFlashDirectFsc}{52.0}

\newcommand{\acccmGemLiteDirectDot}{36.0}

\newcommand{\acccmGemLiteTagCropDot}{58.0}

\newcommand{\acccmGemLiteCotFsc}{41.0}

\newcommand{\acccmQwenThirtyTwoDirectPix}{65.6}

\newcommand{\acccmQwenThirtyTwoTagPix}{78.9}

\newcommand{\rEightyGemLiteFull}{20}
\newcommand{\rEightyGemLiteDirect}{7}

\newcommand{\accabGemLiteTagNoNudgeAll}{45.5}

\newcommand{\accabGemLiteTagGridAll}{45.5}

\newcommand{\accabGemLiteTagAll}{39.9}

\newcommand{\accRegGemLiteDirect}{51.4}
\newcommand{\accCluGemLiteDirect}{66.7}
\newcommand{\accRndGemLiteDirect}{58.3}

\newcommand{\accRegGemLiteTag}{90.3}
\newcommand{\accCluGemLiteTag}{70.8}
\newcommand{\accRndGemLiteTag}{79.2}

\newcommand{\nAbItems}{143}

\newcommand{\orderAgreeGemLiteDirect}{36.4}
\newcommand{\orderAgreeGemLiteTag}{39.2}

\providecommand{\accTOnecmpGemLite}{\XX}
\providecommand{\accTcmpGemLite}{\XX}
\providecommand{\cardProfTagQwenEight}{\XX}
\providecommand{\nAbDot}{\XX}
\providecommand{\nAbFsc}{\XX}
\providecommand{\nDotXL}{\XX}
\providecommand{\nImagesTotal}{\XX}
\providecommand{\nListGeTag}{\XX}
\providecommand{\nOrderImages}{\XX}
\providecommand{\nTagListCells}{\XX}
\providecommand{\nTagUpPix}{\XX}
\providecommand{\ndNudgeAccNeqGemLite}{\XX}
\providecommand{\ndNudgeAnsGtTagsGemLite}{\XX}
\providecommand{\ndNudgeMultiGemLite}{\XX}
\providecommand{\ndPlainAnsGtTagsGemLite}{\XX}
\providecommand{\ndPlainMultiGemLite}{\XX}
\providecommand{\ntDeltaCiGptMini}{\XX}
\providecommand{\ntDeltaGptMini}{\XX}
\providecommand{\ntReightyLowGemLite}{\XX}
\providecommand{\ntReightyLowGptMini}{\XX}
\providecommand{\ntReightyOffGemLite}{\XX}
\providecommand{\ntReightyOffGptMini}{\XX}
\providecommand{\omitTOnecmpGemFlash}{\XX}
\providecommand{\phantomProfTagGptMini}{\XX}
\providecommand{\phantomProfTagGrokFour}{\XX}

\providecommand{\dAccQwenEightTagFsc}{}\renewcommand{\dAccQwenEightTagFsc}{+6.0}

\providecommand{\dAccQwenThirtyTwoTagFsc}{}\renewcommand{\dAccQwenThirtyTwoTagFsc}{+7.0}

\providecommand{\dAccQwenThirtyTwoTagPix}{}\renewcommand{\dAccQwenThirtyTwoTagPix}{+13.3}
\providecommand{\dAccCiQwenThirtyTwoTagPix}{}\renewcommand{\dAccCiQwenThirtyTwoTagPix}{[+4,+22]}

\providecommand{\dAccGemLiteTagDot}{}\renewcommand{\dAccGemLiteTagDot}{+14.0}
\providecommand{\dAccCiGemLiteTagDot}{}\renewcommand{\dAccCiGemLiteTagDot}{[+4,+26]}

\providecommand{\dAccGemLiteCotFsc}{}\renewcommand{\dAccGemLiteCotFsc}{+18.0}
\providecommand{\dAccCiGemLiteCotFsc}{}\renewcommand{\dAccCiGemLiteCotFsc}{[+7,+29]}

\providecommand{\dAccGptMiniTagDot}{}\renewcommand{\dAccGptMiniTagDot}{-30.0}

\providecommand{\rxGrokFourDirect}{}\renewcommand{\rxGrokFourDirect}{3}

\providecommand{\rxGemProDirect}{}\renewcommand{\rxGemProDirect}{150}
\providecommand{\rxCiGemProDirect}{}\renewcommand{\rxCiGemProDirect}{[50,200]}

\providecommand{\rxQwenEightDirect}{}\renewcommand{\rxQwenEightDirect}{6}

\providecommand{\rxGemFlashDirect}{}\renewcommand{\rxGemFlashDirect}{150}
\providecommand{\rxCiGemFlashDirect}{}\renewcommand{\rxCiGemFlashDirect}{[100,200]}
\providecommand{\rxGemFlashCot}{}\renewcommand{\rxGemFlashCot}{80}

\providecommand{\rxGemLiteDirect}{}\renewcommand{\rxGemLiteDirect}{7}

\providecommand{\rxQwenThirtyTwoDirect}{}\renewcommand{\rxQwenThirtyTwoDirect}{7}

\providecommand{\nDotXL}{}\renewcommand{\nDotXL}{30}
\providecommand{\nImagesTotal}{}\renewcommand{\nImagesTotal}{\the\numexpr\nImages+\nDotXL\relax}
\providecommand{\nOrderImages}{}\renewcommand{\nOrderImages}{68}
\providecommand{\ordAgreeGemLite}{}\renewcommand{\ordAgreeGemLite}{43}

\providecommand{\ordAccLtrGemLite}{}\renewcommand{\ordAccLtrGemLite}{60}
\providecommand{\ordAccRtlGemLite}{}\renewcommand{\ordAccRtlGemLite}{47}
\providecommand{\ordAgreeGemLite}{}\renewcommand{\ordAgreeGemLite}{44}
\providecommand{\ordSameAgreeGemLite}{}\renewcommand{\ordSameAgreeGemLite}{65}
\providecommand{\ordEffectGemLite}{}\renewcommand{\ordEffectGemLite}{+21.7}
\providecommand{\ordEffectCiGemLite}{}\renewcommand{\ordEffectCiGemLite}{[+14,+31]}

\providecommand{\ordRotEffectGemLite}{}\renewcommand{\ordRotEffectGemLite}{+9.6}
\providecommand{\ordRotEffectCiGemLite}{}\renewcommand{\ordRotEffectCiGemLite}{[+1,+18]}
\providecommand{\ordRotAccGemLite}{}\renewcommand{\ordRotAccGemLite}{63}
\providecommand{\ordAgreeGemFlash}{}\renewcommand{\ordAgreeGemFlash}{88}

\providecommand{\ordAgreeGemFlash}{}\renewcommand{\ordAgreeGemFlash}{86}
\providecommand{\ordSameAgreeGemFlash}{}\renewcommand{\ordSameAgreeGemFlash}{87}
\providecommand{\ordEffectGemFlash}{}\renewcommand{\ordEffectGemFlash}{+0.7}

\providecommand{\ordRotEffectGemFlash}{}\renewcommand{\ordRotEffectGemFlash}{+0.0}

\providecommand{\ordAgreeQwenThirtyTwo}{}\renewcommand{\ordAgreeQwenThirtyTwo}{26}

\providecommand{\ordAgreeQwenThirtyTwo}{}\renewcommand{\ordAgreeQwenThirtyTwo}{24}
\providecommand{\ordSameAgreeQwenThirtyTwo}{}\renewcommand{\ordSameAgreeQwenThirtyTwo}{40}
\providecommand{\ordEffectQwenThirtyTwo}{}\renewcommand{\ordEffectQwenThirtyTwo}{+15.4}
\providecommand{\ordEffectCiQwenThirtyTwo}{}\renewcommand{\ordEffectCiQwenThirtyTwo}{[+8,+23]}

\providecommand{\accTcmpGemLite}{}\renewcommand{\accTcmpGemLite}{54.4}

\providecommand{\phTcmpGemLite}{}\renewcommand{\phTcmpGemLite}{21.6}
\providecommand{\accTOnecmpGemLite}{}\renewcommand{\accTOnecmpGemLite}{51.5}

\providecommand{\phTOnecmpGemLite}{}\renewcommand{\phTOnecmpGemLite}{6.1}

\providecommand{\omitTOnecmpGemFlash}{}\renewcommand{\omitTOnecmpGemFlash}{67.9}

\providecommand{\cardProfTagQwenEight}{}\renewcommand{\cardProfTagQwenEight}{95.1}

\providecommand{\perrTagGemLite}{}\renewcommand{\perrTagGemLite}{5.1}

\providecommand{\perrTagGptMini}{}\renewcommand{\perrTagGptMini}{40.4}

\providecommand{\phantomProfTagGptMini}{}\renewcommand{\phantomProfTagGptMini}{31.9}

\providecommand{\perrTagGemFlash}{}\renewcommand{\perrTagGemFlash}{1.6}

\providecommand{\perrTagGrokFour}{}\renewcommand{\perrTagGrokFour}{34.1}

\providecommand{\phantomProfTagGrokFour}{}\renewcommand{\phantomProfTagGrokFour}{52.7}

\providecommand{\tagEqDirectFscMin}{}\renewcommand{\tagEqDirectFscMin}{10}
\providecommand{\tagEqDirectFscMax}{}\renewcommand{\tagEqDirectFscMax}{53}
\providecommand{\rhoFscMin}{}\renewcommand{\rhoFscMin}{-0.19}
\providecommand{\rhoFscMax}{}\renewcommand{\rhoFscMax}{+0.62}
\providecommand{\closerPooled}{}\renewcommand{\closerPooled}{83}

\providecommand{\pertShiftGemLiteDotMinus}{}\renewcommand{\pertShiftGemLiteDotMinus}{-1.75}

\providecommand{\pertFollowImgAllGemLiteMinus}{}\renewcommand{\pertFollowImgAllGemLiteMinus}{87}
\providecommand{\pertRepairAllGemLiteMinus}{}\renewcommand{\pertRepairAllGemLiteMinus}{57}

\providecommand{\pertRepairAllGptMiniMinus}{}\renewcommand{\pertRepairAllGptMiniMinus}{33}

\providecommand{\pertFollowImgAllGemFlashMinus}{}\renewcommand{\pertFollowImgAllGemFlashMinus}{84}
\providecommand{\pertRepairAllGemFlashMinus}{}\renewcommand{\pertRepairAllGemFlashMinus}{84}

\providecommand{\ntReightyOffGemLite}{}\renewcommand{\ntReightyOffGemLite}{4}
\providecommand{\ntReightyLowGemLite}{}\renewcommand{\ntReightyLowGemLite}{7}

\providecommand{\ntDeltaGptMini}{}\renewcommand{\ntDeltaGptMini}{-9.4}
\providecommand{\ntDeltaCiGptMini}{}\renewcommand{\ntDeltaCiGptMini}{[-13,-5]}

\providecommand{\ntReightyOffGptMini}{}\renewcommand{\ntReightyOffGptMini}{2}
\providecommand{\ntReightyLowGptMini}{}\renewcommand{\ntReightyLowGptMini}{7}

\providecommand{\rtokDirectGemFlash}{}\renewcommand{\rtokDirectGemFlash}{480}

\providecommand{\rtokDirectGemPro}{}\renewcommand{\rtokDirectGemPro}{482}

\providecommand{\rtokDirectGrokFour}{}\renewcommand{\rtokDirectGrokFour}{1894}

\providecommand{\rtokDirectGemFlashOneFifty}{}\renewcommand{\rtokDirectGemFlashOneFifty}{3146}

\providecommand{\latRndQwenEightTag}{}\renewcommand{\latRndQwenEightTag}{25}

\providecommand{\latHexQwenEightTag}{}\renewcommand{\latHexQwenEightTag}{25}
\providecommand{\latHolesQwenEightTag}{}\renewcommand{\latHolesQwenEightTag}{25}

\providecommand{\latRndGemLiteTag}{}\renewcommand{\latRndGemLiteTag}{55}
\providecommand{\latRotGemLiteTag}{}\renewcommand{\latRotGemLiteTag}{30}
\providecommand{\latHexGemLiteTag}{}\renewcommand{\latHexGemLiteTag}{85}
\providecommand{\latHolesGemLiteTag}{}\renewcommand{\latHolesGemLiteTag}{90}

\providecommand{\jitCorrGemLite}{}\renewcommand{\jitCorrGemLite}{0.86}
\providecommand{\jitDgtGemLite}{}\renewcommand{\jitDgtGemLite}{3.3}
\providecommand{\jitDidealGemLite}{}\renewcommand{\jitDidealGemLite}{14.1}

\providecommand{\jitCorrGemFlash}{}\renewcommand{\jitCorrGemFlash}{0.99}

\providecommand{\dTagListQwenEightDot}{}\renewcommand{\dTagListQwenEightDot}{+12.0}

\providecommand{\dTagListGemLiteFsc}{}\renewcommand{\dTagListGemLiteFsc}{-9.0}
\providecommand{\dTagListCiGemLiteFsc}{}\renewcommand{\dTagListCiGemLiteFsc}{[-17,-1]}

\providecommand{\dTagListGptMiniDot}{}\renewcommand{\dTagListGptMiniDot}{-22.0}

\providecommand{\nListGeTag}{}\renewcommand{\nListGeTag}{11}
\providecommand{\nTagListCells}{}\renewcommand{\nTagListCells}{15}
\providecommand{\ndNudgeAnsGtTagsGemLite}{}\renewcommand{\ndNudgeAnsGtTagsGemLite}{13}

\providecommand{\ndNudgeMultiGemLite}{}\renewcommand{\ndNudgeMultiGemLite}{27}

\providecommand{\ndNudgeAccNeqGemLite}{}\renewcommand{\ndNudgeAccNeqGemLite}{17}
\providecommand{\ndPlainAnsGtTagsGemLite}{}\renewcommand{\ndPlainAnsGtTagsGemLite}{6}

\providecommand{\ndPlainMultiGemLite}{}\renewcommand{\ndPlainMultiGemLite}{38}

\providecommand{\nAbDot}{}\renewcommand{\nAbDot}{68}
\providecommand{\nAbFsc}{}\renewcommand{\nAbFsc}{75}
\providecommand{\tilesPixMeanGtGemLite}{}\renewcommand{\tilesPixMeanGtGemLite}{6.0}
\providecommand{\tilesPixMeanAnsGemLite}{}\renewcommand{\tilesPixMeanAnsGemLite}{9.9}

\providecommand{\cotRowN}{}\renewcommand{\cotRowN}{6}

\providecommand{\cotRegN}{}\renewcommand{\cotRegN}{24}
\providecommand{\cotRegExact}{}\renewcommand{\cotRegExact}{8}
\providecommand{\cotXlN}{}\renewcommand{\cotXlN}{30}
\providecommand{\cotXlMeanErr}{}\renewcommand{\cotXlMeanErr}{-7.2}

\providecommand{\nTagUpPix}{}\renewcommand{\nTagUpPix}{4}

\providecommand{\tagOneCostRatioGemLite}{}\renewcommand{\tagOneCostRatioGemLite}{12}
\providecommand{\tagOneCostRatioGemFlash}{}\renewcommand{\tagOneCostRatioGemFlash}{15}
\providecommand{\nOrderImages}{}\renewcommand{\nOrderImages}{68}

\newcommand{\XX}{\textcolor{red}{XX}}
\newcommand{\tool}[1]{\texttt{#1}}
\newcommand{\arm}[1]{\textsc{#1}}
\definecolor{tagred}{RGB}{220,20,40}

\title{Give the Model a Finger: Developmentally Aligned\\ Evaluation of Counting in Vision--Language Models}

\author{%
  Yingxin Jia\\
  Independent Researcher\\
  \texttt{jia\_yingxin@hotmail.com}
  \And
  Hui Geng\\
  Independent Researcher\\
  \texttt{huigeng97@gmail.com}
  \And
  Wenxin Yang\\
  Independent Researcher\\
  \texttt{wh.wenxin.yang@gmail.com}
  \And
  Mohan Zhang\\
  Independent Researcher\\
  \texttt{mohanz@antigma.ai}
}

\begin{document}

\maketitle

\begin{abstract}
Vision--language models (VLMs) are exact on small sets and drift into estimation on larger ones, a profile reminiscent of children before they master counting. Children cross the barrier with epistemic actions: pointing at each item while reciting number words, moving counted items aside, partitioning the array. We give VLMs the same actions as \emph{dumb} image tools with no visual intelligence of their own (a numbered tag at a coordinate, a zoom, an eraser, a grid) in a training-free agent loop. Our central claim is about measurement, not intervention: a model that counts by marking leaves a normally hidden serial procedure on the image, and the developmental principles of counting (used as measurement theory, not as a claim that VLMs are children) turn those marks into an interpretable audit of serial visual reasoning: every tag is classified against point-level ground truth as on-target, duplicate, phantom, or omitted (one-to-one correspondence), the answer is checked against the number of tags (cardinality), and order irrelevance is tested by forcing opposite traversal orders. Across \nModels{} models, \nImagesTotal{} images (a controlled dot-array battery; FSC-147, 8--80 objects; PixMo-Count, 2--10), and \nArms{} harness conditions we find that (i) the exact-counting range is a model-specific signature spanning nearly two orders of magnitude ($R_{80}$ from \rxGrokFourDirect{} to \rxGemFlashDirect{} dots); (ii) as an intervention the finger mostly fails: tags help \nTagUpPix{} of 9 models on small-count photographs (CI-solid for one), hurt imprecise pointers, and rarely beat a verbal routine or a blind point list; what the loop buys is an auditable trace, not accuracy; (iii) as measurement it succeeds: a tag-perturbation probe shows that precise pointers read their answers off the tags, pointing precision tracks which models benefit, dense scenes fail by omission (consistent with children's keeping-track errors), and models rarely revise after seeing their own tags; and (iv) the capacities dissociate: models follow an instructed traversal order yet often change their answer when it reverses, beyond repeat and rotation controls, and children's regular-over-scattered advantage appears only under the routine and survives lattices that defeat a rows-times-columns shortcut.
\end{abstract}

\section{Introduction}
Ask a three-year-old how many crackers are on the table and, if there are two or three, the answer is immediate and exact. Ask about twelve and the child either guesses or starts to \emph{do} something: touch each cracker in turn, say ``one, two, three\dots'', slide the counted ones aside, and announce the last word as the total \citep{gelman1978child,fuson1988children,wynn1990children}. Two systems are at work: a fast, capacity-limited individuation of up to about four items (\emph{subitizing}, \citealp{kaufman1949discrimination,trick1994small}) and an approximate sense of magnitude \citep{feigenson2004core,xu2000large}. Exact enumeration beyond that range is a learned \emph{procedure} that recruits the hands and the space around the child: pointing keeps items and number words in one-to-one correspondence \citep{alibali1999function,saxe1977developmental,graham1999role}, moving items aside and scanning row by row prevent double counts and skips \citep{fuson1988children,potter1968spatial,beckwith1966process}, and partitioning turns one big count into several small ones \citep{vanoeffelen1982configurational}: \emph{epistemic actions}, performed not to change the world but to make a computation easier \citep{kirsh1994distinguishing,kirsh1995intelligent}.

\begin{figure}[t]
\centering
\includegraphics[width=\linewidth]{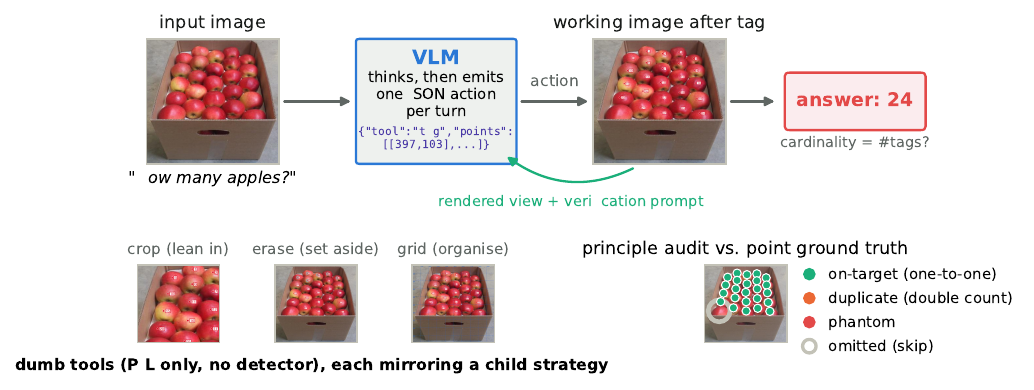}
\caption{The CountAgent loop, drawn with real artifacts from one trajectory (Gemini 3.1 Flash-Lite, FSC-147, 25 apples). Each turn the VLM emits one JSON action; the harness executes it with dumb PIL primitives, re-renders the working image, and returns it with a verification prompt, for at most $T$ tool calls. The final tags are audited against point ground truth, decomposing a wrong count into principle violations; here the model draws 24 tags, all on-target, misses one apple, and answers 24 against a ground truth of 25.}
\label{fig:method}
\end{figure}

Vision--language models (VLMs) look strikingly like the pre-procedural child: near-perfect on one to four objects, sharply worse after that, systematically under-counting dense sets, erring like noisy estimators rather than enumerators \citep{testolin2025visual,tamarapalli2025countqa,che2026counting,budny2025visual}, with the count present in the visual features but mis-read at the output \citep{pang2026unveiling,elshangiti2026count}. VLMs have a number sense but no counting routine. The claim is not that a VLM \emph{is} a child or models one, but that the developmental account of counting supplies testable hypotheses (which external scaffolds should help) and diagnostic variables (what to measure when they do not).

We therefore ask: \emph{if we give a VLM the epistemic actions that children use, does it count like a child who has learned to count, and does it fail like one?} We build a training-free harness in which the model can draw a numbered tag at any image coordinate and \emph{see} the tagged image, zoom into a region, grey out counted items, and overlay a coordinate grid. Every tool is a plain pixel operation with no detector, segmenter, or OCR behind it, so all perception and all decisions remain the model's own; the harness only lends it a finger, a magnifier, and an eraser.

The deeper contribution is measurement. A model that counts by marking externalises, mark by mark, a serial procedure that is normally hidden in its activations; the developmental principles of counting give that trace a grammar, turning model-generated marks into an interpretable audit of serial visual reasoning. This is what developmentally aligned evaluation adds over a benchmark score: not \emph{whether} a model counts, but \emph{which} constituent capacities it has, pointing, one-to-one tracking, cardinal read-out, checking, order invariance, each measured on its own. Following the classic analyses of children's counting \citep{gelman1978child,gelman1983preschoolers,fuson1988children}, we classify every tag as on-target, duplicate, or phantom and every object as counted or omitted against point-level ground truth, test \emph{cardinality} against the tags and \emph{order irrelevance} by forcing opposite traversal orders, and measure \emph{pointing precision}, on a controlled dot-array battery and two photographic benchmarks, FSC-147 \citep{ranjan2021learning} (8--80 dot-annotated instances) and PixMo-Count \citep{deitke2024molmo} (2--10).

\paragraph{Contributions.} (1) A developmentally aligned evaluation in which counting principles turn model-generated marks into an interpretable audit of serial visual reasoning: each principle becomes a measurable prediction (\cref{tab:predictions}), and a wrong count decomposes into principle violations against point-level ground truth (\cref{sec:protocol}). (2) A training-free harness of \emph{dumb epistemic tools} for counting, each mapped to a child strategy and a counting principle (\cref{sec:method}). (3) An empirical study over \nModels{} models and \nImagesTotal{} images with paired bootstrap uncertainty on every headline contrast (\cref{sec:results}), showing that the constituent capacities dissociate across models rather than forming one ``counting ability'' (\cref{tab:profile}), and that the tools are diagnostic rather than therapeutic: a finger helps only models that can already point, and where it fails, the audit localises the missing prerequisite; the study costs under \totalCostUSD{} US dollars, and all code, stimuli, and trajectories will be released.

\section{Background}
\label{sec:background}
\paragraph{How children count.} \citet{gelman1978child} described counting as governed by five principles: \emph{one-to-one} (each item receives exactly one tag), \emph{stable order}, \emph{cardinality} (the last tag names the set size), \emph{abstraction} (anything can be counted), and \emph{order-irrelevance}. Children acquire number-word meanings slowly \citep{wynn1990children,wynn1992children,lecorre2007one,sarnecka2008how,carey2009origin}, and even children who know the routine make \emph{keeping-track} errors on large or disorganised arrays, skipping or double counting \citep{fuson1988children,gelman1983preschoolers}. Accuracy improves when children point at or touch each item \citep{alibali1999function,graham1999role,saxe1977developmental}, when the array is regular rather than scattered \citep{potter1968spatial,beckwith1966process}, and when large sets are counted in groups \citep{vanoeffelen1982configurational}. Small sets are subitized \citep{kaufman1949discrimination,mandler1982subitizing}, large sets estimated by an approximate number system shared with infants and animals \citep{feigenson2004core,dehaene2011number}; number-sensitive units emerge spontaneously in deep networks \citep{stoianov2012emergence,nasr2019number,testolin2020numerosity}.

\paragraph{How VLMs count.} Counting benchmarks range from small sets \citep{paiss2023teaching,deitke2024molmo,acharya2019tallyqa} to dense scenes \citep{ranjan2021learning} and in-the-wild or adversarial questions \citep{tamarapalli2025countqa,deng2026holocount}. VLMs are accurate up to four or five objects and then degrade \citep{testolin2025visual,che2026counting}; failure grows with overlap, occlusion, compositional structure, and prior-knowledge bias \citep{rahmanzadehgervi2024vision,pothiraj2025capture,guo2025your,vo2025vision} and tracks the serial demand that also slows humans \citep{budny2025visual}. Remedies include count-aware training \citep{paiss2023teaching}, pointing supervision \citep{deitke2024molmo,jiang2026rexomni}, and point-then-count prompting \citep{alghisi2026getting,hou2025assessing}.

\paragraph{Tools and drawing for visual reasoning.} Language models can call vision tools \citep{gupta2023visual,suris2023vipergpt,hu2024sketchpad,zhao2025pyvision,su2025thinking}, receive marks from a segmenter \citep{yang2023som} or a coordinate scaffold \citep{lei2024scaffolding}, or zoom into regions \citep{wu2024vstar,shen2024zoomeye,zheng2025deepeyes}. Closest to us, Autonomous Imagination \citep{liu2024autonomous}, SketchVLM \citep{collins2026sketchvlm}, point-then-count prompting \citep{alghisi2026getting}, LVLM-Count \citep{fetrat2024lvlmcount}, and ABACUS \citep{mondal2026abacus} let a model mark, cursor over, or tile the image, and treat the marks as an intervention or an explanation judged by task accuracy (positioning table in \cref{tab:relwork}). We treat them as \emph{measurement}: dumb pixel tools keep every mark the model's own, and point-level ground truth decomposes the trace into omissions, duplicates, phantoms, and broken cardinality. The matching is standard (point-based crowd-localization metrics, \citealp{idrees2018composition,gao2020nwpu,song2021rethinking}; TIDE's error taxonomy, \citealp{bolya2020tide}); new are the loop that makes the marks and the principle tests that read them.

\section{CountAgent: dumb epistemic tools for counting}
\label{sec:method}
\paragraph{Design principle: dumb tools, smart eyes.} Every tool is a pure PIL operation: the harness never detects, segments, recognises, or reads anything; it draws and crops what it is told, and the only intelligence outside the model is geometric bookkeeping (numbering and capping tags), so the study tests \emph{epistemic actions}, not tool quality.

\paragraph{Tools and the child strategies they mirror.} \Cref{tab:tools} (appendix) maps each tool to a child strategy and a principle. \tool{tag} draws a dot at the requested coordinate and a numbered box beside it, the finger plus the number word (numbering continues across calls, so the harness supplies the \emph{stable order}, which is scaffolded, not tested); \tool{crop} returns an enlarged view with tags visible (leaning in); \tool{erase} greys out counted objects (moving them aside); \tool{grid} overlays a coordinate grid with 0--1000 ticks \citep{lei2024scaffolding}. A harness-driven \tool{tiles} routine ($2{\times}2$ overlapping partition, one point-listing call per tile, geometric de-duplication in the overlap) mirrors counting by groups without any agency on the model's side.

\paragraph{Loop.} \Cref{fig:method} shows the loop: a system prompt fixes the tools, the coordinate frame, and a budget of $T{=}6$ tool calls; each reply must end with one JSON action; after a \tool{tag} call the verification prompt asks whether each tag sits on a distinct object, any object is untagged, or two tags share one, then to tag only the missing ones or answer. The loop ends when the model answers; on budget exhaustion it is asked once more for a number, then the tag count is used. Prompts are verbatim in \cref{app:prompts}; rendering, context policy, and the (make-or-break) coordinate-frame choice are in \cref{app:tools}.

\paragraph{Harness conditions (arms).} \arm{direct}: answer with a single number (\emph{answer-only}, not reasoning-free: reasoning models still spend hidden reasoning tokens at the low effort we use, which \cref{app:reason} quantifies together with a thinking-off control). \arm{cot}: think step by step, then answer. \arm{list}: emit one coordinate per object and a count in one turn, never seeing the points rendered (the ``blind pointing'' protocol of \citealp{alghisi2026getting}). \arm{tag}: \tool{tag} with the verification loop. \arm{tag+crop}: adds \tool{crop}. \arm{full}: \tool{tag}, \tool{crop}, \tool{erase}, \tool{grid}. \arm{tiles}: the scripted partition. \arm{tag-1}: \tool{tag} restricted to one point per call, so every mark is placed only after the previous one is seen, enforcing strict look--mark--look seriality (run on the ablation subset with a budget of $T{=}80$ calls, so no object goes untagged by construction). Ablations remove the verification prompt, replace numbers by plain marks, add single tools, and vary $T$.

\section{A developmentally aligned evaluation protocol}
\label{sec:protocol}
\Cref{tab:predictions} (appendix) lists the predictions the developmental account makes for any counter and how we test each.
\paragraph{Stimuli.} \emph{DotArrays} (\nDot{} images) is a controlled battery in the tradition of child enumeration studies: black discs on white with $n \in \{1,\dots,10,12,15,20,25,30,40,50,60\}$, three arrangements (near-regular lattice, uniformly random, clustered), four seeds each, plus four factor sets: heterogeneous shapes and colours (abstraction), red discs among as many blue ones (distractors), touching discs (dense), and 6-px discs (small); an \emph{XL extension} (\nDotXL{} images, $n \in \{80,\dots,200\}$) locates the cliff of the two models that stay above 80\,\% in the main battery, and 60 lattice-control images (rotated, hexagonal, holed) test the arrangement effect (\cref{app:data}). \emph{FSC-147} (\nFsc{} test images, 29 classes) supplies real photographs with 8--80 instances and a dot per instance \citep{ranjan2021learning}, stratified into five count bins. \emph{PixMo-Count} (\nPixmo{} test images) supplies human-verified photographs with 2--10 instances \citep{deitke2024molmo}, twenty per count.

\paragraph{Models.} We evaluate \nModels{} models through one API (OpenRouter) with temperature 0 and low reasoning effort where the model supports it. Three inexpensive tier-1 models run every arm on every image (Gemini 3.1 Flash-Lite, Qwen3-VL-32B, Qwen3-VL-8B); two tier-2 models (GPT-5 mini, Gemini 3.7 Flash) skip \arm{tag+crop} and \arm{tiles}; four tier-3 models (Gemini 3.1 Pro, Claude Sonnet 5, Grok 4.20, Grok 4.5) run \arm{direct}, \arm{cot}, \arm{tag}, and \arm{full} on a stratified subset of \nCommon{} images. Cross-model comparisons use that \emph{common subset} (50 DotArrays, 100 FSC-147, 90 PixMo-Count images, stratified by count); the \emph{ablation subset} (\nAbItems{} images: \nAbDot{} DotArrays arrays with $n\ge5$ and \nAbFsc{} FSC-147 photographs) hosts the ablations and the order, perturbation, and strict-seriality experiments. A contrast is \emph{CI-solid} when its 95\,\% paired bootstrap interval over images excludes zero. Identifiers, prices, and costs are in \cref{app:cost}; Qwen3-VL-32B was withdrawn by the provider on 2026-10-09 and is absent from the controls added in revision.

\paragraph{Count metrics.} Exact-match accuracy (an unparseable answer counts as wrong), mean absolute error (MAE), and the sign of the error. On DotArrays we summarise each model--arm accuracy-versus-$n$ curve by its \emph{counting range} $R_{80}$: the largest $n$ up to which a non-increasing (isotonic) fit of exact accuracy stays at or above 80\,\%, the analogue of a developmental milestone; $R_{80}$ is right-censored when the fit is still above 80\,\% at the largest tested $n$, and the battery is extended for the two models this concerns. Bootstrap confidence intervals accompany $R_{80}$ (resampling images within each level) and every headline arm contrast (paired resampling over images).

\paragraph{Principle audit.} For every arm that produces points or tags and every image with point ground truth (DotArrays, FSC-147), we match tags to objects by minimum-cost assignment with a per-image distance threshold $\tau$ (half the median nearest-neighbour distance among ground-truth points, floor 15\,px). A matched tag is \emph{on-target}; an unmatched tag within $\tau$ of an already-matched object is a \emph{duplicate} (double count); an unmatched tag farther than $\tau$ from every object is a \emph{phantom}; an unmatched object is an \emph{omission} (skip). We report these per 100 ground-truth objects. \emph{Cardinality consistency} is the fraction of trajectories whose announced count equals the number of tags; \emph{pointing error} is the median, over images, of the per-image median distance between an on-target tag and its object, in thousandths of the shorter image side (DotArrays and FSC-147 pooled); \emph{abstraction} and \emph{arrangement} effects come from the DotArrays factor sets. \emph{Order irrelevance} is tested directly: the \arm{tag} loop is instructed to traverse the array left-to-right/top-to-bottom or the reverse; the realised traversal is read off the tag sequence, and the principle demands the same count either way (a same-order repeat control separates order dependence from inference instability). A $180^\circ$-rotation control counts the rotated image in reading order, which visits the objects in the reverse sequence under the natural instruction, separating order dependence from the difficulty of an unnatural scan; \emph{mirror consistency} (same count on an image and its horizontal mirror) is a coarser invariance probe. Finally, a \emph{tag-perturbation probe} asks whether the announced number is read off the tags at all: after the model's first tag call the harness silently removes one on-target tag, or adds one on an untagged object, before returning the image. Sensitivity to $\tau$ is reported in \cref{app:audit}.

\begin{table}[t]
\centering\tiny
\setlength{\tabcolsep}{4.5pt}
\renewcommand{\arraystretch}{0.96}
\caption{Exact-count accuracy (\%) per model and arm on the common subset every model saw (\nCommon{} images: 50 DotArrays, 100 FSC-147, 90 PixMo-Count); best arm per row in bold; $\uparrow$/$\downarrow$ mark CI-solid changes against \arm{direct} (95\,\% paired bootstrap CI over images excludes zero; all intervals in \cref{tab:delta}); unparseable answers count as wrong; tier-2/3 models ran fewer arms. MAE, \arm{tag+crop}, and \arm{tiles} are in \cref{tab:maincommonfull,tab:mainfull}.}
\label{tab:main}
\begin{tabular}{@{}lrrrrr@{}}
\toprule
model & direct & CoT & list & tag & full toolbox \\
\midrule\multicolumn{6}{@{}l}{\emph{DotArrays (synthetic, $n\le60$)}} \\
Qwen3-VL-8B & \textbf{52.0} & 22.0$^{\downarrow}$ & 26.0 & 38.0$^{\downarrow}$ & 32.0$^{\downarrow}$ \\
Qwen3-VL-32B & \textbf{48.0} & 28.0$^{\downarrow}$ & 44.0 & 36.0 & 34.0 \\
Gemini 3.1 Flash-Lite & 36.0 & \textbf{58.0$^{\uparrow}$} & 54.0 & 50.0$^{\uparrow}$ & 52.0$^{\uparrow}$ \\
GPT-5 mini & \textbf{48.0} & 40.0 & 40.0 & 18.0$^{\downarrow}$ & 28.0$^{\downarrow}$ \\
Gemini 3.7 Flash & \textbf{100.0} & 98.0 & 94.0 & 94.0 & 96.0 \\
Gemini 3.1 Pro & 96.0 & 96.0 & -- & 96.0 & \textbf{100.0} \\
Grok 4.20 & \textbf{66.0} & \textbf{66.0} & -- & 50.0$^{\downarrow}$ & 42.0$^{\downarrow}$ \\
Grok 4.5 & 54.0 & \textbf{66.0} & -- & 28.0$^{\downarrow}$ & 44.0 \\
Claude Sonnet 5 & 72.0 & \textbf{80.0} & -- & 62.0 & 58.0 \\
\midrule\multicolumn{6}{@{}l}{\emph{FSC-147 (photos, 8--80 objects)}} \\
Qwen3-VL-8B & 11.0 & 12.0 & 16.0 & \textbf{17.0} & 12.0 \\
Qwen3-VL-32B & 25.0 & 21.0 & 27.0 & \textbf{32.0} & 26.0 \\
Gemini 3.1 Flash-Lite & 23.0 & \textbf{41.0$^{\uparrow}$} & 31.0 & 22.0 & 29.0 \\
GPT-5 mini & 12.0 & 12.0 & \textbf{14.0} & 12.0 & 8.0 \\
Gemini 3.7 Flash & \textbf{52.0} & \textbf{52.0} & 47.0 & 45.0 & 47.0 \\
Gemini 3.1 Pro & \textbf{51.0} & \textbf{51.0} & -- & 50.0 & 49.0 \\
Grok 4.20 & 22.0 & \textbf{24.0} & -- & 19.0 & 13.0$^{\downarrow}$ \\
Grok 4.5 & 14.0 & \textbf{19.0} & -- & 12.0 & 12.0 \\
Claude Sonnet 5 & 24.0 & \textbf{25.0} & -- & 21.0 & 21.0 \\
\midrule\multicolumn{6}{@{}l}{\emph{PixMo-Count (photos, 2--10 objects)}} \\
Qwen3-VL-8B & 62.2 & \textbf{64.4} & 53.3 & 57.8 & 53.3 \\
Qwen3-VL-32B & 65.6 & 75.6 & \textbf{81.1} & 78.9$^{\uparrow}$ & 78.9$^{\uparrow}$ \\
Gemini 3.1 Flash-Lite & 71.1 & \textbf{78.9} & 72.2 & 71.1 & 77.8 \\
GPT-5 mini & 66.7 & 65.6 & \textbf{68.9} & 65.6 & 61.1 \\
Gemini 3.7 Flash & \textbf{88.9} & \textbf{88.9} & 87.8 & 86.7 & 83.3 \\
Gemini 3.1 Pro & \textbf{92.2} & \textbf{92.2} & -- & 90.0 & \textbf{92.2} \\
Grok 4.20 & 74.4 & 75.6 & -- & 75.6 & \textbf{78.9} \\
Grok 4.5 & 58.9 & \textbf{76.7$^{\uparrow}$} & -- & 67.8 & 68.9$^{\uparrow}$ \\
Claude Sonnet 5 & 74.4 & 76.7 & -- & \textbf{82.2} & 78.9 \\
\bottomrule
\end{tabular}
\end{table}

\section{Results}
\label{sec:results}
\subsection{The exact-counting range is a model-specific signature}
\label{sec:res-cliff}
\Cref{fig:curvesall} (appendix) plots exact accuracy against $n$ on the DotArrays main battery for every model; \cref{fig:r80} summarises each curve by $R_{80}$ (CIs in \cref{tab:reighty}) and \cref{fig:xl} shows the two strongest models on the extended range. With \arm{direct} answers every model is at ceiling for $n\le4$, but the range over which it stays exact spans nearly two orders of magnitude: $R_{80}$ is \rxGrokFourDirect{} for Grok 4.20, \rxQwenEightDirect{}--\rxQwenThirtyTwoDirect{} for the Qwen models, and \rxGemLiteDirect{} for Gemini 3.1 Flash-Lite and GPT-5 mini, while the two strongest Geminis are still at or near 80\,\% at $n{=}60$, so the original battery only right-censors their range. Extending it to 200 dots (\cref{fig:xl}, appendix) locates their cliffs as well: $R_{80}{=}\rxGemFlashDirect{}$ (95\,\% CI \rxCiGemFlashDirect{}) for Gemini 3.7 Flash and \rxGemProDirect{} (CI \rxCiGemProDirect{}, wide because one mid-battery level dips) for Gemini 3.1 Pro, with the usual under-counting beyond. At this scale the \emph{verbal} routine breaks before the answer-only arm does: Gemini 3.7 Flash's CoT range is \rxGemFlashCot{} versus \rxGemFlashDirect{} (\cref{sec:discussion}). Beyond its cliff, every model mostly under-counts, and photographs move every cliff to the left: on FSC-147 (8--80 objects) \arm{direct} exact accuracy on the common subset ranges from \acccmQwenEightDirectFsc{}\,\% (Qwen3-VL-8B) to \acccmGemFlashDirectFsc{}\,\% (Gemini 3.7 Flash), and even the strongest models fall to about 20\,\% in the 61--80 bin (\cref{tab:fscbins}); PixMo-Count, whose counts lie inside or just beyond the subitizing range, is much easier (\acccmQwenEightDirectPix{}--\acccmGemProDirectPix{}\,\%).

\paragraph{What \arm{direct} measures.} \arm{direct} is answer-only, not reasoning-free: at low effort the provider reports a median of \rtokDirectGemFlash{} hidden reasoning tokens per answer for Gemini 3.7 Flash (\rtokDirectGemFlashOneFifty{} at $n{=}150$), \rtokDirectGemPro{} for Gemini 3.1 Pro, and \rtokDirectGrokFour{} for Grok 4.20, none for the Qwen models. A thinking-off control (\cref{tab:reason}) leaves Flash-Lite's accuracy unchanged but shrinks its range ($R_{80}$ \ntReightyLowGemLite{} to \ntReightyOffGemLite{}), costs GPT-5 mini \ntDeltaGptMini{}\,pp (CI \ntDeltaCiGptMini{}) and its range (\ntReightyLowGptMini{} to \ntReightyOffGptMini{}), and barely moves Gemini 3.7 Flash, which cannot be switched off. The contrasts below are thus ``no visible routine'' versus ``an externalised one''.

\begin{table}[tb]
\centering\scriptsize
\setlength{\tabcolsep}{2.6pt}
\caption{Capacity profile per model, with 95\,\% bootstrap CIs over images: counting range $R_{80}$ under \arm{direct} ($\ge$: right-censored at the largest $n$ tested); pointing error of the \arm{tag} arm (median over images of the per-image median on-target tag-to-object distance, in thousandths of the shorter side; DotArrays + FSC-147); omitted objects, duplicate and phantom tags per 100 ground-truth objects (\arm{tag}, DotArrays + FSC-147); cardinality (announced count $=$ number of tags, \%); order effect (excess disagreement from reversing the traversal beyond the same-order repeat control, pp; \cref{tab:order}). Tier-3 models ran the common subset only.}
\label{tab:profile}
\begin{tabular}{@{}lrrrrrrr@{}}
\toprule
model & $R_{80}$ & point.\ err & omitted & duplicate & phantom & card.\ (\%) & order (pp) \\
\midrule
Qwen3-VL-8B & 6 {\tiny[6,7]} & 11.9 {\tiny[10.6,12.9]} & 46.6 {\tiny[42.6,50.5]} & 10.6 {\tiny[7.8,14.0]} & 45.3 {\tiny[37.3,54.4]} & 95.1 {\tiny[93.0,96.9]} & +51.5 {\tiny[41.2,61.8]} \\
Qwen3-VL-32B & 7 {\tiny[6,9]} & 8.7 {\tiny[7.6,10.3]} & 30.8 {\tiny[26.9,34.6]} & 29.7 {\tiny[22.6,37.5]} & 85.7 {\tiny[70.8,101.3]} & 81.0 {\tiny[77.3,84.5]} & +15.4 {\tiny[8.1,23.2]} \\
Gemini 3.1 Flash-Lite & 7 {\tiny[4,9]} & 5.1 {\tiny[4.0,7.0]} & 27.6 {\tiny[24.1,30.9]} & 3.9 {\tiny[2.4,5.9]} & 21.2 {\tiny[17.6,24.9]} & 88.5 {\tiny[85.8,91.3]} & +21.7 {\tiny[13.6,30.5]} \\
GPT-5 mini & 7 {\tiny[3,12]} & 40.4 {\tiny[38.3,43.8]} & 79.4 {\tiny[76.7,81.9]} & 2.2 {\tiny[1.7,2.8]} & 31.9 {\tiny[28.3,35.7]} & 64.7 {\tiny[60.1,69.1]} & -- \\
Gemini 3.7 Flash & 150 {\tiny[100,200]} & 1.6 {\tiny[1.5,1.8]} & 14.5 {\tiny[11.2,17.9]} & 1.3 {\tiny[0.6,2.2]} & 12.6 {\tiny[9.6,15.7]} & 89.9 {\tiny[87.1,92.5]} & +0.7 {\tiny[-1.5,3.3]} \\
Gemini 3.1 Pro & 150 {\tiny[50,200]} & 10.7 {\tiny[7.7,12.8]} & 22.1 {\tiny[15.9,28.4]} & 1.4 {\tiny[0.5,2.6]} & 22.4 {\tiny[16.6,28.3]} & 89.3 {\tiny[84.0,94.0]} & -- \\
Grok 4.20 & 3 {\tiny[3,20]} & 34.1 {\tiny[31.7,38.4]} & 69.1 {\tiny[64.9,73.2]} & 2.9 {\tiny[1.7,4.4]} & 52.7 {\tiny[47.5,58.1]} & 93.3 {\tiny[89.3,97.3]} & -- \\
Grok 4.5 & 6 {\tiny[6,15]} & 29.4 {\tiny[27.1,32.7]} & 47.4 {\tiny[43.0,51.6]} & 10.2 {\tiny[6.9,14.1]} & 100.4 {\tiny[92.7,108.5]} & 37.3 {\tiny[30.0,45.3]} & -- \\
Claude Sonnet 5 & 20 {\tiny[6,50]} & 33.9 {\tiny[30.7,39.4]} & 78.0 {\tiny[74.0,81.6]} & 8.1 {\tiny[5.8,10.6]} & 55.2 {\tiny[46.5,64.4]} & 51.3 {\tiny[43.3,59.3]} & -- \\
\bottomrule
\end{tabular}
\end{table}

\subsection{A finger helps some models in some regimes, and hurts others}
\label{sec:res-tools}
\Cref{tab:main} compares the arms on the common subset (\cref{tab:maincommonfull,fig:delta} give MAE and paired CIs); three regimes emerge. \emph{Small-count photographs (PixMo-Count).} The tag loop raises the point estimate for \nTagUpPix{} of the nine models, but only the Qwen3-VL-32B gain is CI-solid (\acccmQwenThirtyTwoDirectPix{} to \acccmQwenThirtyTwoTagPix{}\,\%; \dAccQwenThirtyTwoTagPix{}\,pp, 95\,\% CI \dAccCiQwenThirtyTwoTagPix{}); Flash-Lite is flat, and the two Gemini models near ceiling, GPT-5 mini, and Qwen3-VL-8B do not gain. \emph{Dense photographs (FSC-147).} No tool arm moves any model reliably: the largest tag gains (Qwen3-VL-32B \dAccQwenThirtyTwoTagFsc{}\,pp, Qwen3-VL-8B \dAccQwenEightTagFsc{}\,pp) carry CIs that span zero, the rest are flat or slightly worse, and verbal chain of thought is the strongest single intervention for Flash-Lite (\acccmGemLiteCotFsc{}\,\%, \dAccGemLiteCotFsc{}\,pp \dAccCiGemLiteCotFsc{}). \emph{Dot arrays.} The success stories of \cref{fig:hero,fig:hero2} (appendix) are real but not the average: Flash-Lite moves from \acccmGemLiteDirectDot{} to \acccmGemLiteTagCropDot{}\,\% (\arm{tag+crop}; \arm{tag} \dAccGemLiteTagDot{}\,pp \dAccCiGemLiteTagDot{}) and its counting range from $R_{80}{=}\rEightyGemLiteDirect{}$ to \rEightyGemLiteFull{}, but every other model is flat (Gemini 3.1 Pro) or loses, CI-solidly so for GPT-5 mini (\dAccGptMiniTagDot{}\,pp), Grok 4.20, Grok 4.5, and Qwen3-VL-8B (\cref{tab:delta}). \emph{Seeing the marks versus listing them.} Blind pointing (\arm{list}) matches or beats \arm{tag} in \nListGeTag{} of the \nTagListCells{} cells where both ran (\cref{tab:taglist}); the paired difference is CI-solid for \arm{list} for Flash-Lite on FSC-147 (\dTagListGemLiteFsc{}\,pp, CI \dTagListCiGemLiteFsc{}) and GPT-5 mini on dots (\dTagListGptMiniDot{}\,pp), and for \arm{tag} only for Qwen3-VL-8B on dots (\dTagListQwenEightDot{}\,pp). Seeing the marks buys little accuracy; it buys a cleaner trace (on FSC-147, Flash-Lite's blind lists carry \phantomGemLiteListFsc{} phantom points per 100 objects against \phantomGemLiteTagFsc{} for \arm{tag}), and the rendered, numbered trace is what the audit reads. The scripted partition (\arm{tiles}) is the best arm on sparse arrays for Flash-Lite (MAE \maeGemLiteTilesDot{}) but collapses on photographs (\accGemLiteTilesFsc{}\,\% on FSC-147): objects cut by tile borders are double counted and empty tiles invite phantoms; on PixMo-Count, where most objects exceed the tile overlap, it double counts by construction (mean answer \tilesPixMeanAnsGemLite{} for a truth of \tilesPixMeanGtGemLite{}) and is not reported.

\subsection{The principle audit explains who benefits}
\label{sec:res-audit}
\Cref{tab:profile} gives each model's capacity profile with bootstrap intervals (per dataset and arm: \cref{fig:audit,tab:audit}), and \cref{fig:precision} (appendix) plots each model's gain from tagging against its pointing error. \emph{(i) Pointing precision is the prerequisite.} The median on-target tag lands \perrTagGemFlash{} thousandths of the image side from its object for Gemini 3.7 Flash and \perrTagGemLite{} for Flash-Lite, but \perrTagGrokFour{} for Grok 4.20 and \perrTagGptMini{} for GPT-5 mini; the imprecise pointers add \phantomProfTagGptMini{}--\phantomProfTagGrokFour{} phantom tags per 100 objects (\cref{tab:profile}) and are exactly the models that lose accuracy with the tag loop. \emph{(ii) In dense scenes the routine breaks by omission.} For the models that can point, the dominant residual on FSC-147 is skipping (\omitGemLiteTagFsc{} omissions per 100 objects for Flash-Lite, \omitGemProTagFsc{} for Gemini 3.1 Pro), then phantoms; duplicates are rare (\dupGemLiteTagFsc{} and \dupGemProTagFsc{}), as in children, who lose their place more often than they double count. \emph{(iii) Models rarely revise after looking:} only \multiGemLiteTagAll{}\,\% of Flash-Lite trajectories issue a second tag call (\multiGemProTagAll{}\,\% for Gemini 3.1 Pro): one-shot tagging without the checking step children learn. \emph{(iv) Cardinality is mostly respected} (\cardGemLiteTagDot{}\,\% of Flash-Lite and \cardGemProTagDot{}\,\% of Gemini 3.1 Pro trajectories on DotArrays announce exactly the number of tags), but the imprecise pointers decouple the two (\cardGptMiniTagDot{}\,\% for GPT-5 mini): they tag, distrust the tags, and estimate anyway. Since no tool deletes a tag, a departure from the tag count may be a correction, and mostly is (closer to the truth in \closerPooled{}\,\% of cases; \cref{tab:coupling}). \emph{(v) The answer is read off the tags, not the tags off the answer.} A model could estimate a number and draw that many points; two tests say the precise pointers do not. On FSC-147 the tag count equals the model's own \arm{direct} answer on only \tagEqDirectFscMin{}--\tagEqDirectFscMax{}\,\% of images, with per-image errors weakly to moderately correlated ($\rho$ \rhoFscMin{} to \rhoFscMax{}; \cref{tab:coupling}). And when the harness silently removes one on-target tag after the first tag call (\cref{tab:perturb}), Gemini 3.7 Flash re-tags the missing object in \pertRepairAllGemFlashMinus{}\,\% of trajectories and Flash-Lite in \pertRepairAllGemLiteMinus{}\,\%, and their final answer equals the number of tags visible, not the number emitted, in \pertFollowImgAllGemFlashMinus{} and \pertFollowImgAllGemLiteMinus{}\,\% (where Flash-Lite does not repair, its answer moves down with the tag: \pertShiftGemLiteDotMinus{} on dots relative to the unperturbed run); GPT-5 mini, the decoupled pointer, repairs in only \pertRepairAllGptMiniMinus{}\,\%. For the models that can point, the trace is the computation.

\subsection{Developmental signatures and ablations}
\label{sec:res-dev}
\paragraph{Arrangement.} Children count regular arrays more accurately than scattered ones \citep{potter1968spatial}. \Cref{fig:arrangement} (appendix) shows a qualitatively analogous ordering for models once they count with tags (regular $>$ random $>$ clustered; Flash-Lite: \accRegGemLiteTag{}/\accRndGemLiteTag{}/\accCluGemLiteTag{}\,\%), whereas \arm{direct} estimation shows no advantage for regular arrays (\accRegGemLiteDirect{}/\accRndGemLiteDirect{}/\accCluGemLiteDirect{}\,\%): the arrangement effect is a signature of the routine, not of the number sense. Two controls rule out a rows-times-columns shortcut (\cref{tab:lattice,tab:jitter}). The lattices are jittered and the tags follow the jitter: Flash-Lite's on-target tags land \jitDgtGemLite{}\,px from the dot but \jitDidealGemLite{}\,px from the cell centre, with tag and dot offsets correlated at $r{=}\jitCorrGemLite{}$ (\jitCorrGemFlash{} for Gemini 3.7 Flash), so the marks sit on dots, not on an imagined grid. On new hexagonal and holed lattices, which offer no product to recite, Flash-Lite's tag advantage persists (\latHexGemLiteTag{} and \latHolesGemLiteTag{}\,\% against \latRndGemLiteTag{}\,\% on random arrays of the same $n$) but Qwen3-VL-8B's does not (\latHexQwenEightTag{}/\latHolesQwenEightTag{} vs.\ \latRndQwenEightTag{}\,\%), and a lattice rotated by 20--40$^\circ$ is hard for every model and arm (\latRotGemLiteTag{}\,\% under \arm{tag}): for the one model with a working routine the regular advantage is that of an organised scan path, which a rotated lattice no longer offers in reading order. \emph{Abstraction:} mixed shapes and red-among-blue sets cost \arm{direct} more than the tag arms (\cref{tab:family}).
\paragraph{Order irrelevance.} Instructed to tag the same \nOrderImages{} arrays in opposite traversal orders, models follow the procedure (tag order tracks the instructed scan, \cref{tab:order}). The principle behind the procedure fares worse: the two traversals end in the same count on \ordAgreeGemLite{}\,\% of arrays for Flash-Lite and \ordAgreeQwenThirtyTwo{}\,\% for Qwen3-VL-32B, against \ordAgreeGemFlash{}\,\% for Gemini 3.7 Flash. Two controls separate order dependence from other causes (\cref{tab:order}). Repeating the \emph{same} order agrees on \ordSameAgreeGemLite{}/\ordSameAgreeQwenThirtyTwo{}/\ordSameAgreeGemFlash{}\,\% of arrays, so reversal adds \ordEffectGemLite{}\,pp (CI \ordEffectCiGemLite{}) and \ordEffectQwenThirtyTwo{}\,pp (\ordEffectCiQwenThirtyTwo{}) of disagreement beyond inference instability for the mid-tier models and none for Gemini 3.7 Flash (\ordEffectGemFlash{}\,pp). Because the reverse scan is simply harder (Flash-Lite drops from \ordAccLtrGemLite{} to \ordAccRtlGemLite{}\,\%), a rotation control counts each array rotated by $180^\circ$ in ordinary reading order, visiting the dots in reverse under the natural instruction at undiminished accuracy (\ordRotAccGemLite{}\,\%); it still disagrees with the unrotated runs \ordRotEffectGemLite{}\,pp (CI \ordRotEffectCiGemLite{}) more often than the repeat control for Flash-Lite (\ordRotEffectGemFlash{}\,pp for Gemini 3.7 Flash): part of the raw reversal effect was difficulty, but an order effect remains. The coarser mirror probe agrees: a horizontal flip, which changes the view but not the set, changes Flash-Lite's answer in most trajectories (mirror consistency \orderAgreeGemLiteDirect{}\,\% under \arm{direct}, \orderAgreeGemLiteTag{}\,\% under \arm{tag}; \cref{tab:flip}).
\paragraph{Ablations.}
\label{sec:res-ablation}
\Cref{tab:ablation} varies the harness on the ablation subset for the two tier-1 models with usable pointing. \emph{Removing} the verification prompt raises Flash-Lite's accuracy (\accabGemLiteTagAll{} to \accabGemLiteTagNoNudgeAll{}\,\%), not because the checked arm adds tags (it draws no more and re-tags less often, \ndNudgeMultiGemLite{} vs.\ \ndPlainMultiGemLite{}\,\%) but because, asked whether any object is still untagged, the model re-estimates verbally instead of re-inspecting: it announces more than it tagged in \ndNudgeAnsGtTagsGemLite{}\,\% of trajectories against \ndPlainAnsGtTagsGemLite{}\,\% without the prompt, and is then right only \ndNudgeAccNeqGemLite{}\,\% of the time. Plain marks match numbered tags on accuracy but lower cardinality consistency; the grid is the most useful addition (\accabGemLiteTagGridAll{}\,\%); \tool{erase}, \tool{crop}, and the turn budget hardly matter. Strict seriality (\arm{tag-1}, one point per look, $T{=}80$; \cref{tab:tagone}) changes the errors more than the accuracy (Flash-Lite \accTOnecmpGemLite{} vs.\ \accTcmpGemLite{}\,\%): phantoms collapse (\phTOnecmpGemLite{} vs.\ \phTcmpGemLite{} per 100 objects) and the failure becomes abandoning the traversal (Gemini 3.7 Flash leaves \omitTOnecmpGemFlash{}\,\% untagged yet answers well), at \tagOneCostRatioGemLite{}--\tagOneCostRatioGemFlash{}$\times$ the cost; batching is not what breaks the routine, and the batched trace is a fair record of it.

\section{Discussion and limitations}
\label{sec:discussion}
\paragraph{What transfers from the developmental account, and what does not.} Both a VLM and a pre-procedural child have a serviceable approximate number sense and no reliable serial routine, so the same external scaffolds can help or fail to help both. Pointing helps only if pointing is precise: children whose gestures are not yet coordinated with their number words gain less from gesturing \citep{alibali1999function}, and neither do our imprecise pointers (\cref{tab:profile}). Two failures our audit quantified, answers that come apart from the tags and the missing habit of checking one's count, have close parallels in preschoolers \citep{gelman1983preschoolers}. Two parts do not transfer: children practise and internalise the routine, whereas our models neither learn from the tags nor carry the routine to the next image (a prosthesis, not a curriculum); and the model's ``motor'' error is a coordinate-prediction error without proprioceptive feedback, hence the outsized effect of the coordinate frame and the grid.

\paragraph{Why the verbal routine collapses first at scale.} Gemini 3.7 Flash's verbal cliff (CoT $R_{80}{=}\rxGemFlashCot{}$ against \rxGemFlashDirect{}, \cref{fig:xl}) looks backwards only if chain of thought is a count. Past about 80 dots it is not: of the \cotXlN{} CoT replies at $n\ge80$, the \cotRowN{} that tally row by row are all exact, whereas the other \cotRegN{} partition the image into regions, estimate each, and add the estimates, which is exact \cotRegExact{} times and under-counts by \cotXlMeanErr{} on average. A verbal routine has no external record, so beyond the span it can tally in text it falls back on the approximate system applied piecewise, whose under-counting bias accumulates, a verbal analogue of children's keeping-track failures \citep{fuson1988children}; the answer-only arm, free to spend hidden reasoning (\cref{app:reason}), and the tag loop, which externalises the record, degrade more slowly. A routine helps only while its bookkeeping holds, and external marks extend it. Three levers follow from the audit: calibrate pointing (a grid helped), train the verification step (the current prompt makes the model re-estimate rather than re-inspect), and combine the verbal and visual routines.

\paragraph{Limitations.} All models are accessed through one API at temperature 0 with low reasoning effort; \arm{direct} is answer-only rather than reasoning-free, two endpoints could not have reasoning disabled (\cref{tab:reason}), and Qwen3-VL-32B was withdrawn by the provider during revision. The dot-array battery is deliberately simple, the photographic sets cover only 8--80 and 2--10 objects, FSC-147 dots mark approximate centres, so tags on elongated objects can be scored as phantoms ($\tau$-sensitivity in \cref{app:audit}). Finally, the developmental alignment is in the \emph{evaluation}: our claims are qualitative correspondences with the child literature, not psychometric fits, and all subjects are frozen general-purpose VLMs; running the battery on models trained on child-scale data, and asking whether they could \emph{acquire} the routine rather than borrow it, is the natural next step.

\section{Conclusion}
Dumb tools let one mid-tier VLM count fifty dots exactly where it used to guess, and left most others flat or worse. The lasting contribution is what the marks make visible: read through the developmental principles of counting, a wrong count becomes a diagnosis --- imprecise pointing, lost track, an unchecked answer, an order-dependent count. Can a model \emph{learn} the routine, and the habit of checking it, as children do?

\bibliographystyle{plainnat}
\bibliography{refs}

\begin{thebibliography}{61}
\providecommand{\natexlab}[1]{#1}
\providecommand{\url}[1]{\texttt{#1}}
\expandafter\ifx\csname urlstyle\endcsname\relax
  \providecommand{\doi}[1]{doi: #1}\else
  \providecommand{\doi}{doi: \begingroup \urlstyle{rm}\Url}\fi

\bibitem[Acharya et~al.(2019)Acharya, Kafle, and Kanan]{acharya2019tallyqa}
Manoj Acharya, Kushal Kafle, and Christopher Kanan.
\newblock {TallyQA}: Answering complex counting questions.
\newblock In \emph{Proceedings of the AAAI Conference on Artificial
  Intelligence}, volume~33, pages 8076--8084, 2019.
\newblock \doi{10.1609/aaai.v33i01.33018076}.

\bibitem[Alghisi et~al.(2026)Alghisi, Rizzoli, Mousavi, and
  Riccardi]{alghisi2026getting}
Simone Alghisi, Massimo Rizzoli, Seyed~Mahed Mousavi, and Giuseppe Riccardi.
\newblock Getting to the point: Pointing improves {LVLMs} at counting.
\newblock \emph{arXiv preprint arXiv:2603.21746}, 2026.

\bibitem[Alibali and DiRusso(1999)]{alibali1999function}
Martha~Wagner Alibali and Alyssa~A. DiRusso.
\newblock The function of gesture in learning to count: More than keeping
  track.
\newblock \emph{Cognitive Development}, 14\penalty0 (1):\penalty0 37--56, 1999.
\newblock \doi{10.1016/S0885-2014(99)80017-3}.

\bibitem[Beckwith and Restle(1966)]{beckwith1966process}
Mary Beckwith and Frank Restle.
\newblock Process of enumeration.
\newblock \emph{Psychological Review}, 73\penalty0 (5):\penalty0 437--444,
  1966.
\newblock \doi{10.1037/h0023650}.

\bibitem[Bolya et~al.(2020)Bolya, Foley, Hays, and Hoffman]{bolya2020tide}
Daniel Bolya, Sean Foley, James Hays, and Judy Hoffman.
\newblock {TIDE}: A general toolbox for identifying object detection errors.
\newblock In \emph{European Conference on Computer Vision (ECCV)}, pages
  558--573. Springer, 2020.
\newblock \doi{10.1007/978-3-030-58580-8_33}.

\bibitem[Budny et~al.(2025)Budny, Ghods, Campbell, Marjieh, Joshi, Kumar,
  Cohen, Webb, and Griffiths]{budny2025visual}
Nicholas Budny, Kia Ghods, Declan Campbell, Raja Marjieh, Amogh Joshi, Sreejan
  Kumar, Jonathan~D. Cohen, Taylor~W. Webb, and Thomas~L. Griffiths.
\newblock Visual serial processing deficits explain divergences in human and
  {VLM} reasoning.
\newblock \emph{arXiv preprint arXiv:2509.25142}, 2025.

\bibitem[Carey(2009)]{carey2009origin}
Susan Carey.
\newblock \emph{The Origin of Concepts}.
\newblock Oxford University Press, New York, 2009.
\newblock \doi{10.1093/acprof:oso/9780195367638.001.0001}.

\bibitem[Che et~al.(2026)Che, Xue, Quan, Liu, Shi, Hurst, Feldman, Tang,
  Krishna, and Pavlovic]{che2026counting}
Liwei Che, Zhiyu Xue, Yihao Quan, Benlin Liu, Zeru Shi, Michelle Hurst, Jacob
  Feldman, Ruixiang Tang, Ranjay Krishna, and Vladimir Pavlovic.
\newblock Counting circuits: Mechanistic interpretability of visual reasoning
  in {LVLMs}.
\newblock \emph{arXiv preprint arXiv:2603.18523}, 2026.

\bibitem[Collins et~al.(2026)Collins, Bolton, Nguyen, Taesiri, Bui, and
  Nguyen]{collins2026sketchvlm}
Brandon Collins, Logan Bolton, Hung~Huy Nguyen, Mohammad~Reza Taesiri, Trung
  Bui, and Anh~Totti Nguyen.
\newblock {SketchVLM}: Vision language models can annotate images to explain
  thoughts and guide users.
\newblock \emph{arXiv preprint arXiv:2604.22875}, 2026.

\bibitem[Dehaene(2011)]{dehaene2011number}
Stanislas Dehaene.
\newblock \emph{The Number Sense: How the Mind Creates Mathematics}.
\newblock Oxford University Press, New York, revised and updated edition, 2011.

\bibitem[Deitke et~al.(2025)Deitke, Clark, Lee, Tripathi, Yang, Park, Salehi,
  Muennighoff, Lo, Soldaini, Lu, Anderson, Bransom, Ehsani, Ngo, Chen, Patel,
  Yatskar, Callison-Burch, Head, Hendrix, Bastani, VanderBilt, Lambert, Chou,
  Chheda, Sparks, Skjonsberg, Schmitz, Sarnat, Bischoff, Walsh, Newell,
  Wolters, Gupta, Zeng, Borchardt, Groeneveld, Nam, Lebrecht, Wittlif,
  Schoenick, Michel, Krishna, Weihs, Smith, Hajishirzi, Girshick, Farhadi, and
  Kembhavi]{deitke2024molmo}
Matt Deitke, Christopher Clark, Sangho Lee, Rohun Tripathi, Yue Yang, Jae~Sung
  Park, Mohammadreza Salehi, Niklas Muennighoff, Kyle Lo, Luca Soldaini, Jiasen
  Lu, Taira Anderson, Erin Bransom, Kiana Ehsani, Huong Ngo, YenSung Chen, Ajay
  Patel, Mark Yatskar, Chris Callison-Burch, Andrew Head, Rose Hendrix, Favyen
  Bastani, Eli VanderBilt, Nathan Lambert, Yvonne Chou, Arnavi Chheda, Jenna
  Sparks, Sam Skjonsberg, Michael Schmitz, Aaron Sarnat, Byron Bischoff, Pete
  Walsh, Chris Newell, Piper Wolters, Tanmay Gupta, Kuo-Hao Zeng, Jon
  Borchardt, Dirk Groeneveld, Crystal Nam, Sophie Lebrecht, Caitlin Wittlif,
  Carissa Schoenick, Oscar Michel, Ranjay Krishna, Luca Weihs, Noah~A. Smith,
  Hannaneh Hajishirzi, Ross Girshick, Ali Farhadi, and Aniruddha Kembhavi.
\newblock Molmo and {PixMo}: Open weights and open data for state-of-the-art
  vision-language models.
\newblock In \emph{Proceedings of the IEEE/CVF Conference on Computer Vision
  and Pattern Recognition (CVPR)}, 2025.
\newblock URL \url{https://arxiv.org/abs/2409.17146}.

\bibitem[Deng et~al.(2026)Deng, Qiao, and Wan]{deng2026holocount}
Jinhong Deng, Limeng Qiao, and Guanglu Wan.
\newblock {HoloCount}: A holistic visual counting benchmark for {MLLMs}.
\newblock \emph{arXiv preprint arXiv:2607.06420}, 2026.

\bibitem[El-Shangiti et~al.(2026)El-Shangiti, Nurgazy, AlQuabeh, Rozanov, and
  Inui]{elshangiti2026count}
Ahmed~Oumar El-Shangiti, Abzal Nurgazy, Hilal AlQuabeh, Nikolai Rozanov, and
  Kentaro Inui.
\newblock The count is there, but misaligned: Understanding and correcting
  counting failures in {VLMs}.
\newblock \emph{arXiv preprint arXiv:2607.09544}, 2026.

\bibitem[Feigenson et~al.(2004)Feigenson, Dehaene, and
  Spelke]{feigenson2004core}
Lisa Feigenson, Stanislas Dehaene, and Elizabeth Spelke.
\newblock Core systems of number.
\newblock \emph{Trends in Cognitive Sciences}, 8\penalty0 (7):\penalty0
  307--314, 2004.
\newblock \doi{10.1016/j.tics.2004.05.002}.

\bibitem[{Fetrat Qharabagh} et~al.(2024){Fetrat Qharabagh}, Ghofrani, and
  Fountoulakis]{fetrat2024lvlmcount}
Muhammad {Fetrat Qharabagh}, Mohammadreza Ghofrani, and Kimon Fountoulakis.
\newblock {LVLM-Count}: Enhancing the counting ability of large vision-language
  models.
\newblock \emph{arXiv preprint arXiv:2412.00686}, 2024.

\bibitem[Fuson(1988)]{fuson1988children}
Karen~C. Fuson.
\newblock \emph{Children's Counting and Concepts of Number}.
\newblock Springer-Verlag, New York, 1988.
\newblock \doi{10.1007/978-1-4612-3754-9}.

\bibitem[Gelman and Gallistel(1978)]{gelman1978child}
Rochel Gelman and C.~R. Gallistel.
\newblock \emph{The Child's Understanding of Number}.
\newblock Harvard University Press, Cambridge, MA, 1978.

\bibitem[Gelman and Meck(1983)]{gelman1983preschoolers}
Rochel Gelman and Elizabeth Meck.
\newblock Preschoolers' counting: Principles before skill.
\newblock \emph{Cognition}, 13\penalty0 (3):\penalty0 343--359, 1983.
\newblock \doi{10.1016/0010-0277(83)90014-8}.

\bibitem[Graham(1999)]{graham1999role}
Theresa~A. Graham.
\newblock The role of gesture in children's learning to count.
\newblock \emph{Journal of Experimental Child Psychology}, 74\penalty0
  (4):\penalty0 333--355, 1999.
\newblock \doi{10.1006/jecp.1999.2520}.

\bibitem[Guo et~al.(2025)Guo, Huang, Shi, Song, and Zhang]{guo2025your}
Xuyang Guo, Zekai Huang, Zhenmei Shi, Zhao Song, and Jiahao Zhang.
\newblock Your vision-language model can't even count to 20: Exposing the
  failures of {VLMs} in compositional counting.
\newblock \emph{arXiv preprint arXiv:2510.04401}, 2025.

\bibitem[Gupta and Kembhavi(2023)]{gupta2023visual}
Tanmay Gupta and Aniruddha Kembhavi.
\newblock Visual programming: Compositional visual reasoning without training.
\newblock In \emph{Proceedings of the IEEE/CVF Conference on Computer Vision
  and Pattern Recognition (CVPR)}, 2023.
\newblock URL \url{https://arxiv.org/abs/2211.11559}.

\bibitem[Hou et~al.(2026)Hou, Mi, Zorzi, Ballan, and
  Testolin]{hou2025assessing}
Kuinan Hou, Jing Mi, Marco Zorzi, Lamberto Ballan, and Alberto Testolin.
\newblock Assessing the visual enumeration abilities of specialized counting
  architectures and vision-language models.
\newblock In \emph{Pattern Recognition (Lecture Notes in Computer Science)},
  pages 498--512. Springer, 2026.
\newblock \doi{10.1007/978-3-032-31452-9_33}.
\newblock arXiv:2512.15254.

\bibitem[Hu et~al.(2024)Hu, Shi, Fu, Roth, Ostendorf, Zettlemoyer, Smith, and
  Krishna]{hu2024sketchpad}
Yushi Hu, Weijia Shi, Xingyu Fu, Dan Roth, Mari Ostendorf, Luke Zettlemoyer,
  Noah~A Smith, and Ranjay Krishna.
\newblock Visual sketchpad: Sketching as a visual chain of thought for
  multimodal language models.
\newblock In \emph{Advances in Neural Information Processing Systems
  (NeurIPS)}, 2024.
\newblock URL \url{https://arxiv.org/abs/2406.09403}.

\bibitem[Idrees et~al.(2018)Idrees, Tayyab, Athrey, Zhang, Al-Maadeed, Rajpoot,
  and Shah]{idrees2018composition}
Haroon Idrees, Muhmmad Tayyab, Kishan Athrey, Dong Zhang, Somaya Al-Maadeed,
  Nasir Rajpoot, and Mubarak Shah.
\newblock Composition loss for counting, density map estimation and
  localization in dense crowds.
\newblock In \emph{European Conference on Computer Vision (ECCV)}, pages
  544--559. Springer, 2018.
\newblock \doi{10.1007/978-3-030-01216-8_33}.

\bibitem[Jiang et~al.(2026)Jiang, Huo, Chen, Xiong, Zeng, Chen, Ren, Yu, and
  Zhang]{jiang2026rexomni}
Qing Jiang, Junan Huo, Xingyu Chen, Yuda Xiong, Zhaoyang Zeng, Yihao Chen,
  Tianhe Ren, Junzhi Yu, and Lei Zhang.
\newblock Detect anything via next point prediction.
\newblock In \emph{Proceedings of the IEEE/CVF Conference on Computer Vision
  and Pattern Recognition (CVPR)}, pages 25472--25483, 2026.
\newblock URL \url{https://arxiv.org/abs/2510.12798}.

\bibitem[Kaufman et~al.(1949)Kaufman, Lord, Reese, and
  Volkmann]{kaufman1949discrimination}
E.~L. Kaufman, M.~W. Lord, T.~W. Reese, and J.~Volkmann.
\newblock The discrimination of visual number.
\newblock \emph{The American Journal of Psychology}, 62\penalty0 (4):\penalty0
  498--525, 1949.
\newblock \doi{10.2307/1418556}.

\bibitem[Kirsh(1995)]{kirsh1995intelligent}
David Kirsh.
\newblock The intelligent use of space.
\newblock \emph{Artificial Intelligence}, 73\penalty0 (1--2):\penalty0 31--68,
  1995.
\newblock \doi{10.1016/0004-3702(94)00017-U}.

\bibitem[Kirsh and Maglio(1994)]{kirsh1994distinguishing}
David Kirsh and Paul Maglio.
\newblock On distinguishing epistemic from pragmatic action.
\newblock \emph{Cognitive Science}, 18\penalty0 (4):\penalty0 513--549, 1994.
\newblock \doi{10.1207/s15516709cog1804_1}.

\bibitem[Le~Corre and Carey(2007)]{lecorre2007one}
Mathieu Le~Corre and Susan Carey.
\newblock One, two, three, four, nothing more: An investigation of the
  conceptual sources of the verbal counting principles.
\newblock \emph{Cognition}, 105\penalty0 (2):\penalty0 395--438, 2007.
\newblock \doi{10.1016/j.cognition.2006.10.005}.

\bibitem[Lei et~al.(2025)Lei, Yang, Chen, Li, and Liu]{lei2024scaffolding}
Xuanyu Lei, Zonghan Yang, Xinrui Chen, Peng Li, and Yang Liu.
\newblock Scaffolding coordinates to promote vision-language coordination in
  large multi-modal models.
\newblock In \emph{Proceedings of the 31st International Conference on
  Computational Linguistics (COLING)}, pages 2886--2903, 2025.
\newblock URL \url{https://aclanthology.org/2025.coling-main.195}.

\bibitem[Liu et~al.(2025)Liu, Li, Xiao, Jian, Qin, Shao, Ding, and
  Zhou]{liu2024autonomous}
Jingming Liu, Yumeng Li, Boyuan Xiao, Yichang Jian, Ziang Qin, Tianjia Shao,
  Yao-Xiang Ding, and Kun Zhou.
\newblock Autonomous imagination: Closed-loop decomposition of
  visual-to-textual conversion in visual reasoning for multimodal large
  language models.
\newblock \emph{Transactions on Machine Learning Research}, 2025.
\newblock arXiv:2411.18142.

\bibitem[Mandler and Shebo(1982)]{mandler1982subitizing}
George Mandler and Billie~J. Shebo.
\newblock Subitizing: An analysis of its component processes.
\newblock \emph{Journal of Experimental Psychology: General}, 111\penalty0
  (1):\penalty0 1--22, 1982.
\newblock \doi{10.1037/0096-3445.111.1.1}.

\bibitem[Mondal et~al.(2026)Mondal, Nag, and Dutta]{mondal2026abacus}
Anindya Mondal, Sauradip Nag, and Anjan Dutta.
\newblock {ABACUS}: Adapting unified foundation model for bridging image count
  understanding and generation.
\newblock \emph{arXiv preprint arXiv:2606.23835}, 2026.

\bibitem[Nasr et~al.(2019)Nasr, Viswanathan, and Nieder]{nasr2019number}
Khaled Nasr, Pooja Viswanathan, and Andreas Nieder.
\newblock Number detectors spontaneously emerge in a deep neural network
  designed for visual object recognition.
\newblock \emph{Science Advances}, 5\penalty0 (5):\penalty0 eaav7903, 2019.
\newblock \doi{10.1126/sciadv.aav7903}.

\bibitem[Paiss et~al.(2023)Paiss, Ephrat, Tov, Zada, Mosseri, Irani, and
  Dekel]{paiss2023teaching}
Roni Paiss, Ariel Ephrat, Omer Tov, Shiran Zada, Inbar Mosseri, Michal Irani,
  and Tali Dekel.
\newblock Teaching {CLIP} to count to ten.
\newblock In \emph{Proceedings of the IEEE/CVF International Conference on
  Computer Vision (ICCV)}, 2023.
\newblock URL \url{https://arxiv.org/abs/2302.12066}.

\bibitem[Pang et~al.(2026)Pang, Hou, Wang, and Sachan]{pang2026unveiling}
Xingzhou Pang, Yifan Hou, Junling Wang, and Mrinmaya Sachan.
\newblock Unveiling the visual counting bottleneck in vision-language models.
\newblock In \emph{Proceedings of the International Conference on Machine
  Learning (ICML)}, 2026.
\newblock URL \url{https://arxiv.org/abs/2605.30170}.

\bibitem[Pothiraj et~al.(2025)Pothiraj, Stengel-Eskin, Cho, and
  Bansal]{pothiraj2025capture}
Atin Pothiraj, Elias Stengel-Eskin, Jaemin Cho, and Mohit Bansal.
\newblock {CAPTURe}: Evaluating spatial reasoning in vision language models via
  occluded object counting.
\newblock In \emph{Proceedings of the IEEE/CVF International Conference on
  Computer Vision (ICCV)}, pages 8001--8010, 2025.
\newblock \doi{10.1109/ICCV51701.2025.00750}.

\bibitem[Potter and Levy(1968)]{potter1968spatial}
Mary~C. Potter and Ellen~I. Levy.
\newblock Spatial enumeration without counting.
\newblock \emph{Child Development}, 39\penalty0 (1):\penalty0 265--272, 1968.
\newblock \doi{10.2307/1127377}.

\bibitem[Rahmanzadehgervi et~al.(2024)Rahmanzadehgervi, Bolton, Taesiri, and
  Nguyen]{rahmanzadehgervi2024vision}
Pooyan Rahmanzadehgervi, Logan Bolton, Mohammad~Reza Taesiri, and Anh~Totti
  Nguyen.
\newblock Vision language models are blind.
\newblock In \emph{Proceedings of the Asian Conference on Computer Vision
  (ACCV)}, pages 18--34, 2024.
\newblock URL \url{https://arxiv.org/abs/2407.06581}.

\bibitem[Ranjan et~al.(2021)Ranjan, Sharma, Nguyen, and
  Hoai]{ranjan2021learning}
Viresh Ranjan, Udbhav Sharma, Thu Nguyen, and Minh Hoai.
\newblock Learning to count everything.
\newblock In \emph{Proceedings of the IEEE/CVF Conference on Computer Vision
  and Pattern Recognition (CVPR)}, 2021.
\newblock URL \url{https://arxiv.org/abs/2104.08391}.

\bibitem[Sarnecka and Carey(2008)]{sarnecka2008how}
Barbara~W. Sarnecka and Susan Carey.
\newblock How counting represents number: What children must learn and when
  they learn it.
\newblock \emph{Cognition}, 108\penalty0 (3):\penalty0 662--674, 2008.
\newblock \doi{10.1016/j.cognition.2008.05.007}.

\bibitem[Saxe(1977)]{saxe1977developmental}
Geoffrey~B. Saxe.
\newblock A developmental analysis of notational counting.
\newblock \emph{Child Development}, 48\penalty0 (4):\penalty0 1512--1520, 1977.
\newblock \doi{10.2307/1128514}.

\bibitem[Shen et~al.(2025)Shen, Zhao, Zhao, Xu, Zhang, Zhu, and
  Yin]{shen2024zoomeye}
Haozhan Shen, Kangjia Zhao, Tiancheng Zhao, Ruochen Xu, Zilun Zhang, Mingwei
  Zhu, and Jianwei Yin.
\newblock {ZoomEye}: Enhancing multimodal {LLMs} with human-like zooming
  capabilities through tree-based image exploration.
\newblock In \emph{Proceedings of the Conference on Empirical Methods in
  Natural Language Processing (EMNLP)}, pages 6613--6629, 2025.
\newblock \doi{10.18653/v1/2025.emnlp-main.335}.

\bibitem[Song et~al.(2021)Song, Wang, Jiang, Wang, Tai, Wang, Li, Huang, and
  Wu]{song2021rethinking}
Qingyu Song, Changan Wang, Zhengkai Jiang, Yabiao Wang, Ying Tai, Chengjie
  Wang, Jilin Li, Feiyue Huang, and Yang Wu.
\newblock Rethinking counting and localization in crowds: A purely point-based
  framework.
\newblock In \emph{IEEE/CVF International Conference on Computer Vision
  (ICCV)}, pages 3345--3354, 2021.
\newblock \doi{10.1109/ICCV48922.2021.00335}.

\bibitem[Stoianov and Zorzi(2012)]{stoianov2012emergence}
Ivilin Stoianov and Marco Zorzi.
\newblock Emergence of a `visual number sense' in hierarchical generative
  models.
\newblock \emph{Nature Neuroscience}, 15\penalty0 (2):\penalty0 194--196, 2012.
\newblock \doi{10.1038/nn.2996}.

\bibitem[Su et~al.(2025)Su, Xia, Guo, Liu, Ma, Qu, Liu, Li, Zeng, Yang, Li,
  Cheng, Ji, He, and Fung]{su2025thinking}
Zhaochen Su, Peng Xia, Hangyu Guo, Zhenhua Liu, Yan Ma, Xiaoye Qu, Jiaqi Liu,
  Yanshu Li, Kaide Zeng, Zhengyuan Yang, Linjie Li, Yu~Cheng, Heng Ji, Junxian
  He, and Yi~R. Fung.
\newblock Thinking with images for multimodal reasoning: Foundations, methods,
  and future frontiers.
\newblock \emph{arXiv preprint arXiv:2506.23918}, 2025.

\bibitem[Surís et~al.(2023)Surís, Menon, and Vondrick]{suris2023vipergpt}
Dídac Surís, Sachit Menon, and Carl Vondrick.
\newblock {ViperGPT}: Visual inference via {Python} execution for reasoning.
\newblock In \emph{Proceedings of the IEEE/CVF International Conference on
  Computer Vision (ICCV)}, 2023.
\newblock URL \url{https://arxiv.org/abs/2303.08128}.

\bibitem[Tamarapalli et~al.(2025)Tamarapalli, Grover, Pande, and
  Yerramilli]{tamarapalli2025countqa}
Jayant~Sravan Tamarapalli, Rynaa Grover, Nilay Pande, and Sahiti Yerramilli.
\newblock {CountQA}: How well do {MLLMs} count in the wild?
\newblock \emph{arXiv preprint arXiv:2508.06585}, 2025.

\bibitem[Testolin et~al.(2020)Testolin, Zou, and
  McClelland]{testolin2020numerosity}
Alberto Testolin, Will~Y. Zou, and James~L. McClelland.
\newblock Numerosity discrimination in deep neural networks: Initial
  competence, developmental refinement and experience statistics.
\newblock \emph{Developmental Science}, 23\penalty0 (5):\penalty0 e12940, 2020.
\newblock \doi{10.1111/desc.12940}.

\bibitem[Testolin et~al.(2025)Testolin, Hou, and Zorzi]{testolin2025visual}
Alberto Testolin, Kuinan Hou, and Marco Zorzi.
\newblock Visual enumeration remains challenging for multimodal generative
  {AI}.
\newblock \emph{PLOS ONE}, 20\penalty0 (9):\penalty0 e0331566, 2025.
\newblock \doi{10.1371/journal.pone.0331566}.

\bibitem[Trick and Pylyshyn(1994)]{trick1994small}
Lana~M. Trick and Zenon~W. Pylyshyn.
\newblock Why are small and large numbers enumerated differently? {A}
  limited-capacity preattentive stage in vision.
\newblock \emph{Psychological Review}, 101\penalty0 (1):\penalty0 80--102,
  1994.
\newblock \doi{10.1037/0033-295X.101.1.80}.

\bibitem[van Oeffelen and Vos(1982)]{vanoeffelen1982configurational}
Michiel~P. van Oeffelen and Peter~G. Vos.
\newblock Configurational effects on the enumeration of dots: Counting by
  groups.
\newblock \emph{Memory \& Cognition}, 10\penalty0 (4):\penalty0 396--404, 1982.
\newblock \doi{10.3758/BF03202432}.

\bibitem[Vo et~al.(2025)Vo, Nguyen, Taesiri, Dang, Nguyen, and
  Kim]{vo2025vision}
An~Vo, Khai-Nguyen Nguyen, Mohammad~Reza Taesiri, Vy~Tuong Dang, Anh~Totti
  Nguyen, and Daeyoung Kim.
\newblock Vision language models are biased.
\newblock \emph{arXiv preprint arXiv:2505.23941}, 2025.

\bibitem[Wang et~al.(2021)Wang, Gao, Lin, and Li]{gao2020nwpu}
Qi~Wang, Junyu Gao, Wei Lin, and Xuelong Li.
\newblock {NWPU-Crowd}: A large-scale benchmark for crowd counting and
  localization.
\newblock \emph{IEEE Transactions on Pattern Analysis and Machine
  Intelligence}, 43\penalty0 (6):\penalty0 2141--2149, 2021.
\newblock \doi{10.1109/TPAMI.2020.3013269}.

\bibitem[Wu and Xie(2024)]{wu2024vstar}
Penghao Wu and Saining Xie.
\newblock {V*}: Guided visual search as a core mechanism in multimodal {LLMs}.
\newblock In \emph{Proceedings of the IEEE/CVF Conference on Computer Vision
  and Pattern Recognition (CVPR)}, 2024.
\newblock URL \url{https://arxiv.org/abs/2312.14135}.

\bibitem[Wynn(1990)]{wynn1990children}
Karen Wynn.
\newblock Children's understanding of counting.
\newblock \emph{Cognition}, 36\penalty0 (2):\penalty0 155--193, 1990.
\newblock \doi{10.1016/0010-0277(90)90003-3}.

\bibitem[Wynn(1992)]{wynn1992children}
Karen Wynn.
\newblock Children's acquisition of the number words and the counting system.
\newblock \emph{Cognitive Psychology}, 24\penalty0 (2):\penalty0 220--251,
  1992.
\newblock \doi{10.1016/0010-0285(92)90008-P}.

\bibitem[Xu and Spelke(2000)]{xu2000large}
Fei Xu and Elizabeth~S. Spelke.
\newblock Large number discrimination in 6-month-old infants.
\newblock \emph{Cognition}, 74\penalty0 (1):\penalty0 B1--B11, 2000.
\newblock \doi{10.1016/S0010-0277(99)00066-9}.

\bibitem[Yang et~al.(2023)Yang, Zhang, Li, Zou, Li, and Gao]{yang2023som}
Jianwei Yang, Hao Zhang, Feng Li, Xueyan Zou, Chunyuan Li, and Jianfeng Gao.
\newblock Set-of-mark prompting unleashes extraordinary visual grounding in
  {GPT-4V}.
\newblock \emph{arXiv preprint arXiv:2310.11441}, 2023.

\bibitem[Zhao et~al.(2025)Zhao, Zhang, Lin, Li, Wu, Zhang, and
  Wei]{zhao2025pyvision}
Shitian Zhao, Haoquan Zhang, Shaoheng Lin, Ming Li, Qilong Wu, Kaipeng Zhang,
  and Chen Wei.
\newblock {PyVision}: Agentic vision with dynamic tooling.
\newblock \emph{arXiv preprint arXiv:2507.07998}, 2025.

\bibitem[Zheng et~al.(2025)Zheng, Yang, Hong, Zhao, Xu, Yang, Shen, and
  Yu]{zheng2025deepeyes}
Ziwei Zheng, Michael Yang, Jack Hong, Chenxiao Zhao, Guohai Xu, Le~Yang, Chao
  Shen, and Xing Yu.
\newblock {DeepEyes}: Incentivizing ``thinking with images'' via reinforcement
  learning.
\newblock \emph{arXiv preprint arXiv:2505.14362}, 2025.

\end{thebibliography}

\newpage
\appendix
\section{Developmental predictions}
\label{app:predictions}
\Cref{tab:predictions} lists the predictions the developmental account makes for any counter, procedural or not, how the protocol of \cref{sec:protocol} operationalises each, and the verdict of \cref{sec:results}.
\begin{table}[h]
\centering\scriptsize
\caption{The developmental predictions our protocol operationalises. One-to-one correspondence and cardinality are audited directly; abstraction, arrangement, and order effects are behavioural comparisons; stable order is supplied by the harness (\cref{tab:tools}), so it is scaffolded, not tested.}
\label{tab:predictions}
\setlength{\tabcolsep}{4pt}
\begin{tabular}{@{}p{4.6cm} p{4.5cm} p{3.6cm}@{}}
\toprule
developmental prediction & operationalisation & verdict \\
\midrule
small sets are exact without any routine \citep{trick1994small} & accuracy vs.\ $n$: ceiling for $n\le4$ & supported (\cref{sec:res-cliff}) \\
pointing supports one-to-one tracking \citep{alibali1999function} & \arm{tag} vs.\ \arm{direct}, conditioned on pointing precision & precise pointers only (\cref{sec:res-audit}) \\
keeping-track errors are skips, rarely double counts \citep{fuson1988children} & omission/duplicate/phantom audit & supported (\cref{sec:res-audit}) \\
abstraction: anything can be counted \citep{gelman1978child} & mixed shapes and colours; red discs among blue & hurts estimation more than the routine (\cref{sec:res-dev}) \\
regular arrays are easier than scattered ones \citep{potter1968spatial} & regular/random/clustered; rotated, hexagonal, holed lattices & only under the routine (\cref{sec:res-dev}) \\
cardinality: the last tag names the set size \citep{gelman1978child} & announced answer $=$ number of tags? & model-dependent (\cref{sec:res-audit}) \\
order irrelevance: any traversal gives one count \citep{gelman1978child} & forced opposite traversal orders, with repeat and rotation controls; mirror flip & often violated (\cref{sec:res-dev}) \\
\bottomrule
\end{tabular}
\end{table}

\section{Prompts}
\label{app:prompts}
All model-facing text, verbatim. Placeholders in braces are filled per item (\texttt{\{obj\}} is the plural object name from the dataset, e.g.\ ``apples'', \texttt{\{obj\_sing\}} its singular, \texttt{\{T\}} the tool budget). Coordinates are always requested in the normalised $[0,1000]^2$ frame.
\paragraph{Direct}
\begin{small}\begin{verbatim}
How many {obj} are in this image? Reply with only one line of the form
'ANSWER: <integer>'.
\end{verbatim}\end{small}
\paragraph{Chain of thought}
\begin{small}\begin{verbatim}
Count the {obj} in this image carefully. Think step by step (for example,
describe where each one is or count region by region), then finish with a
final line of the form 'ANSWER: <integer>'.
\end{verbatim}\end{small}
\paragraph{List (blind pointing)}
\begin{small}\begin{verbatim}
Locate every {obj_sing} in this image. Coordinates use a normalized frame: x
and y are integers in [0,1000], where (0,0) is the top-left and (1000,1000)
the bottom-right corner. Output a JSON object on one line: {"points": [[x,y],
[x,y], ...]} with exactly one point per {obj_sing}, placed at its center. Then
finish with a final line of the form 'ANSWER: <integer>' giving the number of
{obj}.
\end{verbatim}\end{small}
\paragraph{Tool-loop system prompt (\{tool\_lines\} is filled with the lines below for the arm's tools)}
\begin{small}\begin{verbatim}
You are counting {obj} in an image. Coordinates use a NORMALIZED frame: x and
y are integers in [0,1000]; (0,0) is the top-left corner and (1000,1000) the
bottom-right corner of the ORIGINAL image; enlarged crops keep this same
frame.
In every reply, think briefly and then output EXACTLY ONE JSON action as the
last line (no code fences). Available actions:
{tool_lines}
Procedure: tag the {obj} you see; inspect the returned image; if a tag is off
any {obj_sing} or two tags share one {obj_sing}, say so and do not count
those; tag ONLY the {obj} that are still untagged; then answer. You have at
most {T} tool calls before you must answer.
\end{verbatim}\end{small}
\paragraph{Tool lines}
\begin{small}\begin{verbatim}
 {"tool":"tag","points":[[x,y],...]}   draw numbered tags (numbering continues
from the last tag) at these points on the working image; you will see the
result. Tag each {obj_sing} exactly once, at its center. Never re-tag an
object that already has a tag. To start over, add "mode":"replace" (all
previous tags are removed and numbering restarts at 1).
 {"tool":"crop","box":[x1,y1,x2,y2]}   see an enlarged view of this region of
the working image (existing tags stay visible; coordinates stay in the full-
image frame).
 {"tool":"erase","points":[[x,y],...]} grey out already-counted objects at
these points so they cannot be counted twice.
 {"tool":"grid","on":true}              overlay a light coordinate grid with
0-1000 tick labels to help you place tags precisely.
 {"tool":"answer","count":N}           final answer: N = number of distinct
{obj} in the image (normally the number of correctly placed tags).
\end{verbatim}\end{small}
\paragraph{First user message}
\begin{small}\begin{verbatim}
Count the {obj} in this image.
\end{verbatim}\end{small}
\paragraph{Traversal-order suffixes (order experiment; appended to the first user message)}
\begin{small}\begin{verbatim}
 Tag the {obj} strictly in reading order: start at the top-left, sweep left to
right, then move down and sweep the next band left to right, until you reach
the bottom-right.
 Tag the {obj} strictly in reverse reading order: start at the bottom-right,
sweep right to left, then move up and sweep the next band right to left, until
you reach the top-left.
\end{verbatim}\end{small}
\paragraph{One-tag-per-call suffix (tag-1 arm; appended to the first user message)}
\begin{small}\begin{verbatim}
 Tag exactly ONE {obj_sing} per tag call: output a single [x,y] point, look at
the returned image, then tag the next untagged {obj_sing} with the next call,
until every {obj_sing} is tagged; then answer.
\end{verbatim}\end{small}
\paragraph{Verification nudge after a tag call}
\begin{small}\begin{verbatim}
Tags {a}..{b} are now drawn (red dot = exact point, numbered box beside it).
Check the image: (a) is each tag on a distinct {obj_sing}? (b) is any
{obj_sing} still untagged? (c) do two tags share one {obj_sing}? Then tag ONLY
the missing ones, use another tool, or answer with the number of distinct
correctly tagged {obj}. Tool calls left: {left}.
\end{verbatim}\end{small}
\paragraph{Plain nudge (ablation without verification)}
\begin{small}\begin{verbatim}
Tags {a}..{b} are now drawn. Continue. Tool calls left: {left}.
\end{verbatim}\end{small}
\paragraph{Observation after crop / erase / grid}
\begin{small}\begin{verbatim}
Enlarged view of the requested region (coordinates still refer to the full
image). Continue. Tool calls left: {left}.
The requested points are greyed out on the working image. Continue. Tool calls
left: {left}.
Grid overlay enabled on the working image. Continue. Tool calls left: {left}.
\end{verbatim}\end{small}
\paragraph{Invalid action}
\begin{small}\begin{verbatim}
Your last line was not a valid JSON action. Reply with exactly one JSON action
as described. Tool calls left: {left}.
\end{verbatim}\end{small}
\paragraph{Forced answer}
\begin{small}\begin{verbatim}
You have used all tool calls. Reply now with {"tool":"answer","count":N} only.
\end{verbatim}\end{small}
\paragraph{Tile prompt (scripted tiles arm)}
\begin{small}\begin{verbatim}
This is an enlarged crop of a larger image; it may contain zero {obj}. First
state how many {obj} have their center inside THIS crop (it is common for a
crop to contain none). Then output one line {"points": [[x,y], ...]} with
exactly one point per {obj_sing} at its center, using a normalized frame of
THIS crop (x and y integers in [0,1000], (0,0) top-left, (1000,1000) bottom-
right); output {"points": []} if there are none. Finish with a final line
'ANSWER: <integer>'.
\end{verbatim}\end{small}

\section{Tools and rendering}
\label{app:tools}
\begin{table}[h]
\centering\scriptsize
\caption{Positioning relative to prior mark-based methods (cells summarise the published descriptions; ``grounding'' = pointing accuracy is analysed, but wrong counts are not decomposed into principle violations). Our distinction is not that a VLM can make and inspect marks, it is that the marks are used as a measurement trace: audited point-by-point against ground truth and read through the developmental principles of counting.}
\label{tab:relwork}
\setlength{\tabcolsep}{3pt}
\begin{tabular}{@{}l l l l l@{}}
\toprule
method & marks come from & vision modules in loop & sees \& revises marks & audit of the marks \\
\midrule
Set-of-Mark \citep{yang2023som} & segmenter & segmenter & no & none \\
Autonomous Imag.\ \citep{liu2024autonomous} & model cursor & SAM2, inpainting & yes & none \\
SketchVLM \citep{collins2026sketchvlm} & model coords & none (rendering) & yes & none \\
Point-then-count \citep{alghisi2026getting} & model coords & none (unrendered) & no & grounding \\
LVLM-Count \citep{fetrat2024lvlmcount} & detector & detector & no & none \\
ABACUS \citep{mondal2026abacus} & trained model & trained counter & no & none \\
CountAgent (ours) & model coords & none (PIL only) & yes & point-GT principles \\
\bottomrule
\end{tabular}
\end{table}
\begin{table}[h]
\centering\small
\caption{Epistemic tools, the child strategy each mirrors, and the counting principle each supports. All tools are PIL operations with no visual intelligence. Coordinates are in a normalised $[0,1000]^2$ frame of the original image. The harness itself supplies the stable-order component (tags are numbered consecutively across calls), so stable order is scaffolded, not independently tested.}
\label{tab:tools}
\setlength{\tabcolsep}{4pt}
\begin{tabular}{@{}l p{3.9cm} p{3.7cm} p{3.4cm}@{}}
\toprule
tool & signature & child strategy & principle(s) \\
\midrule
\tool{tag} & points $\to$ numbered tags on the working image & pointing while reciting ``1, 2, 3'' & one-to-one, stable order (scaffolded), cardinality \\
\tool{crop} & box $\to$ enlarged view (tags visible) & leaning in; partitioning & one-to-one in dense regions \\
\tool{erase} & points $\to$ greyed-out items & moving counted items aside & one-to-one (no double count) \\
\tool{grid} & on $\to$ coordinate grid overlay & organised row-by-row scanning & pointing precision; partition \\
\tool{tiles} (scripted) & $k{\times}k$ overlapping tiles, per-tile points, merge & counting by groups & cardinality by addition \\
\tool{answer} & $N$ & announcing the last number word & cardinality \\
\bottomrule
\end{tabular}
\end{table}
\paragraph{Working state.} The harness keeps the original image, an ordered list of tags (pixel coordinates), a list of erased regions, and a grid flag; the working image is re-rendered from this state after every action, so any intermediate view can be reproduced exactly from the stored trajectory (this is how \cref{fig:hero} was produced).
\paragraph{Tag rendering.} Each tag is a filled disc of radius $\max(3, \min(6, 0.006\,s))$ px, where $s$ is the shorter image side, drawn at the exact requested point with a white outline, plus a white box with a red border containing the tag number in a sans-serif font of size $\max(11, \min(20, 0.024\,s))$. The box is placed at one of four offsets (right, left, above, below) chosen greedily to minimise overlap with previously drawn boxes. Plain-mark ablations omit the box.
\paragraph{Crop.} The requested normalised box is cut from the \emph{working} image (tags visible) and upscaled so that the shorter side is at least 768\,px (longer side at most 1024\,px). Coordinates in the model's subsequent replies remain in the full-image frame.
\paragraph{Erase.} A grey ($\mathrm{RGB}=160$, $\alpha=0.9$) disc of radius $0.02\,s$ (or a grey rectangle for a box) is composited under the tags.
\paragraph{Grid.} Lines every 100 normalised units in both directions with tick labels, drawn in translucent blue.
\paragraph{Caps and salvage.} At most 100 points per tag call and 300 tags per trajectory are drawn; points outside $[-5,1005]$ are dropped (degenerate replies often continue an arithmetic sequence off the image); truncated JSON is salvaged by extracting $[x,y]$ pairs; a tag call may set \texttt{"mode":"replace"} to clear all tags and start over. Unparseable replies are answered with a request to re-issue a valid action, at most three times.
\paragraph{Context policy.} Only the two most recent images stay in the request; earlier image parts are replaced by the text ``[image shown earlier omitted]''. Every request asks the provider for token usage and cost, which are logged per call.
\paragraph{Coordinate frame.} In a pilot with a pixel frame, Gemini 3.1 Flash-Lite returned coordinates in its native $[0,1000]$ frame; on a $768$-px image every tag landed roughly $0.77\times$ too close to the origin, and the model re-issued the same wrong points four times without noticing. All reported experiments therefore use the normalised frame for every model.

\section{Datasets}
\label{app:data}
\paragraph{DotArrays.} $768{\times}768$ white canvases. Main battery: black discs of radius 14\,px, $n\in\{1,\dots,10,12,15,20,25,30,40,50,60\}$, arrangements \emph{regular} (near-square lattice with $\pm15\%$ jitter), \emph{random} (uniform with a minimum gap of 8\,px between discs), and \emph{clustered} (2--4 Gaussian clusters, $\sigma=0.09$ of the side), four seeds each ($18\times3\times4=216$ images). Factor sets (four seeds each): \emph{heterogeneous} (circles, squares, triangles in six colours; question ``shapes''; $n\in\{8,15,25,40\}$), \emph{distractor} (count red discs among an equal number of blue discs; $n\in\{6,12,20,30\}$), \emph{dense} (clustered discs allowed to touch; $n\in\{15,30,50\}$), \emph{small} (radius 6\,px; $n\in\{20,40,60\}$). Total \nDot{} images with exact centre coordinates.
\paragraph{DotArrays lattice controls.} Sixty further images ($n\in\{12,20,30,40,50\}$, four seeds each) on the main canvas with the main disc radius and jitter: a square lattice rotated by an angle drawn from $[20^\circ,40^\circ]$ about the canvas centre, a hexagonal lattice, and a square lattice with $\lceil n/0.75\rceil$ cells of which $n$ are filled. Ideal cell centres are stored with the points.
\paragraph{DotArrays XL.} The extension battery holds \nDotXL{} images with $n\in\{80,100,120,150,200\}$, arrangements \emph{regular} and \emph{random} (the clustered generator does not reach these densities at its main-battery dispersion), three seeds each, same canvas, disc radius, and rendering as the main battery. It exists to resolve the right-censoring of $R_{80}$ for the two models that stay above 80\,\% at $n{=}60$ (Gemini 3.7 Flash and Gemini 3.1 Pro, which ran \arm{direct}, \arm{cot}, \arm{tag}, and \arm{full} on all of it).
\paragraph{FSC-147.} From the official test split \citep{ranjan2021learning} we sample 40 images per count bin $\{8$--$15, 16$--$25, 26$--$40, 41$--$60, 61$--$80\}$ with round-robin class balancing (29 classes; the class name is used as the object name, lightly normalised, e.g.\ ``deers'' $\to$ ``deer''). Images (native height 384\,px) are upscaled to height 768\,px and the dot annotations scaled accordingly.
\paragraph{Image size.} Images are resized to at most 1024\,px on the long side before any arm sees them.
\paragraph{PixMo-Count.} From the human-verified test split \citep{deitke2024molmo} (540 images, counts 2--10) we sample 20 images per count and fetch the images from their Flickr URLs (dead links are skipped); the dataset provides no points, so PixMo-Count contributes to count metrics only.

\section{Audit details}
\label{app:audit}
\paragraph{Matching.} For each image with point ground truth we form the distance matrix between the model's tags (pixel coordinates of the exact tag points) and the ground-truth points, and solve a minimum-cost one-to-one assignment (Hungarian algorithm) in which pairs farther apart than a per-image threshold $\tau$ are forbidden. $\tau$ is half the median nearest-neighbour distance among the ground-truth points of that image, floored at 15\,px; the median $\tau$ is 44\,px on DotArrays (range 17--272, wide because sparse arrays have large nearest-neighbour distances) and 41\,px on FSC-147 (range 15--138). A matched tag is \emph{on-target}; an unmatched tag within $\tau$ of some (necessarily already-matched) object is a \emph{duplicate}; an unmatched tag farther than $\tau$ from every object is a \emph{phantom}; a ground-truth point left unmatched is an \emph{omission}. Scaling $\tau$ inside an image cannot create matches beyond one per object, so duplicates and omissions are conservative with respect to $\tau$; the phantom/duplicate boundary is the sensitive one.
\paragraph{Sensitivity.} \Cref{tab:tau} recomputes the audit for the \arm{tag} arm with $\tau$ scaled by $0.75$ and $1.5$. On-target rates move by a few points and some phantoms become duplicates as $\tau$ grows, but the ordering of models and the qualitative conclusions (precise vs.\ imprecise pointers; omissions dominating dense scenes) are unchanged at both scales.
\paragraph{Known biases.} FSC-147 dots mark approximate object centres placed by annotators; for elongated or heavily occluded objects a semantically correct tag can fall outside $\tau$ and be scored as a phantom plus an omission, which slightly inflates both categories on FSC-147 relative to DotArrays, where centres are exact by construction. Cardinality consistency compares the announced count with the number of drawn tags and is independent of $\tau$.
\begin{table}[h]
\centering\scriptsize
\caption{Sensitivity of the audit to the matching threshold (tag arm, DotArrays + FSC-147): rates per 100 ground-truth objects.}
\label{tab:tau}
\begin{tabular}{@{}llrrrr@{}}
\toprule
model & $\tau$ scale & on-target & omitted & duplicate & phantom \\
\midrule
Grok 4.20 & 0.75 & 19.0 & 81.0 & 0.6 & 66.9 \\
Grok 4.5 & 0.75 & 35.5 & 64.5 & 2.4 & 125.4 \\
Gemini 3.1 Pro & 0.75 & 73.1 & 26.9 & 0.4 & 28.2 \\
Claude Sonnet 5 & 0.75 & 14.1 & 85.9 & 3.2 & 68.1 \\
Qwen3-VL-8B & 0.75 & 46.4 & 53.6 & 6.6 & 56.3 \\
Gemini 3.7 Flash & 0.75 & 82.5 & 17.5 & 0.6 & 16.3 \\
GPT-5 mini & 0.75 & 13.4 & 86.6 & 0.5 & 40.7 \\
Gemini 3.1 Flash-Lite & 0.75 & 66.5 & 33.5 & 2.4 & 28.6 \\
Qwen3-VL-32B & 0.75 & 62.1 & 37.9 & 19.7 & 102.9 \\
Grok 4.20 & 1.0 & 30.9 & 69.1 & 2.9 & 52.7 \\
Grok 4.5 & 1.0 & 52.6 & 47.4 & 10.2 & 100.4 \\
Gemini 3.1 Pro & 1.0 & 77.9 & 22.1 & 1.4 & 22.4 \\
Claude Sonnet 5 & 1.0 & 22.0 & 78.0 & 8.1 & 55.2 \\
Qwen3-VL-8B & 1.0 & 53.4 & 46.6 & 10.6 & 45.3 \\
Gemini 3.7 Flash & 1.0 & 85.5 & 14.5 & 1.3 & 12.6 \\
GPT-5 mini & 1.0 & 20.6 & 79.4 & 2.2 & 31.9 \\
Gemini 3.1 Flash-Lite & 1.0 & 72.4 & 27.6 & 3.9 & 21.2 \\
Qwen3-VL-32B & 1.0 & 69.2 & 30.8 & 29.7 & 85.7 \\
Grok 4.20 & 1.5 & 46.7 & 53.3 & 8.3 & 31.4 \\
Grok 4.5 & 1.5 & 73.3 & 26.7 & 32.9 & 57.1 \\
Gemini 3.1 Pro & 1.5 & 83.6 & 16.4 & 4.4 & 13.7 \\
Claude Sonnet 5 & 1.5 & 33.3 & 66.7 & 18.6 & 33.5 \\
Qwen3-VL-8B & 1.5 & 61.6 & 38.4 & 18.9 & 28.8 \\
Gemini 3.7 Flash & 1.5 & 89.0 & 11.0 & 3.0 & 7.3 \\
GPT-5 mini & 1.5 & 29.9 & 70.1 & 7.2 & 17.6 \\
Gemini 3.1 Flash-Lite & 1.5 & 78.6 & 21.4 & 7.7 & 11.3 \\
Qwen3-VL-32B & 1.5 & 76.3 & 23.7 & 49.0 & 59.4 \\
\bottomrule
\end{tabular}
\end{table}

\section{Additional results}
\label{app:more}
\begin{figure}[h]
\centering
\includegraphics[width=\linewidth]{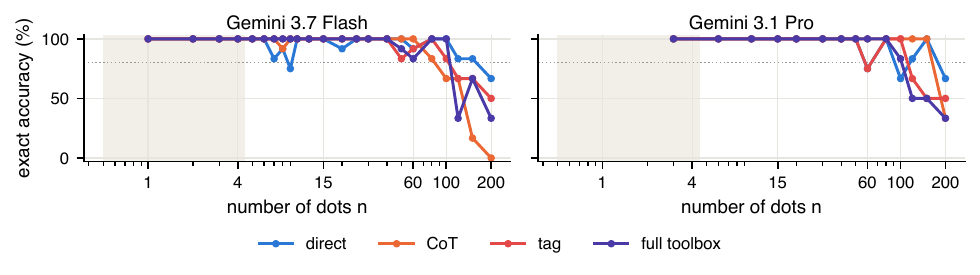}
\caption{Exact accuracy versus $n$ for the two models whose $R_{80}$ is right-censored on the main battery ($n\le60$), continued onto the XL battery ($n$ up to 200, \cref{app:data}); the shaded band is the subitizing range, the dotted line the 80\,\% criterion. Both cliffs appear near 150; the verbal routine (CoT) collapses before direct estimation does.}
\label{fig:xl}
\end{figure}

\begin{figure}[h]
\centering
\includegraphics[width=\linewidth]{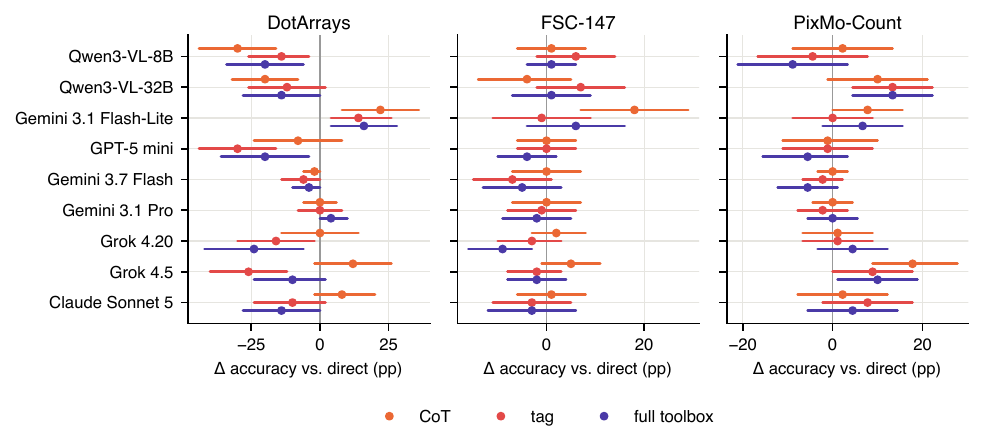}
\caption{Paired change in exact accuracy of each arm over \arm{direct} on the common subset, with 95\,\% paired bootstrap CIs over images ($B{=}10^4$; numbers in \cref{tab:delta}). With 50--100 images per cell most single effects are within noise; CI-solid are Flash-Lite's gains and the imprecise pointers' losses on DotArrays, Qwen3-VL-32B's gain on PixMo-Count, and CoT for Flash-Lite.}
\label{fig:delta}
\end{figure}

\begin{figure}[h]
\centering
\includegraphics[width=\linewidth]{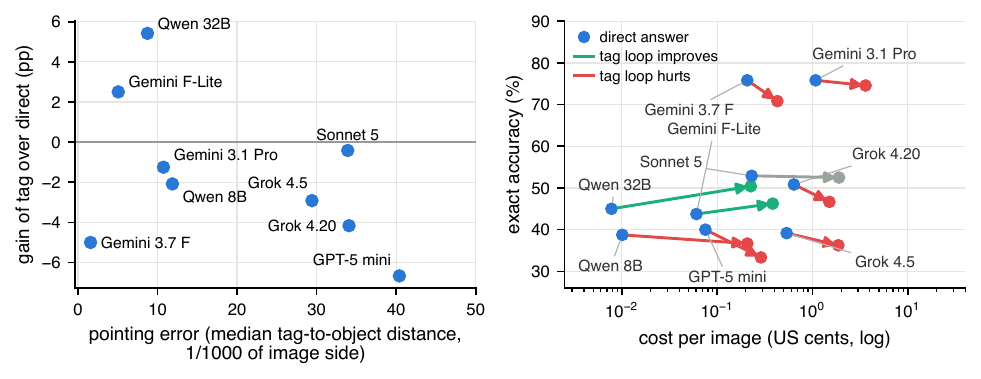}
\caption{Left: gain of \arm{tag} over \arm{direct} on the common subset against pointing error (median tag-to-object distance of on-target tags, DotArrays + FSC-147); the models that gain are the models that point. Right: the same comparison in cost--accuracy space, each arrow running from a model's \arm{direct} answer (blue dot) to its \arm{tag} loop (green improves, red hurts, grey within one point); the loop costs roughly $3\times$ more per image.}
\label{fig:precision}
\end{figure}

\begin{figure}[h]
\centering
\begin{subfigure}[t]{0.42\linewidth}\centering\includegraphics[width=\linewidth]{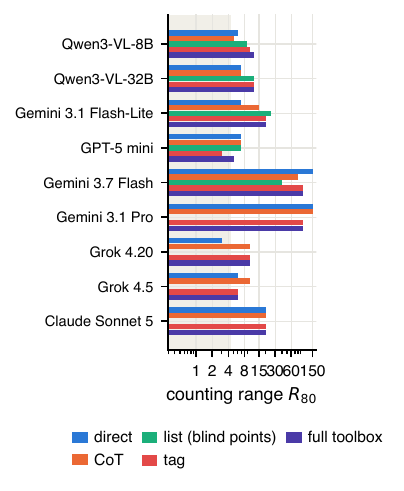}\caption{Counting range $R_{80}$ per model and arm.}\label{fig:r80}\end{subfigure}\hfill
\begin{subfigure}[t]{0.56\linewidth}\centering\includegraphics[width=\linewidth]{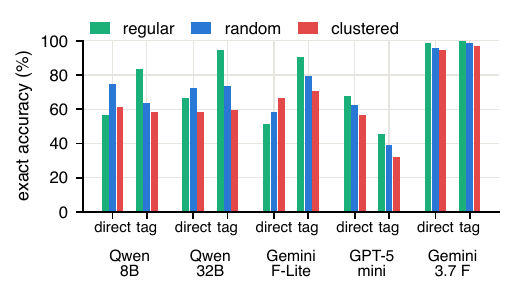}\caption{Arrangement effect on the DotArrays main battery.}\label{fig:arrangement}\end{subfigure}
\caption{Developmental signatures. (a) The largest $n$ up to which the isotonic fit of exact accuracy stays $\ge 80\,\%$ (shaded: subitizing range; tier-3 models were run on a subset of $n$; the two Geminis include the XL extension, and $\ge$ marks ranges still censored at the largest tested $n$; CIs in \cref{tab:reighty}). (b) Exact accuracy on regular, random, and clustered arrays for the models that saw all three arrangements.}
\label{fig:dev}
\end{figure}

\begin{figure}[h]
\centering
\includegraphics[width=\linewidth]{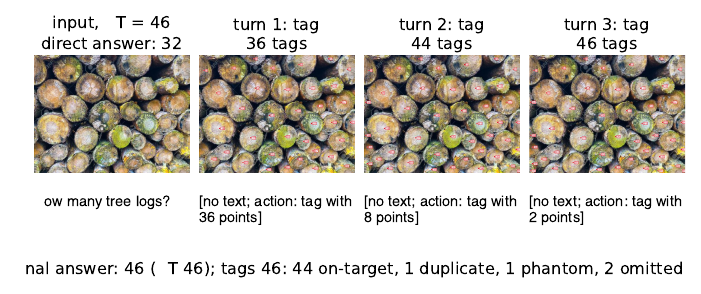}
\caption{A real trajectory, re-rendered from the stored actions: Gemini 3.1 Pro on an FSC-147 photograph with 46 tree logs. The \arm{direct} answer is 32; three rounds of tagging reach 46, and the audit of the final tags against the dot annotations finds 44 on-target tags, one duplicate, one phantom, and two omissions. Panels show the working image after each tool call.}
\label{fig:hero}
\end{figure}

\begin{figure}[h]\centering\includegraphics[width=\linewidth]{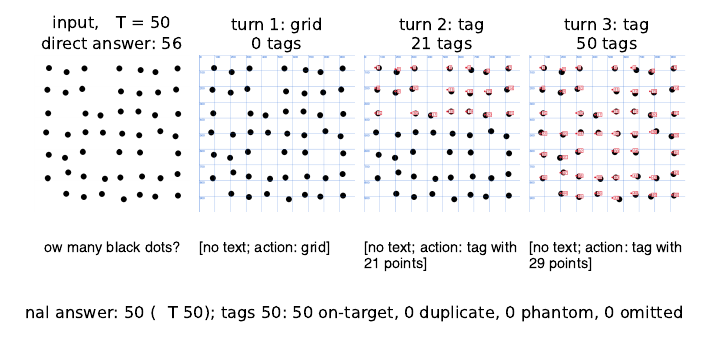}
\caption{Second real trajectory: Gemini 3.1 Flash-Lite with the full toolbox on a 50-dot regular array. \arm{direct} says 56; the model calls \tool{grid}, tags 21 dots, then the remaining 29, and answers 50 with every tag on target.}\label{fig:hero2}\end{figure}

\begin{figure}[h]\centering\includegraphics[width=\linewidth]{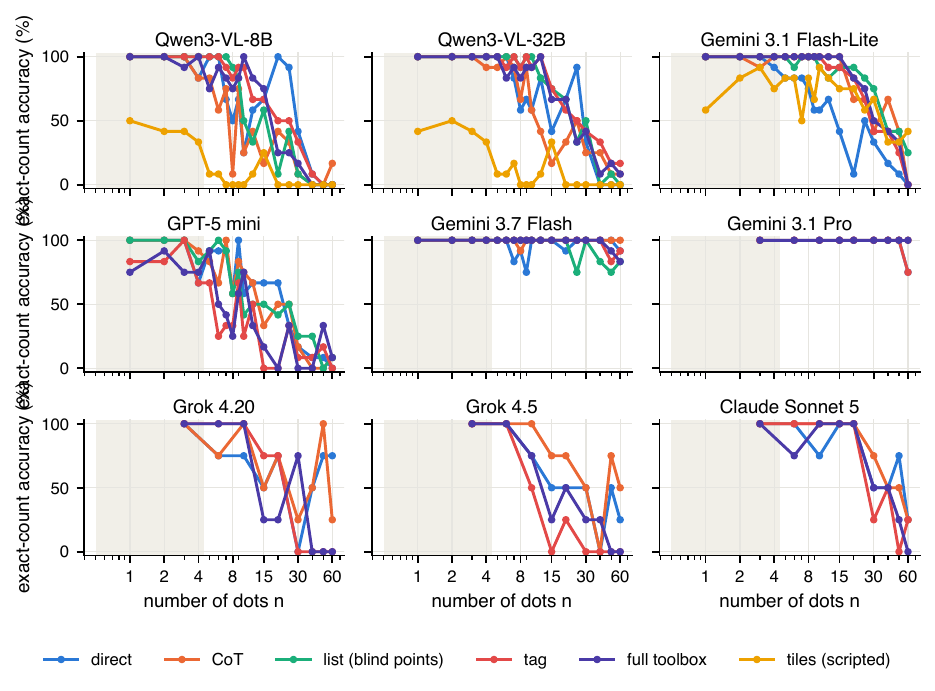}
\caption{Exact-count accuracy versus number of dots on the DotArrays main battery (three arrangements, four seeds per $n$), one panel per model, one line per harness arm; the shaded band marks the subitizing range $n\le4$. Tier-2 models ran fewer arms; tier-3 models ran a subset of $n$ (two arrangements, two seeds per level).}\label{fig:curvesall}\end{figure}

\begin{figure}[h]\centering\includegraphics[width=0.9\linewidth]{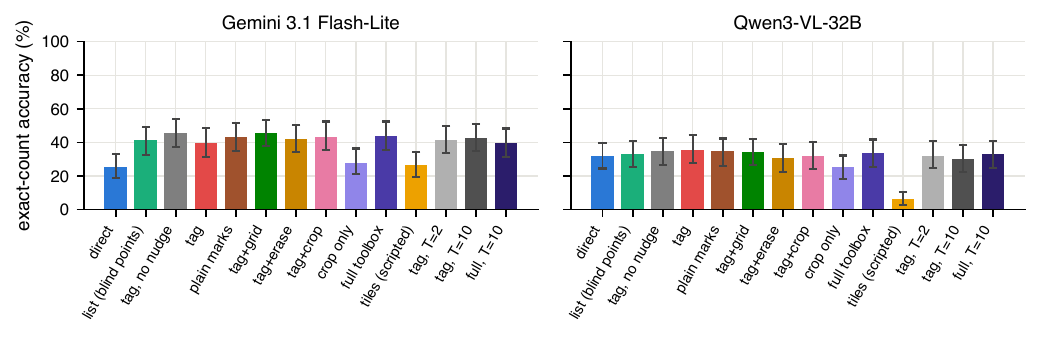}
\caption{Ablations: exact accuracy on the ablation subset with bootstrap 95\,\% CIs.}\label{fig:ablation}\end{figure}

This section collects the material referenced from \cref{sec:results}: the developmental-signature panels (\cref{fig:dev}), the XL extension curves (\cref{fig:xl}), the gain-versus-pointing-error and cost panels (\cref{fig:precision}), the bootstrap table for $R_{80}$ (\cref{tab:reighty}), the paired-delta table behind \cref{fig:delta} (\cref{tab:delta}), the traversal-order experiment with its repeat and rotation controls (\cref{tab:order}), the strict-seriality comparison (\cref{tab:tagone}), two full trajectories (\cref{fig:hero,fig:hero2}), the accuracy-versus-$n$ curves for all nine models (\cref{fig:curvesall}), the principle audit for all point-producing arms including \arm{list} and \arm{tiles} (\cref{fig:auditfull}, with the full numbers in \cref{tab:audit}), the ablation bars (\cref{fig:ablation}), and the long-form tables: all seven arms on the common subset with MAE (\cref{tab:maincommonfull}), the main comparison on the full image sets rather than the common subset (\cref{tab:mainfull}), all ablation arms with their audits (\cref{tab:ablation}), FSC-147 by count bin (\cref{tab:fscbins}), PixMo-Count by ground-truth count (\cref{tab:pixmo}), the DotArrays factor sets (\cref{tab:family}), and the mirror-consistency table (\cref{tab:flip}). The controls added in revision (coupling of tags and answers, the tag-perturbation probe, hidden reasoning and the thinking-off control, the lattice and jitter checks, and the tag-versus-list contrasts) are in \cref{app:controls}. Three regularities are worth noting beyond the main text. On the \emph{full} DotArrays battery (\cref{tab:mainfull}) tagging raises Qwen3-VL-32B's exact accuracy (\accQwenThirtyTwoDirectDot{} to \accQwenThirtyTwoTagDot{}\,\%) while multiplying its MAE by ten (\maeQwenThirtyTwoDirectDot{} to \maeQwenThirtyTwoTagDot{}) through occasional runaway coordinate sequences, so accuracy, MAE, and the audit must be read together. Second, in \cref{tab:fscbins} every model's accuracy decays with the count bin under every arm; the tools change \emph{where} the decay starts for at most one model but never remove it. Third, in \cref{tab:pixmo} the per-count profile of \arm{direct} answers is not monotone for several models (dips at 6--7 with recovery at 8--10), a pattern consistent with counting-by-configuration rather than enumeration on small photographed sets.

\paragraph{Traversal-order experiment.} The two order arms are the standard \arm{tag} loop with one sentence appended to the first user message requesting a left-to-right/top-to-bottom or a right-to-left/bottom-to-top traversal (verbatim in \cref{app:prompts}); Gemini 3.1 Flash-Lite, Qwen3-VL-32B, and Gemini 3.7 Flash ran both arms on the \nOrderImages{} DotArrays images of the ablation subset. Because tags are numbered in the order they are drawn, the realised traversal is observable: $\bar\tau$ in \cref{tab:order} is the Kendall correlation between tag index and vertical position, averaged over trajectories with at least five tags. Order irrelevance predicts identical final counts under the two traversals; agreement below the model's own exact accuracy on these arrays means the traversal, which should be irrelevant, changes the answer. Because hosted models are not perfectly deterministic even at temperature 0, each order arm was run twice: the repeat control in \cref{tab:order} measures how often the \emph{same} order repeated yields a different count (averaged over both orders), the cross agreement averages over all four pairs of opposite-order runs, and the order effect is their difference, so the three columns are consistent by construction. The rotation control runs the reading-order arm on every array rotated by $180^\circ$ (twice; Qwen3-VL-8B replaces the withdrawn Qwen3-VL-32B), which visits the dots in the reverse sequence under the natural instruction; its agreement is averaged over the four pairs of (unrotated, rotated) reading-order runs and its effect is again the excess over the repeat control.
\paragraph{Strict seriality (tag-1).} The batched \tool{tag} call can place many points before the model sees any of them rendered, so the standard trace is an ordered externalised enumeration, not proof of item-by-item inspection. The \arm{tag-1} arm closes that gap: each call may add exactly one point (extra points are discarded with a notice), so every mark is placed only after the previous one is observed. \Cref{tab:tagone} compares \arm{direct}, batched \arm{tag}, and \arm{tag-1} for the two precise pointers on the DotArrays ablation images.
\begin{table}[h]
\centering\scriptsize
\caption{Counting range $R_{80}$ per model and arm on the DotArrays main battery (for Gemini 3.7 Flash and Gemini 3.1 Pro including the XL extension to $n{=}200$), with 95\,\% bootstrap confidence intervals (images resampled within each level, $B{=}10^3$). $\ge$ marks right-censored values: the isotonic accuracy fit is still $\ge 80\,\%$ at the largest tested $n$.}
\label{tab:reighty}
\begin{tabular}{@{}lrrrr@{}}
\toprule
model & direct & CoT & tag & full \\
\midrule
Qwen3-VL-8B & 6 {\tiny[6,7]} & 5 {\tiny[3,7]} & 10 {\tiny[7,15]} & 12 {\tiny[4,15]} \\
Qwen3-VL-32B & 7 {\tiny[6,9]} & 7 {\tiny[6,10]} & 12 {\tiny[12,15]} & 12 {\tiny[12,20]} \\
Gemini 3.1 Flash-Lite & 7 {\tiny[4,9]} & 15 {\tiny[15,25]} & 20 {\tiny[12,25]} & 20 {\tiny[15,25]} \\
GPT-5 mini & 7 {\tiny[3,12]} & 7 {\tiny[4,12]} & 3 {\tiny[1,5]} & 5 {\tiny[0,5]} \\
Gemini 3.7 Flash & 150 {\tiny[100,200]} & 80 {\tiny[60,120]} & 100 {\tiny[50,150]} & 100 {\tiny[50,100]} \\
Gemini 3.1 Pro & 150 {\tiny[50,200]} & 150 {\tiny[150,200]} & 100 {\tiny[100,150]} & 100 {\tiny[80,120]} \\
Grok 4.20 & 3 {\tiny[3,20]} & 10 {\tiny[3,20]} & 10 {\tiny[10,20]} & 10 {\tiny[10,10]} \\
Grok 4.5 & 6 {\tiny[6,15]} & 10 {\tiny[10,30]} & 6 {\tiny[6,10]} & 6 {\tiny[6,10]} \\
Claude Sonnet 5 & 20 {\tiny[6,50]} & 20 {\tiny[20,40]} & 20 {\tiny[20,20]} & 20 {\tiny[20,40]} \\
\bottomrule
\end{tabular}
\end{table}

\begin{table}[h]
\centering\scriptsize
\setlength{\tabcolsep}{2pt}
\caption{Paired change in exact accuracy (percentage points) of each arm over \arm{direct} on the common subset, with 95\,\% paired bootstrap confidence intervals over images ($B{=}10^4$). Bold: the interval excludes zero.}
\label{tab:delta}
\begin{tabular}{@{}lrrrrrrrrr@{}}
\toprule
& \multicolumn{3}{c}{DotArrays} & \multicolumn{3}{c}{FSC-147} & \multicolumn{3}{c}{PixMo-Count} \\
\cmidrule(lr){2-4}\cmidrule(lr){5-7}\cmidrule(lr){8-10}
model & CoT & tag & full & CoT & tag & full & CoT & tag & full \\
\midrule
Qwen3-VL-8B & \textbf{-30 {\tiny[-44,-16]}} & \textbf{-14 {\tiny[-26,-4]}} & \textbf{-20 {\tiny[-34,-6]}} & +1 {\tiny[-6,+8]} & +6 {\tiny[-2,+14]} & +1 {\tiny[-4,+6]} & +2 {\tiny[-9,+13]} & -4 {\tiny[-17,+8]} & -9 {\tiny[-21,+3]} \\
Qwen3-VL-32B & \textbf{-20 {\tiny[-32,-8]}} & -12 {\tiny[-26,+2]} & -14 {\tiny[-28,+0]} & -4 {\tiny[-14,+5]} & +7 {\tiny[-2,+16]} & +1 {\tiny[-7,+9]} & +10 {\tiny[-1,+21]} & \textbf{+13 {\tiny[+4,+22]}} & \textbf{+13 {\tiny[+4,+22]}} \\
Gemini 3.1 Flash-Lite & \textbf{+22 {\tiny[+8,+36]}} & \textbf{+14 {\tiny[+4,+26]}} & \textbf{+16 {\tiny[+4,+28]}} & \textbf{+18 {\tiny[+7,+29]}} & -1 {\tiny[-11,+9]} & +6 {\tiny[-4,+16]} & +8 {\tiny[+0,+16]} & +0 {\tiny[-9,+9]} & +7 {\tiny[-2,+16]} \\
GPT-5 mini & -8 {\tiny[-24,+8]} & \textbf{-30 {\tiny[-44,-16]}} & \textbf{-20 {\tiny[-36,-4]}} & +0 {\tiny[-6,+6]} & +0 {\tiny[-6,+6]} & -4 {\tiny[-10,+2]} & -1 {\tiny[-11,+10]} & -1 {\tiny[-11,+9]} & -6 {\tiny[-16,+3]} \\
Gemini 3.7 Flash & -2 {\tiny[-6,+0]} & -6 {\tiny[-14,+0]} & -4 {\tiny[-10,+0]} & +0 {\tiny[-7,+7]} & -7 {\tiny[-15,+1]} & -5 {\tiny[-13,+3]} & +0 {\tiny[-3,+3]} & -2 {\tiny[-7,+2]} & -6 {\tiny[-12,+1]} \\
Gemini 3.1 Pro & +0 {\tiny[-6,+6]} & +0 {\tiny[-8,+8]} & +4 {\tiny[+0,+10]} & +0 {\tiny[-7,+7]} & -1 {\tiny[-8,+6]} & -2 {\tiny[-9,+5]} & +0 {\tiny[-4,+4]} & -2 {\tiny[-8,+3]} & +0 {\tiny[-6,+6]} \\
Grok 4.20 & +0 {\tiny[-14,+14]} & \textbf{-16 {\tiny[-30,-2]}} & \textbf{-24 {\tiny[-42,-6]}} & +2 {\tiny[-3,+8]} & -3 {\tiny[-10,+3]} & \textbf{-9 {\tiny[-16,-3]}} & +1 {\tiny[-7,+9]} & +1 {\tiny[-7,+9]} & +4 {\tiny[-3,+12]} \\
Grok 4.5 & +12 {\tiny[-2,+26]} & \textbf{-26 {\tiny[-40,-12]}} & -10 {\tiny[-24,+2]} & +5 {\tiny[-1,+11]} & -2 {\tiny[-8,+3]} & -2 {\tiny[-8,+4]} & \textbf{+18 {\tiny[+9,+28]}} & +9 {\tiny[+0,+18]} & \textbf{+10 {\tiny[+1,+19]}} \\
Claude Sonnet 5 & +8 {\tiny[-2,+20]} & -10 {\tiny[-24,+2]} & -14 {\tiny[-28,+0]} & +1 {\tiny[-6,+8]} & -3 {\tiny[-11,+5]} & -3 {\tiny[-12,+6]} & +2 {\tiny[-8,+12]} & +8 {\tiny[-2,+18]} & +4 {\tiny[-6,+14]} \\
\bottomrule
\end{tabular}
\end{table}

\begin{table}[h]
\centering\scriptsize
\setlength{\tabcolsep}{2.6pt}
\caption{Traversal-order experiment (order-irrelevance principle) on the \nOrderImages{} DotArrays images of the ablation subset: the \arm{tag} loop with an instructed scan order, left-to-right/top-to-bottom (\arm{tag}$_{\rightarrow}$) versus right-to-left/bottom-to-top (\arm{tag}$_{\leftarrow}$), each run twice. $\bar\tau$ is the mean Kendall correlation between tag sequence and vertical position (positive = top-down), showing that the instructed order is realised; agree$_{\leftrightarrow}$ = same final count under opposite orders (mean over the four run pairs); agree$_{=}$ = same final count when the \emph{same} order is run twice (repeat control; mean over both orders); order effect = agree$_{=}$ $-$ agree$_{\leftrightarrow}$, the excess disagreement attributable to reversing the traversal, with a 95\,\% paired bootstrap CI over images; acc: exact accuracy under each order and under the uninstructed \arm{tag} loop on the same images. Rotation control: the reading-order arm on the image rotated by $180^\circ$ visits the dots in the reverse sequence under the natural instruction; agree$_{\mathrm{rot}}$ is its agreement with the unrotated reading-order runs and the rotation effect its excess disagreement over the repeat control.}
\label{tab:order}
\begin{tabular}{@{}lrrrrrrrrrrr@{}}
\toprule
model & $\bar\tau_{\rightarrow}$ & $\bar\tau_{\leftarrow}$ & agree$_{\leftrightarrow}$ & agree$_{=}$ & order effect & acc$_{\rightarrow}$ & acc$_{\leftarrow}$ & acc (tag) & agree$_{\mathrm{rot}}$ & rot.\ effect & acc$_{\mathrm{rot}}$ \\
\midrule
Qwen3-VL-8B & +0.56 & -0.39 & 13.2 & 64.7 & +51.5 {\tiny[+41,+62]} & 27.9 & 14.7 & 38.2 & 30.1 & +34.6 {\tiny[+24,+45]} & 36.0 \\
Qwen3-VL-32B & +0.50 & -0.47 & 24.3 & 39.7 & +15.4 {\tiny[+8,+23]} & 44.1 & 26.5 & 45.6 & -- & -- & -- \\
Gemini 3.1 Flash-Lite & +0.71 & -0.59 & 43.8 & 65.4 & +21.7 {\tiny[+14,+31]} & 60.3 & 47.1 & 54.4 & 55.9 & +9.6 {\tiny[+1,+18]} & 63.2 \\
Gemini 3.7 Flash & +0.96 & -0.86 & 86.0 & 86.8 & +0.7 {\tiny[-1,+3]} & 94.1 & 92.6 & 94.1 & 86.8 & +0.0 {\tiny[-4,+4]} & 94.9 \\
\bottomrule
\end{tabular}
\end{table}

\begin{table}[h]
\centering\scriptsize
\caption{Strict seriality (\arm{tag-1}): one point per tag call, so every mark is placed only after the previous one is seen, versus \arm{direct} and the batched \arm{tag} loop ($T{=}6$) on the same DotArrays ablation images. \arm{tag-1} runs with a budget of $T{=}80$ tool calls (the largest array has 60 dots), so no object goes untagged by construction; omissions are the model abandoning the traversal. Error rates are per 100 ground-truth objects; cost in US cents per image.}
\label{tab:tagone}
\begin{tabular}{@{}llrrrrrrr@{}}
\toprule
model & arm & acc (\%) & MAE & omit & dup & phantom & calls & cost \\
\midrule
Gemini 3.1 Flash-Lite & direct & 29.4 & 2.2 & -- & -- & -- & 1.0 & 0.05 \\
Gemini 3.1 Flash-Lite & tag & 54.4 & 5.7 & 14.6 & 3.3 & 21.6 & 2.9 & 0.67 \\
Gemini 3.1 Flash-Lite & tag-1 & 51.5 & 4.2 & 43.6 & 2.7 & 6.1 & 29.3 & 7.79 \\
Gemini 3.7 Flash & direct & 100.0 & 0.0 & -- & -- & -- & 1.0 & 0.24 \\
Gemini 3.7 Flash & tag & 94.1 & 0.1 & 2.3 & 0.5 & 4.0 & 2.8 & 0.41 \\
Gemini 3.7 Flash & tag-1 & 85.3 & 0.5 & 67.9 & 0.1 & 0.1 & 17.7 & 5.95 \\
\bottomrule
\end{tabular}
\end{table}

\begin{figure}[h]
\centering
\includegraphics[width=\linewidth]{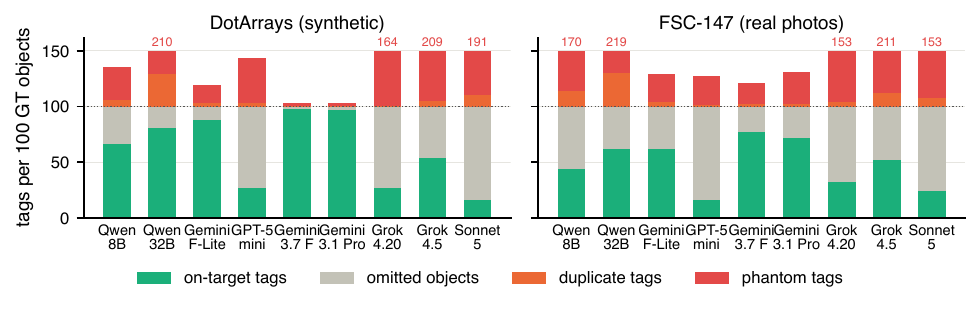}
\caption{Principle audit for the \arm{tag} arm: composition of the model's tags per 100 ground-truth objects on DotArrays and FSC-147. On-target tags plus omitted objects sum to 100; duplicates and phantoms are extra tags; numbers above a bar give totals that exceed the axis.}
\label{fig:audit}
\end{figure}
\begin{figure}[h]\centering\includegraphics[width=\linewidth]{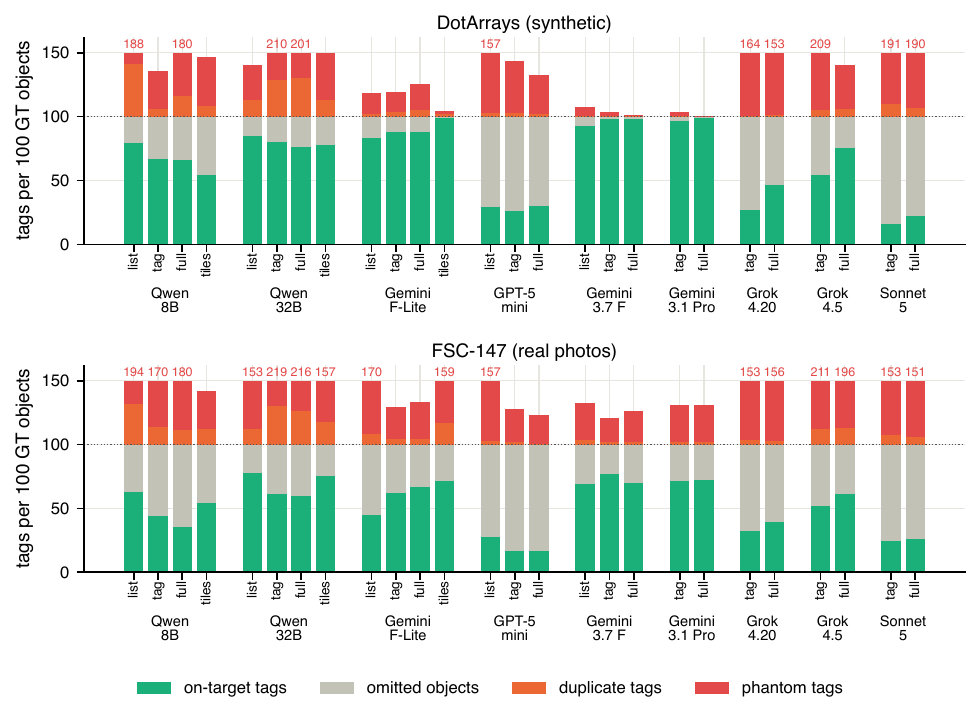}
\caption{Principle audit for all point-producing arms (\arm{list}, \arm{tag}, \arm{full}, \arm{tiles}), as \cref{fig:audit}.}\label{fig:auditfull}\end{figure}
\begin{table}[t]
\centering\scriptsize
\setlength{\tabcolsep}{3pt}
\caption{Principle audit on images with point ground truth (DotArrays + FSC-147). Per 100 ground-truth objects: on-target tags, omissions (skipped objects), duplicates (second tag on a tagged object), phantoms (tag on nothing). Card.: fraction of trajectories whose announced count equals the number of tags. Prec.: median tag-to-object distance of on-target tags in thousandths of the shorter image side (lower is better).}
\label{tab:audit}
\begin{tabular}{@{}llrrrrrr@{}}
\toprule
model & arm & on-target & omitted & duplicate & phantom & card.\,(\%) & prec. \\
\midrule
Qwen3-VL-8B & list & 69.4 & 30.6 & 35.7 & 56.3 & 76.7 & 16.2 \\
 & tag & 53.4 & 46.6 & 10.6 & 45.3 & 94.7 & 14.7 \\
 & full & 48.0 & 52.0 & 13.3 & 66.9 & 85.6 & 17.0 \\
 & tiles & 54.1 & 45.9 & 10.9 & 32.9 & 100.0 & 18.9 \\
Qwen3-VL-32B & list & 80.6 & 19.4 & 12.7 & 34.9 & 58.0 & 9.7 \\
 & tag & 69.2 & 30.8 & 29.7 & 85.7 & 80.2 & 13.1 \\
 & full & 66.3 & 33.7 & 27.9 & 81.7 & 77.4 & 12.9 \\
 & tiles & 76.1 & 23.9 & 15.7 & 38.2 & 100.0 & 18.0 \\
Gemini 3.1 Flash-Lite & list & 60.5 & 39.5 & 5.9 & 42.9 & 93.9 & 14.3 \\
 & tag & 72.4 & 27.6 & 3.9 & 21.2 & 87.7 & 9.1 \\
 & full & 75.7 & 24.3 & 5.0 & 25.3 & 92.6 & 9.9 \\
 & tiles & 82.9 & 17.1 & 10.7 & 25.6 & 100.0 & 8.5 \\
GPT-5 mini & list & 28.3 & 71.7 & 2.8 & 54.6 & 90.8 & 36.7 \\
 & tag & 20.6 & 79.4 & 2.2 & 31.9 & 64.2 & 41.0 \\
 & full & 21.9 & 78.1 & 1.3 & 25.5 & 68.0 & 37.7 \\
Gemini 3.7 Flash & list & 79.1 & 20.9 & 2.2 & 20.0 & 97.1 & 8.7 \\
 & tag & 85.5 & 14.5 & 1.3 & 12.6 & 89.1 & 7.3 \\
 & full & 81.6 & 18.4 & 1.3 & 14.3 & 88.8 & 8.1 \\
Gemini 3.1 Pro & tag & 77.9 & 22.1 & 1.4 & 22.4 & 92.0 & 8.3 \\
 & full & 79.4 & 20.6 & 1.6 & 21.5 & 90.5 & 8.0 \\
Grok 4.20 & tag & 30.9 & 69.1 & 2.9 & 52.7 & 92.0 & 32.8 \\
 & full & 41.5 & 58.5 & 2.6 & 52.8 & 94.5 & 30.5 \\
Grok 4.5 & tag & 52.6 & 47.4 & 10.2 & 100.4 & 35.0 & 29.7 \\
 & full & 65.0 & 35.0 & 11.4 & 70.1 & 73.0 & 30.1 \\
Claude Sonnet 5 & tag & 22.0 & 78.0 & 8.1 & 55.2 & 52.0 & 32.7 \\
 & full & 25.1 & 74.9 & 6.3 & 54.4 & 67.5 & 33.6 \\
\bottomrule
\end{tabular}
\end{table}

\begin{table}[h]
\centering\scriptsize
\setlength{\tabcolsep}{4.5pt}
\renewcommand{\arraystretch}{0.96}
\caption{All seven arms on the common subset (\nCommon{} images): exact-count accuracy (\%) with MAE in parentheses; best arm per row in bold. \arm{tiles} is not reported on PixMo-Count: its geometric merge only removes near-coincident points, so objects larger than the tile overlap are counted in every tile they touch (\cref{sec:res-tools}).}
\label{tab:maincommonfull}
\begin{tabular}{@{}lrrrrrrr@{}}
\toprule
model & direct & CoT & list & tag & tag+crop & full toolbox & tiles \\
\midrule\multicolumn{8}{@{}l}{\emph{DotArrays (synthetic, $n\le60$)}} \\
Qwen3-VL-8B & \textbf{52.0 {\scriptsize(5.0)}} & 22.0 {\scriptsize(71.5)} & 26.0 {\scriptsize(57.4)} & 38.0 {\scriptsize(8.3)} & 34.0 {\scriptsize(9.2)} & 32.0 {\scriptsize(17.4)} & 2.0 {\scriptsize(9.5)} \\
Qwen3-VL-32B & \textbf{48.0 {\scriptsize(2.7)}} & 28.0 {\scriptsize(5.8)} & 44.0 {\scriptsize(3.9)} & 36.0 {\scriptsize(28.8)} & 38.0 {\scriptsize(17.3)} & 34.0 {\scriptsize(24.5)} & 10.0 {\scriptsize(6.7)} \\
Gemini 3.1 Flash-Lite & 36.0 {\scriptsize(2.0)} & \textbf{58.0 {\scriptsize(4.7)}} & 54.0 {\scriptsize(3.0)} & 50.0 {\scriptsize(6.0)} & \textbf{58.0 {\scriptsize(3.6)}} & 52.0 {\scriptsize(3.3)} & 54.0 {\scriptsize(1.6)} \\
GPT-5 mini & \textbf{48.0 {\scriptsize(2.2)}} & 40.0 {\scriptsize(1.7)} & 40.0 {\scriptsize(2.3)} & 18.0 {\scriptsize(4.0)} & -- & 28.0 {\scriptsize(2.6)} & -- \\
Gemini 3.7 Flash & \textbf{100.0 {\scriptsize(0.0)}} & 98.0 {\scriptsize(0.1)} & 94.0 {\scriptsize(0.4)} & 94.0 {\scriptsize(0.2)} & -- & 96.0 {\scriptsize(0.1)} & -- \\
Gemini 3.1 Pro & 96.0 {\scriptsize(0.1)} & 96.0 {\scriptsize(0.1)} & -- & 96.0 {\scriptsize(0.0)} & -- & \textbf{100.0 {\scriptsize(0.0)}} & -- \\
Grok 4.20 & \textbf{66.0 {\scriptsize(1.1)}} & \textbf{66.0 {\scriptsize(1.7)}} & -- & 50.0 {\scriptsize(3.1)} & -- & 42.0 {\scriptsize(3.7)} & -- \\
Grok 4.5 & 54.0 {\scriptsize(1.7)} & \textbf{66.0 {\scriptsize(1.1)}} & -- & 28.0 {\scriptsize(7.1)} & -- & 44.0 {\scriptsize(3.8)} & -- \\
Claude Sonnet 5 & 72.0 {\scriptsize(1.0)} & \textbf{80.0 {\scriptsize(0.3)}} & -- & 62.0 {\scriptsize(7.1)} & -- & 58.0 {\scriptsize(6.0)} & -- \\
\midrule\multicolumn{8}{@{}l}{\emph{FSC-147 (photos, 8--80 objects)}} \\
Qwen3-VL-8B & 11.0 {\scriptsize(11.5)} & 12.0 {\scriptsize(70.4)} & 16.0 {\scriptsize(37.5)} & 17.0 {\scriptsize(22.8)} & \textbf{19.0 {\scriptsize(21.5)}} & 12.0 {\scriptsize(20.6)} & 5.0 {\scriptsize(19.4)} \\
Qwen3-VL-32B & 25.0 {\scriptsize(8.4)} & 21.0 {\scriptsize(31.1)} & 27.0 {\scriptsize(10.4)} & \textbf{32.0 {\scriptsize(23.4)}} & 22.0 {\scriptsize(17.9)} & 26.0 {\scriptsize(23.3)} & 2.0 {\scriptsize(13.9)} \\
Gemini 3.1 Flash-Lite & 23.0 {\scriptsize(7.0)} & \textbf{41.0 {\scriptsize(9.8)}} & 31.0 {\scriptsize(14.6)} & 22.0 {\scriptsize(8.3)} & 29.0 {\scriptsize(8.3)} & 29.0 {\scriptsize(7.6)} & 2.0 {\scriptsize(10.8)} \\
GPT-5 mini & 12.0 {\scriptsize(10.8)} & 12.0 {\scriptsize(8.7)} & \textbf{14.0 {\scriptsize(8.9)}} & 12.0 {\scriptsize(9.5)} & -- & 8.0 {\scriptsize(10.7)} & -- \\
Gemini 3.7 Flash & \textbf{52.0 {\scriptsize(3.1)}} & \textbf{52.0 {\scriptsize(2.5)}} & 47.0 {\scriptsize(2.9)} & 45.0 {\scriptsize(4.3)} & -- & 47.0 {\scriptsize(3.9)} & -- \\
Gemini 3.1 Pro & \textbf{51.0 {\scriptsize(3.2)}} & \textbf{51.0 {\scriptsize(3.0)}} & -- & 50.0 {\scriptsize(3.6)} & -- & 49.0 {\scriptsize(4.3)} & -- \\
Grok 4.20 & 22.0 {\scriptsize(8.1)} & \textbf{24.0 {\scriptsize(7.6)}} & -- & 19.0 {\scriptsize(10.1)} & -- & 13.0 {\scriptsize(11.3)} & -- \\
Grok 4.5 & 14.0 {\scriptsize(12.5)} & \textbf{19.0 {\scriptsize(10.5)}} & -- & 12.0 {\scriptsize(18.1)} & -- & 12.0 {\scriptsize(21.7)} & -- \\
Claude Sonnet 5 & 24.0 {\scriptsize(7.0)} & \textbf{25.0 {\scriptsize(5.2)}} & -- & 21.0 {\scriptsize(8.2)} & -- & 21.0 {\scriptsize(7.9)} & -- \\
\midrule\multicolumn{8}{@{}l}{\emph{PixMo-Count (photos, 2--10 objects)}} \\
Qwen3-VL-8B & 62.2 {\scriptsize(0.6)} & 64.4 {\scriptsize(0.8)} & 53.3 {\scriptsize(4.6)} & 57.8 {\scriptsize(0.6)} & \textbf{66.7 {\scriptsize(0.5)}} & 53.3 {\scriptsize(1.2)} & n/a \\
Qwen3-VL-32B & 65.6 {\scriptsize(0.5)} & 75.6 {\scriptsize(0.7)} & \textbf{81.1 {\scriptsize(0.6)}} & 78.9 {\scriptsize(0.7)} & 78.9 {\scriptsize(0.3)} & 78.9 {\scriptsize(0.4)} & n/a \\
Gemini 3.1 Flash-Lite & 71.1 {\scriptsize(0.5)} & \textbf{78.9 {\scriptsize(0.7)}} & 72.2 {\scriptsize(1.4)} & 71.1 {\scriptsize(0.4)} & 75.6 {\scriptsize(0.4)} & 77.8 {\scriptsize(0.3)} & n/a \\
GPT-5 mini & 66.7 {\scriptsize(0.5)} & 65.6 {\scriptsize(1.1)} & \textbf{68.9 {\scriptsize(0.6)}} & 65.6 {\scriptsize(1.1)} & -- & 61.1 {\scriptsize(0.8)} & n/a \\
Gemini 3.7 Flash & \textbf{88.9 {\scriptsize(0.2)}} & \textbf{88.9 {\scriptsize(0.2)}} & 87.8 {\scriptsize(0.2)} & 86.7 {\scriptsize(0.2)} & -- & 83.3 {\scriptsize(0.2)} & n/a \\
Gemini 3.1 Pro & \textbf{92.2 {\scriptsize(0.1)}} & \textbf{92.2 {\scriptsize(0.1)}} & -- & 90.0 {\scriptsize(0.2)} & -- & \textbf{92.2 {\scriptsize(0.1)}} & n/a \\
Grok 4.20 & 74.4 {\scriptsize(0.6)} & 75.6 {\scriptsize(0.3)} & -- & 75.6 {\scriptsize(0.4)} & -- & \textbf{78.9 {\scriptsize(0.3)}} & n/a \\
Grok 4.5 & 58.9 {\scriptsize(0.7)} & \textbf{76.7 {\scriptsize(0.3)}} & -- & 67.8 {\scriptsize(0.6)} & -- & 68.9 {\scriptsize(0.6)} & n/a \\
Claude Sonnet 5 & 74.4 {\scriptsize(0.4)} & 76.7 {\scriptsize(0.8)} & -- & \textbf{82.2 {\scriptsize(0.3)}} & -- & 78.9 {\scriptsize(0.4)} & n/a \\
\bottomrule
\end{tabular}
\end{table}

\begin{table}[h]
\centering\scriptsize
\setlength{\tabcolsep}{4.5pt}
\renewcommand{\arraystretch}{0.96}
\caption{Exact-count accuracy (\%) with MAE in parentheses on the FULL image sets (\nDot{} DotArrays, \nFsc{} FSC-147, \nPixmo{} PixMo-Count) for the models that ran them; tier-3 models only saw the common subset, so their rows repeat \cref{tab:main}. \arm{tiles} on PixMo-Count is omitted for the reason given in \cref{tab:maincommonfull}.}
\label{tab:mainfull}
\begin{tabular}{@{}lrrrrrrr@{}}
\toprule
model & direct & CoT & list & tag & tag+crop & full toolbox & tiles \\
\midrule\multicolumn{8}{@{}l}{\emph{DotArrays (synthetic, $n\le60$)}} \\
Qwen3-VL-8B & 57.4 {\scriptsize(2.8)} & 41.9 {\scriptsize(67.1)} & 50.4 {\scriptsize(19.9)} & 58.1 {\scriptsize(4.8)} & \textbf{59.2 {\scriptsize(5.4)}} & 54.4 {\scriptsize(11.2)} & 11.8 {\scriptsize(5.9)} \\
Qwen3-VL-32B & 60.7 {\scriptsize(1.6)} & 53.7 {\scriptsize(2.7)} & 66.5 {\scriptsize(6.6)} & \textbf{68.0 {\scriptsize(17.5)}} & 65.1 {\scriptsize(49.4)} & 64.3 {\scriptsize(12.5)} & 15.1 {\scriptsize(5.3)} \\
Gemini 3.1 Flash-Lite & 52.9 {\scriptsize(1.2)} & \textbf{77.2 {\scriptsize(3.0)}} & 75.7 {\scriptsize(1.9)} & 73.9 {\scriptsize(2.5)} & 75.4 {\scriptsize(4.8)} & 76.5 {\scriptsize(2.3)} & 64.0 {\scriptsize(0.8)} \\
GPT-5 mini & \textbf{57.7 {\scriptsize(1.5)}} & 54.4 {\scriptsize(1.4)} & 53.3 {\scriptsize(1.4)} & 36.8 {\scriptsize(2.4)} & -- & 39.7 {\scriptsize(2.3)} & -- \\
Gemini 3.7 Flash & 97.1 {\scriptsize(0.0)} & \textbf{99.3 {\scriptsize(0.0)}} & 95.2 {\scriptsize(0.2)} & 98.2 {\scriptsize(0.0)} & -- & 97.4 {\scriptsize(0.1)} & -- \\
Gemini 3.1 Pro & 96.0 {\scriptsize(0.1)} & 96.0 {\scriptsize(0.1)} & -- & 96.0 {\scriptsize(0.0)} & -- & \textbf{100.0 {\scriptsize(0.0)}} & -- \\
Grok 4.20 & \textbf{66.0 {\scriptsize(1.1)}} & \textbf{66.0 {\scriptsize(1.7)}} & -- & 50.0 {\scriptsize(3.1)} & -- & 42.0 {\scriptsize(3.7)} & -- \\
Grok 4.5 & 54.0 {\scriptsize(1.7)} & \textbf{66.0 {\scriptsize(1.1)}} & -- & 28.0 {\scriptsize(7.1)} & -- & 44.0 {\scriptsize(3.8)} & -- \\
Claude Sonnet 5 & 72.0 {\scriptsize(1.0)} & \textbf{80.0 {\scriptsize(0.3)}} & -- & 62.0 {\scriptsize(7.1)} & -- & 58.0 {\scriptsize(6.0)} & -- \\
\midrule\multicolumn{8}{@{}l}{\emph{FSC-147 (photos, 8--80 objects)}} \\
Qwen3-VL-8B & 12.0 {\scriptsize(9.7)} & 10.0 {\scriptsize(86.1)} & 14.0 {\scriptsize(45.2)} & 12.5 {\scriptsize(19.8)} & \textbf{15.5 {\scriptsize(23.5)}} & 12.0 {\scriptsize(21.3)} & 3.0 {\scriptsize(18.8)} \\
Qwen3-VL-32B & 23.0 {\scriptsize(8.0)} & 19.5 {\scriptsize(21.2)} & 25.0 {\scriptsize(19.8)} & \textbf{27.0 {\scriptsize(20.2)}} & 24.0 {\scriptsize(15.5)} & 26.0 {\scriptsize(17.0)} & 2.5 {\scriptsize(13.3)} \\
Gemini 3.1 Flash-Lite & 20.5 {\scriptsize(11.7)} & \textbf{39.0 {\scriptsize(9.5)}} & 25.5 {\scriptsize(13.1)} & 24.0 {\scriptsize(6.9)} & 25.5 {\scriptsize(7.6)} & 27.0 {\scriptsize(7.5)} & 3.0 {\scriptsize(13.1)} \\
GPT-5 mini & 12.5 {\scriptsize(9.2)} & \textbf{15.5 {\scriptsize(8.2)}} & 15.0 {\scriptsize(8.3)} & 14.0 {\scriptsize(9.4)} & -- & 11.5 {\scriptsize(9.5)} & -- \\
Gemini 3.7 Flash & \textbf{48.5 {\scriptsize(3.1)}} & \textbf{48.5 {\scriptsize(4.8)}} & 45.0 {\scriptsize(4.1)} & 42.5 {\scriptsize(3.5)} & -- & 43.0 {\scriptsize(3.4)} & -- \\
Gemini 3.1 Pro & \textbf{51.0 {\scriptsize(3.2)}} & \textbf{51.0 {\scriptsize(3.0)}} & -- & 50.0 {\scriptsize(3.6)} & -- & 49.0 {\scriptsize(4.3)} & -- \\
Grok 4.20 & 22.0 {\scriptsize(8.1)} & \textbf{24.0 {\scriptsize(7.6)}} & -- & 19.0 {\scriptsize(10.1)} & -- & 13.0 {\scriptsize(11.3)} & -- \\
Grok 4.5 & 14.0 {\scriptsize(12.5)} & \textbf{19.0 {\scriptsize(10.5)}} & -- & 12.0 {\scriptsize(18.1)} & -- & 12.0 {\scriptsize(21.7)} & -- \\
Claude Sonnet 5 & 24.0 {\scriptsize(7.0)} & \textbf{25.0 {\scriptsize(5.2)}} & -- & 21.0 {\scriptsize(8.2)} & -- & 21.0 {\scriptsize(7.9)} & -- \\
\midrule\multicolumn{8}{@{}l}{\emph{PixMo-Count (photos, 2--10 objects)}} \\
Qwen3-VL-8B & \textbf{66.1 {\scriptsize(0.5)}} & 62.8 {\scriptsize(0.6)} & 58.3 {\scriptsize(5.4)} & 59.4 {\scriptsize(0.6)} & 65.0 {\scriptsize(0.6)} & 54.4 {\scriptsize(2.6)} & n/a \\
Qwen3-VL-32B & 69.4 {\scriptsize(0.4)} & 74.4 {\scriptsize(0.5)} & \textbf{82.2 {\scriptsize(0.4)}} & 78.9 {\scriptsize(0.5)} & 77.8 {\scriptsize(0.3)} & 77.8 {\scriptsize(0.4)} & n/a \\
Gemini 3.1 Flash-Lite & 73.9 {\scriptsize(0.4)} & 81.7 {\scriptsize(0.4)} & 77.2 {\scriptsize(0.8)} & 77.2 {\scriptsize(0.3)} & 79.4 {\scriptsize(0.3)} & \textbf{82.2 {\scriptsize(0.2)}} & n/a \\
GPT-5 mini & 68.9 {\scriptsize(0.5)} & 68.9 {\scriptsize(0.7)} & \textbf{71.7 {\scriptsize(0.5)}} & 62.2 {\scriptsize(0.9)} & -- & 66.7 {\scriptsize(0.6)} & n/a \\
Gemini 3.7 Flash & \textbf{90.0 {\scriptsize(0.2)}} & \textbf{90.0 {\scriptsize(0.2)}} & 89.4 {\scriptsize(0.2)} & 87.2 {\scriptsize(0.2)} & -- & 85.0 {\scriptsize(0.2)} & n/a \\
Gemini 3.1 Pro & \textbf{92.2 {\scriptsize(0.1)}} & \textbf{92.2 {\scriptsize(0.1)}} & -- & 90.0 {\scriptsize(0.2)} & -- & \textbf{92.2 {\scriptsize(0.1)}} & n/a \\
Grok 4.20 & 74.4 {\scriptsize(0.6)} & 75.6 {\scriptsize(0.3)} & -- & 75.6 {\scriptsize(0.4)} & -- & \textbf{78.9 {\scriptsize(0.3)}} & n/a \\
Grok 4.5 & 58.9 {\scriptsize(0.7)} & \textbf{76.7 {\scriptsize(0.3)}} & -- & 67.8 {\scriptsize(0.6)} & -- & 68.9 {\scriptsize(0.6)} & n/a \\
Claude Sonnet 5 & 74.4 {\scriptsize(0.4)} & 76.7 {\scriptsize(0.8)} & -- & \textbf{82.2 {\scriptsize(0.3)}} & -- & 78.9 {\scriptsize(0.4)} & n/a \\
\bottomrule
\end{tabular}
\end{table}

\begin{table}[t]
\centering\scriptsize
\setlength{\tabcolsep}{3pt}
\caption{Ablations on the ablation subset (\nAbItems{} images: DotArrays $n\ge5$ and FSC-147). Exact accuracy (\%) with MAE; duplicates and phantoms per 100 ground-truth objects; mean tool calls.}
\label{tab:ablation}
\begin{tabular}{@{}llrrrrr@{}}
\toprule
model & arm & acc (MAE) & on-target & dup. & phantom & calls \\
\midrule
Qwen3-VL-32B & direct & 32.2 {\scriptsize(6.1)} & -- & -- & -- & 1.0 \\
 & list (blind points) & 32.9 {\scriptsize(16.2)} & 76.2 & 12.8 & 39.0 & 1.0 \\
 & tag, no nudge & 35.0 {\scriptsize(26.1)} & 64.5 & 29.2 & 92.8 & 3.4 \\
 & plain marks & 35.0 {\scriptsize(12.5)} & 65.7 & 42.2 & 98.6 & 3.6 \\
 & tag & 35.7 {\scriptsize(23.3)} & 66.7 & 33.5 & 98.8 & 3.4 \\
 & tag+grid & 34.3 {\scriptsize(22.4)} & 67.2 & 16.1 & 85.8 & 3.1 \\
 & tag+erase & 30.8 {\scriptsize(24.6)} & 63.0 & 40.5 & 89.4 & 3.4 \\
 & tag+crop & 32.2 {\scriptsize(15.0)} & 62.9 & 16.0 & 65.4 & 3.6 \\
 & crop only & 25.2 {\scriptsize(8.8)} & 0.0 & 0.0 & 0.0 & 4.1 \\
 & full toolbox & 33.6 {\scriptsize(24.0)} & 63.3 & 28.0 & 98.6 & 3.7 \\
 & tiles (scripted) & 6.3 {\scriptsize(10.5)} & 74.1 & 14.0 & 41.1 & 4.0 \\
 & tag, T=2 & 32.2 {\scriptsize(13.6)} & 67.5 & 19.4 & 71.9 & 2.3 \\
 & tag, T=10 & 30.3 {\scriptsize(23.3)} & 67.9 & 28.6 & 100.5 & 3.8 \\
 & full, T=10 & 32.9 {\scriptsize(26.4)} & 65.2 & 34.2 & 107.5 & 4.5 \\
Gemini 3.1 Flash-Lite & direct & 25.2 {\scriptsize(3.8)} & -- & -- & -- & 1.0 \\
 & list (blind points) & 41.3 {\scriptsize(7.2)} & 57.9 & 5.9 & 39.9 & 1.0 \\
 & tag, no nudge & 45.5 {\scriptsize(6.2)} & 70.7 & 4.9 & 27.8 & 2.8 \\
 & plain marks & 43.4 {\scriptsize(4.0)} & 68.9 & 6.3 & 27.0 & 2.7 \\
 & tag & 39.9 {\scriptsize(6.8)} & 67.9 & 2.4 & 23.3 & 2.6 \\
 & tag+grid & 45.5 {\scriptsize(3.8)} & 71.9 & 2.9 & 19.8 & 2.4 \\
 & tag+erase & 42.0 {\scriptsize(5.6)} & 70.4 & 3.6 & 29.2 & 2.6 \\
 & tag+crop & 43.4 {\scriptsize(5.6)} & 71.0 & 3.8 & 29.0 & 2.6 \\
 & crop only & 28.0 {\scriptsize(10.7)} & 0.0 & 0.0 & 0.0 & 1.3 \\
 & full toolbox & 44.1 {\scriptsize(5.8)} & 71.8 & 4.4 & 27.0 & 3.0 \\
 & tiles (scripted) & 26.6 {\scriptsize(7.6)} & 82.3 & 9.8 & 28.0 & 4.0 \\
 & tag, T=2 & 41.3 {\scriptsize(4.4)} & 70.6 & 2.7 & 21.5 & 2.1 \\
 & tag, T=10 & 42.7 {\scriptsize(5.5)} & 70.6 & 3.9 & 26.5 & 2.6 \\
 & full, T=10 & 39.9 {\scriptsize(5.4)} & 73.8 & 4.4 & 25.5 & 3.0 \\
\bottomrule
\end{tabular}
\end{table}

\begin{table}[h]
\centering\tiny
\setlength{\tabcolsep}{2.5pt}
\caption{FSC-147 by count bin: exact accuracy (\%) with MAE.}
\label{tab:fscbins}
\begin{tabular}{@{}llrrrrr@{}}
\toprule
model & arm & 8-15 & 16-25 & 26-40 & 41-60 & 61-80 \\
\midrule
Qwen3-VL-8B & direct & 40.0 {\scriptsize(3.4)} & 10.0 {\scriptsize(3.1)} & 5.0 {\scriptsize(5.0)} & 2.5 {\scriptsize(16.6)} & 2.5 {\scriptsize(20.4)} \\
 & list & 45.0 {\scriptsize(7.2)} & 15.0 {\scriptsize(50.4)} & 5.0 {\scriptsize(59.6)} & 2.5 {\scriptsize(57.9)} & 2.5 {\scriptsize(52.6)} \\
 & tag & 32.5 {\scriptsize(5.1)} & 17.5 {\scriptsize(5.8)} & 10.0 {\scriptsize(21.0)} & 2.5 {\scriptsize(39.4)} & 0.0 {\scriptsize(32.1)} \\
 & full & 35.0 {\scriptsize(2.5)} & 17.5 {\scriptsize(12.2)} & 5.0 {\scriptsize(28.2)} & 0.0 {\scriptsize(23.4)} & 2.5 {\scriptsize(40.0)} \\
 & tiles & 5.0 {\scriptsize(7.8)} & 5.0 {\scriptsize(8.3)} & 2.5 {\scriptsize(12.7)} & 2.5 {\scriptsize(25.5)} & 0.0 {\scriptsize(39.9)} \\
Qwen3-VL-32B & direct & 60.0 {\scriptsize(2.5)} & 27.5 {\scriptsize(2.7)} & 15.0 {\scriptsize(4.1)} & 10.0 {\scriptsize(12.6)} & 2.5 {\scriptsize(18.0)} \\
 & list & 60.0 {\scriptsize(1.8)} & 32.5 {\scriptsize(1.6)} & 20.0 {\scriptsize(9.4)} & 7.5 {\scriptsize(42.0)} & 5.0 {\scriptsize(44.3)} \\
 & tag & 62.5 {\scriptsize(0.5)} & 42.5 {\scriptsize(3.5)} & 17.5 {\scriptsize(15.6)} & 10.0 {\scriptsize(34.7)} & 2.5 {\scriptsize(51.9)} \\
 & full & 60.0 {\scriptsize(3.0)} & 40.0 {\scriptsize(2.8)} & 17.5 {\scriptsize(19.6)} & 7.5 {\scriptsize(23.6)} & 5.0 {\scriptsize(39.0)} \\
 & tiles & 5.0 {\scriptsize(9.3)} & 2.5 {\scriptsize(7.7)} & 2.5 {\scriptsize(11.7)} & 0.0 {\scriptsize(16.9)} & 2.5 {\scriptsize(20.9)} \\
Gemini 3.1 Flash-Lite & direct & 42.5 {\scriptsize(31.8)} & 22.5 {\scriptsize(2.4)} & 17.5 {\scriptsize(4.0)} & 10.0 {\scriptsize(9.6)} & 10.0 {\scriptsize(11.0)} \\
 & list & 37.5 {\scriptsize(9.2)} & 42.5 {\scriptsize(5.5)} & 27.5 {\scriptsize(10.0)} & 7.5 {\scriptsize(20.5)} & 12.5 {\scriptsize(21.1)} \\
 & tag & 45.0 {\scriptsize(4.5)} & 35.0 {\scriptsize(2.2)} & 20.0 {\scriptsize(3.9)} & 7.5 {\scriptsize(10.6)} & 12.5 {\scriptsize(13.2)} \\
 & full & 50.0 {\scriptsize(2.8)} & 42.5 {\scriptsize(1.7)} & 22.5 {\scriptsize(6.3)} & 7.5 {\scriptsize(10.6)} & 12.5 {\scriptsize(15.8)} \\
 & tiles & 5.0 {\scriptsize(14.3)} & 10.0 {\scriptsize(7.8)} & 0.0 {\scriptsize(11.8)} & 0.0 {\scriptsize(12.1)} & 0.0 {\scriptsize(19.6)} \\
GPT-5 mini & direct & 35.0 {\scriptsize(2.1)} & 12.5 {\scriptsize(3.5)} & 7.5 {\scriptsize(5.5)} & 5.0 {\scriptsize(16.7)} & 2.5 {\scriptsize(18.3)} \\
 & list & 42.5 {\scriptsize(1.9)} & 17.5 {\scriptsize(3.5)} & 5.0 {\scriptsize(5.1)} & 7.5 {\scriptsize(12.3)} & 2.5 {\scriptsize(19.1)} \\
 & tag & 40.0 {\scriptsize(1.2)} & 22.5 {\scriptsize(3.4)} & 0.0 {\scriptsize(6.9)} & 5.0 {\scriptsize(14.0)} & 2.5 {\scriptsize(22.2)} \\
 & full & 37.5 {\scriptsize(1.4)} & 15.0 {\scriptsize(4.4)} & 0.0 {\scriptsize(7.8)} & 2.5 {\scriptsize(14.7)} & 2.5 {\scriptsize(21.3)} \\
Gemini 3.7 Flash & direct & 70.0 {\scriptsize(2.5)} & 70.0 {\scriptsize(1.1)} & 47.5 {\scriptsize(2.0)} & 32.5 {\scriptsize(4.2)} & 22.5 {\scriptsize(5.5)} \\
 & list & 65.0 {\scriptsize(8.4)} & 62.5 {\scriptsize(1.0)} & 45.0 {\scriptsize(2.0)} & 32.5 {\scriptsize(3.6)} & 20.0 {\scriptsize(5.5)} \\
 & tag & 65.0 {\scriptsize(3.0)} & 65.0 {\scriptsize(0.7)} & 40.0 {\scriptsize(1.9)} & 15.0 {\scriptsize(3.9)} & 27.5 {\scriptsize(7.8)} \\
 & full & 70.0 {\scriptsize(2.9)} & 67.5 {\scriptsize(0.7)} & 32.5 {\scriptsize(2.5)} & 22.5 {\scriptsize(4.4)} & 22.5 {\scriptsize(6.5)} \\
Gemini 3.1 Pro & direct & 75.0 {\scriptsize(3.5)} & 75.0 {\scriptsize(0.3)} & 40.0 {\scriptsize(2.4)} & 45.0 {\scriptsize(3.8)} & 20.0 {\scriptsize(6.2)} \\
 & tag & 75.0 {\scriptsize(3.4)} & 85.0 {\scriptsize(0.4)} & 45.0 {\scriptsize(4.0)} & 35.0 {\scriptsize(4.0)} & 10.0 {\scriptsize(6.3)} \\
 & full & 75.0 {\scriptsize(3.4)} & 80.0 {\scriptsize(0.3)} & 30.0 {\scriptsize(4.6)} & 45.0 {\scriptsize(5.8)} & 15.0 {\scriptsize(7.5)} \\
Grok 4.20 & direct & 40.0 {\scriptsize(6.0)} & 45.0 {\scriptsize(1.9)} & 15.0 {\scriptsize(4.3)} & 5.0 {\scriptsize(10.8)} & 5.0 {\scriptsize(17.7)} \\
 & tag & 45.0 {\scriptsize(2.5)} & 35.0 {\scriptsize(2.1)} & 10.0 {\scriptsize(7.9)} & 0.0 {\scriptsize(13.2)} & 5.0 {\scriptsize(24.6)} \\
 & full & 30.0 {\scriptsize(4.8)} & 30.0 {\scriptsize(2.1)} & 5.0 {\scriptsize(10.4)} & 0.0 {\scriptsize(15.1)} & 0.0 {\scriptsize(23.9)} \\
Grok 4.5 & direct & 40.0 {\scriptsize(5.8)} & 20.0 {\scriptsize(3.1)} & 5.0 {\scriptsize(7.8)} & 0.0 {\scriptsize(22.1)} & 5.0 {\scriptsize(23.8)} \\
 & tag & 25.0 {\scriptsize(6.7)} & 25.0 {\scriptsize(5.1)} & 5.0 {\scriptsize(15.6)} & 5.0 {\scriptsize(25.9)} & 0.0 {\scriptsize(37.1)} \\
 & full & 25.0 {\scriptsize(7.1)} & 25.0 {\scriptsize(4.7)} & 10.0 {\scriptsize(18.6)} & 0.0 {\scriptsize(41.1)} & 0.0 {\scriptsize(37.1)} \\
Claude Sonnet 5 & direct & 60.0 {\scriptsize(2.4)} & 30.0 {\scriptsize(1.6)} & 20.0 {\scriptsize(4.5)} & 5.0 {\scriptsize(14.1)} & 5.0 {\scriptsize(12.3)} \\
 & tag & 40.0 {\scriptsize(1.2)} & 40.0 {\scriptsize(3.1)} & 20.0 {\scriptsize(6.5)} & 5.0 {\scriptsize(13.4)} & 0.0 {\scriptsize(16.8)} \\
 & full & 30.0 {\scriptsize(1.6)} & 40.0 {\scriptsize(3.2)} & 20.0 {\scriptsize(8.2)} & 0.0 {\scriptsize(7.8)} & 15.0 {\scriptsize(18.9)} \\
\bottomrule
\end{tabular}
\end{table}

\begin{table}[h]
\centering\tiny
\setlength{\tabcolsep}{2.5pt}
\caption{PixMo-Count by ground-truth count: exact accuracy (\%) with MAE.}
\label{tab:pixmo}
\begin{tabular}{@{}llrrrrrrrrr@{}}
\toprule
model & arm & 2 & 3 & 4 & 5 & 6 & 7 & 8 & 9 & 10 \\
\midrule
Qwen3-VL-8B & direct & 95.0 {\scriptsize(0.1)} & 95.0 {\scriptsize(0.1)} & 75.0 {\scriptsize(0.6)} & 75.0 {\scriptsize(0.5)} & 60.0 {\scriptsize(0.5)} & 65.0 {\scriptsize(0.3)} & 45.0 {\scriptsize(0.7)} & 50.0 {\scriptsize(0.7)} & 35.0 {\scriptsize(1.1)} \\
 & list & 80.0 {\scriptsize(0.2)} & 75.0 {\scriptsize(0.2)} & 70.0 {\scriptsize(8.3)} & 75.0 {\scriptsize(2.2)} & 25.0 {\scriptsize(4.7)} & 45.0 {\scriptsize(13.8)} & 55.0 {\scriptsize(3.6)} & 55.0 {\scriptsize(5.0)} & 45.0 {\scriptsize(11.4)} \\
 & tag & 90.0 {\scriptsize(0.1)} & 100.0 {\scriptsize(0.0)} & 65.0 {\scriptsize(0.7)} & 40.0 {\scriptsize(0.8)} & 40.0 {\scriptsize(1.4)} & 50.0 {\scriptsize(0.6)} & 55.0 {\scriptsize(0.5)} & 60.0 {\scriptsize(0.8)} & 35.0 {\scriptsize(0.6)} \\
 & full & 90.0 {\scriptsize(0.1)} & 90.0 {\scriptsize(0.1)} & 55.0 {\scriptsize(9.2)} & 45.0 {\scriptsize(0.9)} & 40.0 {\scriptsize(5.9)} & 65.0 {\scriptsize(2.1)} & 35.0 {\scriptsize(1.8)} & 45.0 {\scriptsize(1.6)} & 25.0 {\scriptsize(1.7)} \\
 & tiles & 30.0 {\scriptsize(1.6)} & 15.0 {\scriptsize(2.0)} & 15.0 {\scriptsize(3.0)} & 35.0 {\scriptsize(2.0)} & 15.0 {\scriptsize(2.8)} & 10.0 {\scriptsize(4.0)} & 15.0 {\scriptsize(2.9)} & 30.0 {\scriptsize(3.8)} & 10.0 {\scriptsize(5.3)} \\
Qwen3-VL-32B & direct & 95.0 {\scriptsize(0.1)} & 80.0 {\scriptsize(0.3)} & 85.0 {\scriptsize(0.3)} & 80.0 {\scriptsize(0.2)} & 65.0 {\scriptsize(0.3)} & 65.0 {\scriptsize(0.5)} & 65.0 {\scriptsize(0.7)} & 60.0 {\scriptsize(0.6)} & 30.0 {\scriptsize(0.9)} \\
 & list & 95.0 {\scriptsize(0.1)} & 95.0 {\scriptsize(0.1)} & 75.0 {\scriptsize(1.8)} & 90.0 {\scriptsize(0.1)} & 70.0 {\scriptsize(0.6)} & 75.0 {\scriptsize(0.3)} & 85.0 {\scriptsize(0.1)} & 75.0 {\scriptsize(0.3)} & 80.0 {\scriptsize(0.2)} \\
 & tag & 95.0 {\scriptsize(0.1)} & 85.0 {\scriptsize(0.2)} & 85.0 {\scriptsize(1.8)} & 90.0 {\scriptsize(0.1)} & 70.0 {\scriptsize(0.4)} & 70.0 {\scriptsize(0.3)} & 75.0 {\scriptsize(0.2)} & 75.0 {\scriptsize(0.3)} & 65.0 {\scriptsize(0.8)} \\
 & full & 95.0 {\scriptsize(0.1)} & 85.0 {\scriptsize(0.2)} & 90.0 {\scriptsize(0.1)} & 85.0 {\scriptsize(0.1)} & 70.0 {\scriptsize(0.7)} & 70.0 {\scriptsize(0.3)} & 85.0 {\scriptsize(0.1)} & 75.0 {\scriptsize(0.2)} & 45.0 {\scriptsize(1.4)} \\
 & tiles & 30.0 {\scriptsize(1.6)} & 25.0 {\scriptsize(2.1)} & 15.0 {\scriptsize(3.1)} & 30.0 {\scriptsize(3.0)} & 10.0 {\scriptsize(3.0)} & 15.0 {\scriptsize(5.2)} & 20.0 {\scriptsize(3.1)} & 15.0 {\scriptsize(3.2)} & 15.0 {\scriptsize(5.6)} \\
Gemini 3.1 Flash-Lite & direct & 95.0 {\scriptsize(0.1)} & 85.0 {\scriptsize(0.2)} & 90.0 {\scriptsize(0.4)} & 75.0 {\scriptsize(0.2)} & 60.0 {\scriptsize(0.4)} & 50.0 {\scriptsize(0.6)} & 70.0 {\scriptsize(0.3)} & 75.0 {\scriptsize(0.4)} & 65.0 {\scriptsize(0.8)} \\
 & list & 95.0 {\scriptsize(0.1)} & 90.0 {\scriptsize(0.1)} & 95.0 {\scriptsize(0.1)} & 85.0 {\scriptsize(0.1)} & 65.0 {\scriptsize(5.0)} & 60.0 {\scriptsize(0.5)} & 65.0 {\scriptsize(0.3)} & 85.0 {\scriptsize(0.2)} & 55.0 {\scriptsize(0.5)} \\
 & tag & 95.0 {\scriptsize(0.1)} & 95.0 {\scriptsize(0.1)} & 95.0 {\scriptsize(0.2)} & 80.0 {\scriptsize(0.2)} & 55.0 {\scriptsize(0.5)} & 55.0 {\scriptsize(0.7)} & 75.0 {\scriptsize(0.2)} & 80.0 {\scriptsize(0.2)} & 65.0 {\scriptsize(0.7)} \\
 & full & 95.0 {\scriptsize(0.1)} & 95.0 {\scriptsize(0.1)} & 90.0 {\scriptsize(0.2)} & 80.0 {\scriptsize(0.2)} & 75.0 {\scriptsize(0.2)} & 70.0 {\scriptsize(0.3)} & 90.0 {\scriptsize(0.1)} & 75.0 {\scriptsize(0.3)} & 70.0 {\scriptsize(0.4)} \\
 & tiles & 25.0 {\scriptsize(2.5)} & 10.0 {\scriptsize(2.9)} & 10.0 {\scriptsize(3.7)} & 15.0 {\scriptsize(4.1)} & 15.0 {\scriptsize(4.2)} & 15.0 {\scriptsize(5.0)} & 10.0 {\scriptsize(3.5)} & 0.0 {\scriptsize(4.5)} & 0.0 {\scriptsize(6.1)} \\
GPT-5 mini & direct & 95.0 {\scriptsize(0.1)} & 90.0 {\scriptsize(0.1)} & 75.0 {\scriptsize(0.4)} & 75.0 {\scriptsize(0.2)} & 70.0 {\scriptsize(0.3)} & 70.0 {\scriptsize(0.3)} & 55.0 {\scriptsize(0.6)} & 40.0 {\scriptsize(0.7)} & 50.0 {\scriptsize(1.4)} \\
 & list & 100.0 {\scriptsize(0.0)} & 100.0 {\scriptsize(0.0)} & 75.0 {\scriptsize(1.1)} & 65.0 {\scriptsize(0.3)} & 75.0 {\scriptsize(0.4)} & 70.0 {\scriptsize(0.4)} & 60.0 {\scriptsize(0.5)} & 50.0 {\scriptsize(0.6)} & 50.0 {\scriptsize(0.8)} \\
 & tag & 95.0 {\scriptsize(0.1)} & 90.0 {\scriptsize(0.1)} & 70.0 {\scriptsize(2.5)} & 60.0 {\scriptsize(0.6)} & 60.0 {\scriptsize(0.6)} & 50.0 {\scriptsize(0.8)} & 40.0 {\scriptsize(0.7)} & 50.0 {\scriptsize(0.8)} & 45.0 {\scriptsize(1.5)} \\
 & full & 95.0 {\scriptsize(0.1)} & 95.0 {\scriptsize(0.1)} & 80.0 {\scriptsize(1.1)} & 65.0 {\scriptsize(0.7)} & 65.0 {\scriptsize(0.6)} & 65.0 {\scriptsize(0.5)} & 45.0 {\scriptsize(0.7)} & 50.0 {\scriptsize(0.7)} & 40.0 {\scriptsize(1.1)} \\
Gemini 3.7 Flash & direct & 95.0 {\scriptsize(0.1)} & 95.0 {\scriptsize(0.1)} & 85.0 {\scriptsize(0.3)} & 90.0 {\scriptsize(0.1)} & 85.0 {\scriptsize(0.3)} & 85.0 {\scriptsize(0.2)} & 100.0 {\scriptsize(0.0)} & 90.0 {\scriptsize(0.1)} & 85.0 {\scriptsize(0.2)} \\
 & list & 95.0 {\scriptsize(0.1)} & 95.0 {\scriptsize(0.1)} & 85.0 {\scriptsize(0.3)} & 90.0 {\scriptsize(0.1)} & 90.0 {\scriptsize(0.2)} & 85.0 {\scriptsize(0.2)} & 95.0 {\scriptsize(0.1)} & 85.0 {\scriptsize(0.1)} & 85.0 {\scriptsize(0.7)} \\
 & tag & 95.0 {\scriptsize(0.1)} & 95.0 {\scriptsize(0.1)} & 85.0 {\scriptsize(0.3)} & 85.0 {\scriptsize(0.1)} & 85.0 {\scriptsize(0.3)} & 85.0 {\scriptsize(0.2)} & 100.0 {\scriptsize(0.0)} & 70.0 {\scriptsize(0.2)} & 85.0 {\scriptsize(0.3)} \\
 & full & 90.0 {\scriptsize(0.1)} & 95.0 {\scriptsize(0.1)} & 85.0 {\scriptsize(0.3)} & 80.0 {\scriptsize(0.2)} & 80.0 {\scriptsize(0.3)} & 85.0 {\scriptsize(0.2)} & 85.0 {\scriptsize(0.1)} & 85.0 {\scriptsize(0.1)} & 80.0 {\scriptsize(0.3)} \\
Gemini 3.1 Pro & direct & 100.0 {\scriptsize(0.0)} & 100.0 {\scriptsize(0.0)} & 80.0 {\scriptsize(0.2)} & 100.0 {\scriptsize(0.0)} & 90.0 {\scriptsize(0.1)} & 60.0 {\scriptsize(0.6)} & 100.0 {\scriptsize(0.0)} & 100.0 {\scriptsize(0.0)} & 100.0 {\scriptsize(0.0)} \\
 & tag & 100.0 {\scriptsize(0.0)} & 100.0 {\scriptsize(0.0)} & 70.0 {\scriptsize(0.6)} & 100.0 {\scriptsize(0.0)} & 80.0 {\scriptsize(0.2)} & 70.0 {\scriptsize(0.4)} & 100.0 {\scriptsize(0.0)} & 100.0 {\scriptsize(0.0)} & 90.0 {\scriptsize(0.2)} \\
 & full & 100.0 {\scriptsize(0.0)} & 100.0 {\scriptsize(0.0)} & 70.0 {\scriptsize(0.6)} & 100.0 {\scriptsize(0.0)} & 80.0 {\scriptsize(0.2)} & 80.0 {\scriptsize(0.3)} & 100.0 {\scriptsize(0.0)} & 100.0 {\scriptsize(0.0)} & 100.0 {\scriptsize(0.0)} \\
Grok 4.20 & direct & 80.0 {\scriptsize(0.2)} & 90.0 {\scriptsize(0.2)} & 60.0 {\scriptsize(1.0)} & 90.0 {\scriptsize(0.3)} & 90.0 {\scriptsize(0.2)} & 60.0 {\scriptsize(0.6)} & 60.0 {\scriptsize(1.4)} & 80.0 {\scriptsize(0.3)} & 60.0 {\scriptsize(0.8)} \\
 & tag & 80.0 {\scriptsize(0.2)} & 90.0 {\scriptsize(0.1)} & 80.0 {\scriptsize(0.2)} & 80.0 {\scriptsize(0.2)} & 80.0 {\scriptsize(0.3)} & 80.0 {\scriptsize(0.3)} & 60.0 {\scriptsize(0.7)} & 80.0 {\scriptsize(0.3)} & 50.0 {\scriptsize(0.9)} \\
 & full & 90.0 {\scriptsize(0.1)} & 100.0 {\scriptsize(0.0)} & 80.0 {\scriptsize(0.2)} & 80.0 {\scriptsize(0.2)} & 60.0 {\scriptsize(0.5)} & 80.0 {\scriptsize(0.3)} & 70.0 {\scriptsize(0.5)} & 80.0 {\scriptsize(0.3)} & 70.0 {\scriptsize(0.5)} \\
Grok 4.5 & direct & 90.0 {\scriptsize(0.1)} & 80.0 {\scriptsize(0.3)} & 80.0 {\scriptsize(0.5)} & 60.0 {\scriptsize(0.4)} & 40.0 {\scriptsize(0.8)} & 60.0 {\scriptsize(0.5)} & 40.0 {\scriptsize(1.5)} & 40.0 {\scriptsize(1.0)} & 40.0 {\scriptsize(1.4)} \\
 & tag & 90.0 {\scriptsize(0.1)} & 90.0 {\scriptsize(0.1)} & 90.0 {\scriptsize(0.1)} & 80.0 {\scriptsize(0.2)} & 70.0 {\scriptsize(0.5)} & 60.0 {\scriptsize(0.5)} & 30.0 {\scriptsize(1.3)} & 60.0 {\scriptsize(0.8)} & 40.0 {\scriptsize(2.0)} \\
 & full & 90.0 {\scriptsize(0.1)} & 90.0 {\scriptsize(0.1)} & 90.0 {\scriptsize(0.1)} & 70.0 {\scriptsize(0.5)} & 60.0 {\scriptsize(0.7)} & 60.0 {\scriptsize(0.6)} & 50.0 {\scriptsize(1.1)} & 70.0 {\scriptsize(0.3)} & 40.0 {\scriptsize(1.7)} \\
Claude Sonnet 5 & direct & 90.0 {\scriptsize(0.1)} & 100.0 {\scriptsize(0.0)} & 80.0 {\scriptsize(0.5)} & 80.0 {\scriptsize(0.2)} & 90.0 {\scriptsize(0.2)} & 50.0 {\scriptsize(0.8)} & 70.0 {\scriptsize(0.3)} & 70.0 {\scriptsize(0.3)} & 40.0 {\scriptsize(1.2)} \\
 & tag & 90.0 {\scriptsize(0.1)} & 90.0 {\scriptsize(0.2)} & 90.0 {\scriptsize(0.4)} & 100.0 {\scriptsize(0.0)} & 80.0 {\scriptsize(0.2)} & 70.0 {\scriptsize(0.4)} & 80.0 {\scriptsize(0.5)} & 60.0 {\scriptsize(0.9)} & 80.0 {\scriptsize(0.4)} \\
 & full & 90.0 {\scriptsize(0.1)} & 100.0 {\scriptsize(0.0)} & 80.0 {\scriptsize(0.8)} & 90.0 {\scriptsize(0.1)} & 70.0 {\scriptsize(0.8)} & 80.0 {\scriptsize(0.3)} & 60.0 {\scriptsize(0.6)} & 80.0 {\scriptsize(0.2)} & 60.0 {\scriptsize(0.9)} \\
\bottomrule
\end{tabular}
\end{table}

\begin{table}[h]
\centering\scriptsize
\caption{DotArrays factor sets: exact accuracy (\%) with MAE. hetero = mixed shapes/colours (abstraction), distractor = red among blue, dense = touching discs, small = 6-px discs.}
\label{tab:family}
\begin{tabular}{@{}llrrrrr@{}}
\toprule
model & arm & main & hetero & distractor & dense & small \\
\midrule
Qwen3-VL-8B & direct & 64.4 {\scriptsize(2.7)} & 37.5 {\scriptsize(1.4)} & 25.0 {\scriptsize(2.9)} & 33.3 {\scriptsize(4.8)} & 25.0 {\scriptsize(4.8)} \\
 & list & 59.7 {\scriptsize(13.2)} & 12.5 {\scriptsize(34.0)} & 18.8 {\scriptsize(31.1)} & 8.3 {\scriptsize(38.6)} & 16.7 {\scriptsize(88.8)} \\
 & tag & 68.5 {\scriptsize(3.0)} & 25.0 {\scriptsize(3.9)} & 18.8 {\scriptsize(6.5)} & 0.0 {\scriptsize(12.2)} & 25.0 {\scriptsize(29.3)} \\
 & full & 62.5 {\scriptsize(7.5)} & 37.5 {\scriptsize(7.0)} & 25.0 {\scriptsize(21.6)} & 0.0 {\scriptsize(26.6)} & 25.0 {\scriptsize(52.1)} \\
 & tiles & 12.0 {\scriptsize(5.8)} & 6.2 {\scriptsize(3.5)} & 18.8 {\scriptsize(2.5)} & 0.0 {\scriptsize(9.3)} & 16.7 {\scriptsize(12.3)} \\
Qwen3-VL-32B & direct & 65.7 {\scriptsize(1.3)} & 87.5 {\scriptsize(0.1)} & 37.5 {\scriptsize(1.4)} & 25.0 {\scriptsize(3.6)} & 0.0 {\scriptsize(6.6)} \\
 & list & 72.7 {\scriptsize(1.6)} & 68.8 {\scriptsize(1.1)} & 50.0 {\scriptsize(1.9)} & 41.7 {\scriptsize(9.3)} & 0.0 {\scriptsize(107.5)} \\
 & tag & 75.9 {\scriptsize(14.4)} & 50.0 {\scriptsize(7.2)} & 56.2 {\scriptsize(4.8)} & 8.3 {\scriptsize(59.8)} & 25.0 {\scriptsize(74.9)} \\
 & full & 71.3 {\scriptsize(8.3)} & 50.0 {\scriptsize(4.4)} & 56.2 {\scriptsize(25.8)} & 16.7 {\scriptsize(47.7)} & 16.7 {\scriptsize(56.7)} \\
 & tiles & 13.4 {\scriptsize(5.1)} & 18.8 {\scriptsize(2.8)} & 18.8 {\scriptsize(2.2)} & 8.3 {\scriptsize(17.7)} & 41.7 {\scriptsize(2.7)} \\
Gemini 3.1 Flash-Lite & direct & 58.8 {\scriptsize(1.1)} & 37.5 {\scriptsize(1.4)} & 31.2 {\scriptsize(1.2)} & 33.3 {\scriptsize(2.4)} & 16.7 {\scriptsize(2.2)} \\
 & list & 84.3 {\scriptsize(1.8)} & 43.8 {\scriptsize(1.1)} & 62.5 {\scriptsize(0.6)} & 33.3 {\scriptsize(2.9)} & 25.0 {\scriptsize(5.0)} \\
 & tag & 80.1 {\scriptsize(1.9)} & 56.2 {\scriptsize(0.5)} & 62.5 {\scriptsize(0.6)} & 33.3 {\scriptsize(10.0)} & 41.7 {\scriptsize(10.2)} \\
 & full & 82.4 {\scriptsize(2.1)} & 68.8 {\scriptsize(0.5)} & 68.8 {\scriptsize(0.7)} & 33.3 {\scriptsize(7.4)} & 33.3 {\scriptsize(5.1)} \\
 & tiles & 68.5 {\scriptsize(0.7)} & 62.5 {\scriptsize(0.4)} & 62.5 {\scriptsize(0.4)} & 8.3 {\scriptsize(3.7)} & 41.7 {\scriptsize(0.7)} \\
GPT-5 mini & direct & 62.5 {\scriptsize(1.3)} & 68.8 {\scriptsize(0.4)} & 50.0 {\scriptsize(1.1)} & 16.7 {\scriptsize(3.3)} & 8.3 {\scriptsize(4.7)} \\
 & list & 60.6 {\scriptsize(1.1)} & 56.2 {\scriptsize(1.5)} & 18.8 {\scriptsize(4.6)} & 8.3 {\scriptsize(2.0)} & 8.3 {\scriptsize(2.9)} \\
 & tag & 38.9 {\scriptsize(2.3)} & 62.5 {\scriptsize(1.0)} & 18.8 {\scriptsize(2.5)} & 0.0 {\scriptsize(5.5)} & 25.0 {\scriptsize(2.8)} \\
 & full & 43.5 {\scriptsize(1.9)} & 43.8 {\scriptsize(1.6)} & 31.2 {\scriptsize(3.4)} & 0.0 {\scriptsize(5.3)} & 16.7 {\scriptsize(6.2)} \\
Gemini 3.7 Flash & direct & 96.3 {\scriptsize(0.1)} & 100.0 {\scriptsize(0.0)} & 100.0 {\scriptsize(0.0)} & 100.0 {\scriptsize(0.0)} & 100.0 {\scriptsize(0.0)} \\
 & list & 95.4 {\scriptsize(0.2)} & 93.8 {\scriptsize(0.1)} & 100.0 {\scriptsize(0.0)} & 91.7 {\scriptsize(0.5)} & 91.7 {\scriptsize(0.1)} \\
 & tag & 98.6 {\scriptsize(0.0)} & 93.8 {\scriptsize(0.3)} & 100.0 {\scriptsize(0.0)} & 100.0 {\scriptsize(0.0)} & 91.7 {\scriptsize(0.1)} \\
 & full & 98.6 {\scriptsize(0.0)} & 100.0 {\scriptsize(0.0)} & 100.0 {\scriptsize(0.0)} & 91.7 {\scriptsize(0.4)} & 75.0 {\scriptsize(0.5)} \\
Gemini 3.1 Pro & direct & 97.2 {\scriptsize(0.1)} & 100.0 {\scriptsize(0.0)} & 100.0 {\scriptsize(0.0)} & 66.7 {\scriptsize(0.3)} & 100.0 {\scriptsize(0.0)} \\
 & tag & 97.2 {\scriptsize(0.0)} & 100.0 {\scriptsize(0.0)} & 100.0 {\scriptsize(0.0)} & 66.7 {\scriptsize(0.3)} & 100.0 {\scriptsize(0.0)} \\
 & full & 100.0 {\scriptsize(0.0)} & 100.0 {\scriptsize(0.0)} & 100.0 {\scriptsize(0.0)} & 100.0 {\scriptsize(0.0)} & 100.0 {\scriptsize(0.0)} \\
Grok 4.20 & direct & 63.9 {\scriptsize(1.2)} & 100.0 {\scriptsize(0.0)} & 50.0 {\scriptsize(1.8)} & 66.7 {\scriptsize(0.7)} & 66.7 {\scriptsize(1.7)} \\
 & tag & 50.0 {\scriptsize(3.4)} & 100.0 {\scriptsize(0.0)} & 25.0 {\scriptsize(2.8)} & 33.3 {\scriptsize(2.3)} & 33.3 {\scriptsize(5.3)} \\
 & full & 47.2 {\scriptsize(3.6)} & 75.0 {\scriptsize(0.5)} & 25.0 {\scriptsize(2.0)} & 0.0 {\scriptsize(8.7)} & 0.0 {\scriptsize(6.7)} \\
Grok 4.5 & direct & 55.6 {\scriptsize(1.8)} & 75.0 {\scriptsize(0.5)} & 50.0 {\scriptsize(0.8)} & 33.3 {\scriptsize(3.7)} & 33.3 {\scriptsize(1.0)} \\
 & tag & 30.6 {\scriptsize(6.9)} & 75.0 {\scriptsize(0.5)} & 0.0 {\scriptsize(8.2)} & 0.0 {\scriptsize(15.3)} & 0.0 {\scriptsize(9.0)} \\
 & full & 44.4 {\scriptsize(4.1)} & 75.0 {\scriptsize(0.5)} & 25.0 {\scriptsize(4.0)} & 0.0 {\scriptsize(7.0)} & 66.7 {\scriptsize(1.3)} \\
Claude Sonnet 5 & direct & 75.0 {\scriptsize(1.0)} & 75.0 {\scriptsize(0.2)} & 75.0 {\scriptsize(0.2)} & 66.7 {\scriptsize(1.7)} & 33.3 {\scriptsize(2.0)} \\
 & tag & 66.7 {\scriptsize(6.0)} & 50.0 {\scriptsize(2.0)} & 75.0 {\scriptsize(0.5)} & 33.3 {\scriptsize(23.3)} & 33.3 {\scriptsize(19.3)} \\
 & full & 66.7 {\scriptsize(4.6)} & 50.0 {\scriptsize(3.0)} & 50.0 {\scriptsize(4.5)} & 0.0 {\scriptsize(10.3)} & 33.3 {\scriptsize(24.3)} \\
\bottomrule
\end{tabular}
\end{table}

\begin{table}[h]
\centering\scriptsize
\caption{Mirror consistency (an indirect invariance probe motivated by the order-irrelevance principle; the direct traversal-order test is in \cref{tab:order}): agreement between the count on the original image and on its horizontal mirror, and accuracy on each, for the ablation subset.}
\label{tab:flip}
\begin{tabular}{@{}llrrr@{}}
\toprule
model & arm & agree (\%) & acc original & acc mirrored \\
\midrule
Gemini 3.1 Flash-Lite & direct & 36.4 & 25.2 & 25.9 \\
Gemini 3.1 Flash-Lite & full toolbox & 38.5 & 44.1 & 44.1 \\
Gemini 3.1 Flash-Lite & tag & 39.2 & 39.9 & 45.5 \\
Qwen3-VL-32B & direct & 67.8 & 32.2 & 27.3 \\
Qwen3-VL-32B & full toolbox & 33.6 & 33.6 & 31.5 \\
Qwen3-VL-32B & tag & 32.9 & 35.7 & 32.2 \\
Qwen3-VL-8B & direct & 60.1 & 27.3 & 25.2 \\
Qwen3-VL-8B & full toolbox & 18.2 & 23.8 & 19.6 \\
Qwen3-VL-8B & tag & 25.9 & 25.2 & 25.9 \\
\bottomrule
\end{tabular}
\end{table}

\section{Controls added in revision}
\label{app:controls}
\paragraph{Do the tags follow the answer?}
\label{app:coupling}
A model could estimate a number first and then draw that many points, in which case the tags would record the answer rather than produce it. \Cref{tab:coupling} tests this on every image that ran both \arm{direct} and \arm{tag}: how often the number of tags equals the model's own \arm{direct} answer, and how the per-image tag-count error correlates with the \arm{direct} error. On PixMo-Count the two agree often because both are usually right; on the dense photographs of FSC-147 the tag count equals the direct answer on only \tagEqDirectFscMin{}--\tagEqDirectFscMax{}\,\% of images. The right-hand columns show that departures of the announced count from the tag count are mostly corrections: the announced number is then closer to the truth in \closerPooled{}\,\% of trajectories pooled over models. \Cref{tab:perturb} reports the causal version of the test, the tag-perturbation probe of \cref{sec:protocol}: after the model's first \tool{tag} call the harness removes one on-target tag or adds one tag on an untagged object, renumbers, and returns the altered image with the usual verification message, so that only the model's own emission disagrees with what it sees. A model that reads its tags should notice (re-tag the missing object, or start over) or at least announce the visible number; a model that estimates and then draws should announce the number it emitted. The precise pointers do the former: Gemini 3.7 Flash re-tags the removed object in nearly every dot-array trajectory, Flash-Lite in most, and both announce the visible number. GPT-5 mini repairs least often. Qwen3-VL-8B announces the visible number in every trajectory yet seldom repairs: it reads the count off the tags without checking them, which is its cardinality consistency of \cardProfTagQwenEight{}\,\% seen from the other side.
\begin{table}[h]
\centering\scriptsize
\setlength{\tabcolsep}{3pt}
\caption{Do the tags follow the answer, or the answer the tags? For every image that ran both \arm{direct} and \arm{tag}: the fraction of images on which the model drew exactly as many tags as its \arm{direct} answer (= direct, \%, with 95\,\% CI) and the Spearman correlation $\rho$ between the per-image tag-count error and the \arm{direct} error. Right: cardinality (announced count $=$ number of tags; \arm{tag} and \arm{full}, DotArrays + FSC-147), the number of trajectories where they differ, and how often the announced count is then closer to the truth than the tag count (closer, \%).}
\label{tab:coupling}
\begin{tabular}{@{}lrrrrrrrrr@{}}
\toprule
& \multicolumn{2}{c}{DotArrays} & \multicolumn{2}{c}{FSC-147} & \multicolumn{2}{c}{PixMo-Count} & & & \\
\cmidrule(lr){2-3}\cmidrule(lr){4-5}\cmidrule(lr){6-7}
model & = direct & $\rho$ & = direct & $\rho$ & = direct & $\rho$ & card.\ (\%) & differ & closer (\%) \\
\midrule
Qwen3-VL-8B & 44 {\tiny[38,50]} & +0.14 & 10 {\tiny[6,15]} & +0.03 & 43 {\tiny[36,50]} & +0.01 & 91 {\tiny[89,93]} & 84/901 & 96 {\tiny[92,100]} \\
Qwen3-VL-32B & 51 {\tiny[45,57]} & -0.43 & 20 {\tiny[14,25]} & -0.19 & 77 {\tiny[71,83]} & +0.44 & 80 {\tiny[77,83]} & 180/902 & 87 {\tiny[82,92]} \\
Gemini 3.1 Flash-Lite & 52 {\tiny[46,58]} & +0.11 & 16 {\tiny[12,22]} & +0.20 & 66 {\tiny[59,73]} & +0.49 & 91 {\tiny[89,92]} & 89/943 & 69 {\tiny[58,78]} \\
GPT-5 mini & 31 {\tiny[26,37]} & +0.13 & 17 {\tiny[12,22]} & +0.38 & 55 {\tiny[48,62]} & +0.35 & 66 {\tiny[63,69]} & 304/903 & 86 {\tiny[82,90]} \\
Gemini 3.7 Flash & 92 {\tiny[88,95]} & +0.04 & 52 {\tiny[44,58]} & +0.51 & 94 {\tiny[90,97]} & +0.62 & 90 {\tiny[88,92]} & 92/924 & 55 {\tiny[45,65]} \\
Gemini 3.1 Pro & 92 {\tiny[84,98]} & +0.00 & 53 {\tiny[43,63]} & +0.26 & 92 {\tiny[87,98]} & +0.64 & 89 {\tiny[85,92]} & 34/300 & 50 {\tiny[32,68]} \\
Grok 4.20 & 42 {\tiny[28,56]} & +0.01 & 19 {\tiny[12,27]} & +0.39 & 76 {\tiny[67,84]} & +0.63 & 94 {\tiny[92,97]} & 17/300 & 88 {\tiny[71,100]} \\
Grok 4.5 & 16 {\tiny[6,26]} & +0.45 & 15 {\tiny[8,22]} & +0.62 & 67 {\tiny[57,77]} & +0.51 & 54 {\tiny[48,60]} & 138/300 & 96 {\tiny[92,99]} \\
Claude Sonnet 5 & 30 {\tiny[18,44]} & +0.03 & 10 {\tiny[5,16]} & -0.03 & 68 {\tiny[58,77]} & +0.32 & 58 {\tiny[52,63]} & 126/300 & 87 {\tiny[81,93]} \\
\bottomrule
\end{tabular}
\end{table}

\begin{table}[h]
\centering\scriptsize
\setlength{\tabcolsep}{3pt}
\caption{Tag-perturbation probe. After the model's first \tool{tag} call the harness silently removes one on-target tag ($-1$) or adds one tag on a still-untagged object ($+1$) before rendering the working image; tags are renumbered, so the image and the harness message agree and only the model's own emission differs. $n$: trajectories in which the perturbation could be applied (a $+1$ needs an untagged object). follows image: announced count $=$ number of tags visible at the end; follows own: announced count $=$ number of points the model emitted (and $\ne$ visible); repaired: the removed object was re-tagged by the end, or the model started over with \texttt{replace} ($-1$) / started over ($+1$); shift: announced count minus the unperturbed \arm{tag} answer on the same image (mean, 95\,\% CI); $\downarrow/\uparrow$: fraction of images on which the answer moved down/up. Models that read their tags should shift by about $-1$ unless they repair.}
\label{tab:perturb}
\begin{tabular}{@{}llcrrrrrrr@{}}
\toprule
model & data & pert. & $n$ & follows image & follows own & repaired & acc.\ (\%) & shift & $\downarrow$/$\uparrow$ \\
\midrule
Qwen3-VL-8B & DotArrays & $-1$ & 55 & 100 {\tiny[100,100]} & 0 {\tiny[0,0]} & 45 {\tiny[33,58]} & 29 & +1.56 {\tiny[-3.37,6.62]} & 40/33 \\
 & DotArrays & $+1$ & 31 & 97 {\tiny[90,100]} & 3 {\tiny[0,10]} & 0 {\tiny[0,0]} & 10 & +8.52 {\tiny[1.24,16.55]} & 28/62 \\
 & FSC-147 & $-1$ & 54 & 98 {\tiny[94,100]} & 0 {\tiny[0,0]} & 15 {\tiny[6,24]} & 2 & +5.48 {\tiny[-3.40,15.98]} & 58/34 \\
 & FSC-147 & $+1$ & 40 & 100 {\tiny[100,100]} & 0 {\tiny[0,0]} & 0 {\tiny[0,0]} & 8 & +15.16 {\tiny[3.16,30.49]} & 27/70 \\
Gemini 3.1 Flash-Lite & DotArrays & $-1$ & 68 & 91 {\tiny[84,97]} & 1 {\tiny[0,4]} & 81 {\tiny[71,90]} & 57 & -1.75 {\tiny[-4.72,0.87]} & 25/22 \\
 & DotArrays & $+1$ & 35 & 94 {\tiny[86,100]} & 0 {\tiny[0,0]} & 40 {\tiny[23,57]} & 23 & +1.54 {\tiny[-7.69,14.31]} & 46/37 \\
 & FSC-147 & $-1$ & 68 & 82 {\tiny[72,91]} & 3 {\tiny[0,7]} & 34 {\tiny[24,46]} & 21 & -3.04 {\tiny[-7.25,0.25]} & 40/33 \\
 & FSC-147 & $+1$ & 52 & 87 {\tiny[77,94]} & 8 {\tiny[2,15]} & 17 {\tiny[8,29]} & 23 & -0.02 {\tiny[-7.47,7.98]} & 29/55 \\
GPT-5 mini & DotArrays & $-1$ & 43 & 84 {\tiny[72,93]} & 7 {\tiny[0,16]} & 47 {\tiny[33,60]} & 35 & -0.60 {\tiny[-1.65,0.28]} & 40/26 \\
 & DotArrays & $+1$ & 46 & 63 {\tiny[50,76]} & 17 {\tiny[7,28]} & 17 {\tiny[7,28]} & 15 & -0.46 {\tiny[-2.28,1.24]} & 33/39 \\
 & FSC-147 & $-1$ & 54 & 76 {\tiny[65,87]} & 15 {\tiny[6,24]} & 22 {\tiny[11,33]} & 15 & -1.42 {\tiny[-4.37,1.06]} & 40/33 \\
 & FSC-147 & $+1$ & 49 & 61 {\tiny[47,76]} & 24 {\tiny[12,37]} & 12 {\tiny[4,22]} & 16 & -0.02 {\tiny[-1.12,1.19]} & 35/38 \\
Gemini 3.7 Flash & DotArrays & $-1$ & 68 & 90 {\tiny[82,96]} & 0 {\tiny[0,0]} & 100 {\tiny[100,100]} & 90 & +0.19 {\tiny[0.01,0.41]} & 4/12 \\
 & DotArrays & $+1$ & 13 & 62 {\tiny[38,85]} & 15 {\tiny[0,38]} & 85 {\tiny[62,100]} & 85 & +0.62 {\tiny[0.00,1.46]} & 8/31 \\
 & FSC-147 & $-1$ & 73 & 79 {\tiny[70,89]} & 3 {\tiny[0,7]} & 70 {\tiny[59,79]} & 40 & +0.75 {\tiny[-1.01,3.21]} & 30/23 \\
 & FSC-147 & $+1$ & 46 & 72 {\tiny[59,85]} & 11 {\tiny[2,20]} & 52 {\tiny[37,65]} & 26 & +4.07 {\tiny[0.95,8.02]} & 23/51 \\
\bottomrule
\end{tabular}
\end{table}

\paragraph{What \arm{direct} measures.}
\label{app:reason}
\Cref{tab:reason} reports the hidden reasoning tokens that the provider returns for the single-turn arms, and the thinking-off control: the \arm{direct} prompt rerun with reasoning disabled (Gemini 3.1 Flash-Lite, GPT-5 mini) or set to the minimal effort the endpoint accepts (Gemini 3.7 Flash, whose endpoint, like Gemini 3.1 Pro's, refuses to disable reasoning). The paired change is computed on the same images; $R_{80}$ uses the DotArrays main battery (and, for Gemini 3.7 Flash, the XL extension). The Qwen models and, mostly, Claude Sonnet 5 produce no reasoning tokens, so for them \arm{direct} is estimation in the intended sense; for the Gemini and Grok models it is an answer-only arm whose hidden computation grows with $n$.
\begin{table}[h]
\centering\scriptsize
\setlength{\tabcolsep}{3pt}
\caption{What \arm{direct} measures. Left: hidden reasoning tokens reported by the provider for the single-turn arms at the ``low'' effort used throughout (median over all images; $\ge40$: median on DotArrays with $n\ge40$; any: fraction of \arm{direct} replies with at least one reasoning token). Right: the thinking-off control, the same \arm{direct} prompt with hidden reasoning disabled (Flash-Lite: \texttt{enabled=false}; GPT-5 mini: effort \texttt{minimal}, which yields zero reasoning tokens) or minimised where the endpoint refuses to disable it (Gemini 3.7 Flash: effort \texttt{minimal}); exact accuracy on the same images and the paired change (pp, 95\,\% CI), and $R_{80}$ on DotArrays.}
\label{tab:reason}
\begin{tabular}{@{}lrrrrlrrrrrr@{}}
\toprule
& \multicolumn{4}{c}{reasoning tokens (low effort)} & \multicolumn{7}{c}{thinking-off control} \\
\cmidrule(lr){2-5}\cmidrule(lr){6-12}
model & direct & $\ge40$ & any & CoT & setting & tokens & DotArrays & FSC-147 & PixMo & $\Delta$ (pooled) & $R_{80}$ \\
\midrule
Qwen3-VL-8B & 0 & 0 & 0\% & 0 & -- & -- & -- & -- & -- & -- & -- \\
Qwen3-VL-32B & 0 & 0 & 0\% & 0 & -- & -- & -- & -- & -- & -- & -- \\
Gemini 3.1 Flash-Lite & 122 & 120 & 100\% & 133 & off & 0 & 53$\to$60 & 20$\to$18 & 74$\to$72 & +1.8 {\tiny[-1.1,4.8]} & 7$\to$4 \\
GPT-5 mini & 128 & 352 & 97\% & 256 & \texttt{minimal} & 0 & 58$\to$38 & 12$\to$19 & 69$\to$58 & -9.4 {\tiny[-13.2,-5.4]} & 7$\to$2 \\
Gemini 3.7 Flash & 480 & 1720 & 100\% & 596 & \texttt{minimal} & 500 & 97$\to$99 & 48$\to$45 & 90$\to$88 & -0.8 {\tiny[-2.5,0.9]} & 150$\to$120 \\
Gemini 3.1 Pro & 482 & 1294 & 99\% & 505 & -- & -- & -- & -- & -- & -- & -- \\
Grok 4.20 & 1894 & 4288 & 100\% & 2519 & -- & -- & -- & -- & -- & -- & -- \\
Grok 4.5 & 0 & 1744 & 42\% & 1070 & -- & -- & -- & -- & -- & -- & -- \\
Claude Sonnet 5 & 0 & 248 & 10\% & 48 & -- & -- & -- & -- & -- & -- & -- \\
\bottomrule
\end{tabular}
\end{table}

\paragraph{Lattice controls and the jitter check.}
\label{app:lattice}
A near-regular square lattice invites a shortcut: count rows and columns and emit an evenly spaced grid, which would score as on-target without any dot being inspected. \Cref{tab:jitter} checks the existing arrays, whose dots are displaced from their cell centres by up to $\pm15\,\%$ of the cell: for the precise pointers the tag offsets track the dot offsets almost perfectly and tags lie several times closer to the dot than to the cell centre, so the tags are placed on dots. \Cref{tab:lattice} adds three lattice sets built with the main battery's generator, canvas, disc size, jitter, and seeds per level (\cref{app:data}): a square lattice rotated by 20--40$^\circ$, a hexagonal lattice, and a square lattice with about a quarter of its cells left empty, each at $n\in\{12,20,30,40,50\}$, compared with the main-battery regular and random arrays at the same $n$. Hexagonal and holed lattices keep Flash-Lite's advantage under the routine but not Qwen3-VL-8B's, whose regular advantage therefore looks lattice-specific, and the rotated lattice removes the advantage for every model and arm; for the model with a working routine the regular advantage is that of an organised scan path rather than of a recitable product. Gemini 3.7 Flash is at ceiling throughout and GPT-5 mini shows no arrangement effect under either arm.
\begin{table}[h]
\centering\scriptsize
\setlength{\tabcolsep}{3pt}
\caption{Arrangement effect on lattices that defeat a rows-times-columns shortcut. Exact accuracy (\%) on the main-battery near-regular and random arrays at $n\in\{12,20,30,40,50\}$ (20 images each) and on three new lattice sets with the same $n$, jitter, and seeds-per-level (20 images each): a square lattice rotated by 20--40$^\circ$, a hexagonal lattice, and a square lattice with about a quarter of its cells left empty. Right: the regular-minus-random and (rotated+hexagonal+holed)-minus-random differences with 95\,\% bootstrap CIs (pp).}
\label{tab:lattice}
\begin{tabular}{@{}llrrrrrrr@{}}
\toprule
model & arm & regular & random & rotated & hexagonal & holed & regular $-$ random & lattices $-$ random \\
\midrule
Qwen3-VL-8B & direct & 45 & 45 & 35 & 35 & 15 & +0 {\tiny[-30,30]} & -17 {\tiny[-42,8]} \\
 & tag & 60 & 25 & 5 & 25 & 25 & +35 {\tiny[5,65]} & -7 {\tiny[-28,13]} \\
Gemini 3.1 Flash-Lite & direct & 35 & 25 & 10 & 25 & 35 & +10 {\tiny[-20,35]} & -2 {\tiny[-25,20]} \\
 & tag & 85 & 55 & 30 & 85 & 90 & +30 {\tiny[5,55]} & +13 {\tiny[-12,38]} \\
GPT-5 mini & direct & 40 & 40 & 0 & 10 & 25 & +0 {\tiny[-30,30]} & -28 {\tiny[-50,-7]} \\
 & tag & 10 & 20 & 0 & 10 & 30 & -10 {\tiny[-30,10]} & -7 {\tiny[-27,12]} \\
Gemini 3.7 Flash & direct & 100 & 100 & 90 & 100 & 100 & +0 {\tiny[0,0]} & -3 {\tiny[-8,0]} \\
 & tag & 100 & 95 & 85 & 100 & 100 & +5 {\tiny[0,15]} & +0 {\tiny[-10,13]} \\
\bottomrule
\end{tabular}
\end{table}

\begin{table}[h]
\centering\scriptsize
\caption{Grid-guessing check on the main-battery near-regular arrays ($n\ge9$; each dot is displaced from its lattice cell centre by a jitter of up to $\pm15\,\%$ of the cell). For every on-target tag: Pearson correlation and regression slope between the dot's jitter and the tag's offset from the same cell centre (a model that emits an even grid has $r\approx0$); median distance from the tag to the dot versus to the cell centre (px); and the fraction of tags closer to the dot than to the centre.}
\label{tab:jitter}
\begin{tabular}{@{}llrrrrrrr@{}}
\toprule
model & arm & images & tags & $r$ & slope & to dot (px) & to centre (px) & closer to dot (\%) \\
\midrule
Qwen3-VL-8B & list & 40 & 919 & 0.63 & 0.88 & 15.1 & 24.2 & 67 \\
 & tag & 40 & 990 & 0.55 & 0.80 & 16.0 & 21.4 & 64 \\
Qwen3-VL-32B & list & 40 & 1084 & 0.96 & 0.95 & 3.8 & 13.6 & 89 \\
 & tag & 40 & 1075 & 0.85 & 0.90 & 5.2 & 14.1 & 81 \\
Gemini 3.1 Flash-Lite & list & 40 & 1082 & 0.98 & 0.98 & 1.7 & 13.7 & 95 \\
 & tag & 40 & 1070 & 0.86 & 0.89 & 3.3 & 14.1 & 80 \\
GPT-5 mini & list & 38 & 635 & 0.37 & 0.90 & 34.7 & 35.3 & 55 \\
 & tag & 33 & 493 & 0.35 & 0.82 & 40.0 & 40.6 & 56 \\
Gemini 3.7 Flash & list & 40 & 1061 & 0.97 & 1.00 & 0.7 & 14.1 & 99 \\
 & tag & 40 & 1084 & 0.99 & 1.00 & 0.9 & 13.8 & 99 \\
\bottomrule
\end{tabular}
\end{table}

\paragraph{Seeing the marks versus listing them.} \Cref{tab:taglist} gives the paired \arm{tag}$-$\arm{list} contrasts behind \cref{sec:res-tools}.
\begin{table}[h]
\centering\scriptsize
\caption{Seeing the marks versus listing them: paired change in exact accuracy of \arm{tag} over \arm{list} (blind pointing) on the common subset, percentage points with 95\,\% paired bootstrap CIs over images. Bold: the interval excludes zero.}
\label{tab:taglist}
\begin{tabular}{@{}lrrr@{}}
\toprule
model & DotArrays & FSC-147 & PixMo-Count \\
\midrule
Qwen3-VL-8B & \textbf{+12.0 {\tiny[2,22]}} & +1.0 {\tiny[-6,8]} & +4.4 {\tiny[-7,16]} \\
Qwen3-VL-32B & -8.0 {\tiny[-20,4]} & +5.0 {\tiny[-4,14]} & -2.2 {\tiny[-10,6]} \\
Gemini 3.1 Flash-Lite & -4.0 {\tiny[-16,8]} & \textbf{-9.0 {\tiny[-17,-1]}} & -1.1 {\tiny[-11,9]} \\
GPT-5 mini & \textbf{-22.0 {\tiny[-36,-10]}} & -2.0 {\tiny[-8,4]} & -3.3 {\tiny[-13,7]} \\
Gemini 3.7 Flash & +0.0 {\tiny[-8,8]} & -2.0 {\tiny[-10,6]} & -1.1 {\tiny[-6,3]} \\
\bottomrule
\end{tabular}
\end{table}

\section{Cost}
\label{app:cost}
Every API call logs the provider-reported cost (OpenRouter \texttt{usage.cost}), so the accounting below is measured, not estimated. The whole study, roughly 67{,}000 calls over 29{,}000 trajectories (E1 across nine models, the ablation grid, the mirror-flip, traversal-order, repeat-control, and strict-seriality experiments, and the XL extension, including pilots and reruns), cost \totalCostUSDexact{} US dollars. The three tier-1 models, which ran all seven arms on all \nImages{} images, account for about one third of the total; the frontier and contrast models on the 240-image common subset and the XL extension account for most of the rest. \Cref{tab:cost} lists, per model, the OpenRouter identifier, the list prices, the mean cost per image for each arm, and the total spend. Two observations: the tool loop multiplies cost by roughly $3\times$ over \arm{direct} for the cheap models (more calls, re-sent images) but remains below one US cent per image for every tier-1 model; and the cost--accuracy frontier of \cref{fig:precision} (right) is driven almost entirely by the model, not by the arm.
\begin{table}[h]
\centering\scriptsize
\setlength{\tabcolsep}{3pt}
\caption{Models (OpenRouter identifiers, list prices in USD per million input/output tokens on 2026-08-18) and mean cost per image in US cents for each arm; the last column is the total spend per model in USD.}
\label{tab:cost}
\begin{tabular}{@{}llrrrrrrrrr@{}}
\toprule
model & identifier & \$/M in/out & direct & CoT & list & tag & tag+crop & full & tiles & total \\
\midrule
Qwen3-VL-8B & \texttt{qwen/qwen3-vl-8b-instruct} & 0.12/0.45 & 0.01 & 0.16 & 0.16 & 0.19 & 0.34 & 0.15 & 0.12 & 7.34 \\
Qwen3-VL-32B & \texttt{qwen/qwen3-vl-32b-instruct} & 0.10/0.42 & 0.01 & 0.06 & 0.04 & 0.19 & 0.16 & 0.23 & 0.09 & 4.95 \\
Gemini 3.1 Flash-Lite & \texttt{google/gemini-3.1-flash-lite} & 0.25/1.50 & 0.06 & 0.13 & 0.13 & 0.35 & 0.37 & 0.40 & 0.27 & 11.13 \\
GPT-5 mini & \texttt{openai/gpt-5-mini} & 0.25/2.00 & 0.07 & 0.12 & 0.11 & 0.31 & -- & 0.41 & -- & 6.66 \\
Gemini 3.7 Flash & \texttt{google/gemini-3.7-flash} & 0.38/1.88 & 0.19 & 0.25 & 0.16 & 0.40 & -- & 0.42 & -- & 9.26 \\
Gemini 3.1 Pro & \texttt{google/gemini-3.1-pro-preview} & 2.00/12.00 & 1.08 & 1.24 & -- & 3.61 & -- & 3.33 & -- & 22.23 \\
Grok 4.20 & \texttt{x-ai/grok-4.20} & 1.25/2.50 & 0.64 & 0.91 & -- & 1.50 & -- & 2.47 & -- & 13.25 \\
Grok 4.5 & \texttt{x-ai/grok-4.5} & 2.00/6.00 & 0.54 & 1.52 & -- & 1.87 & -- & 1.47 & -- & 12.95 \\
Claude Sonnet 5 & \texttt{anthropic/claude-sonnet-5} & 2.00/10.00 & 0.23 & 0.39 & -- & 1.90 & -- & 2.15 & -- & 11.20 \\
\bottomrule
\end{tabular}
\end{table}

\section{Failure gallery}
\label{app:gallery}
\begin{figure}[h]\centering\includegraphics[width=\linewidth]{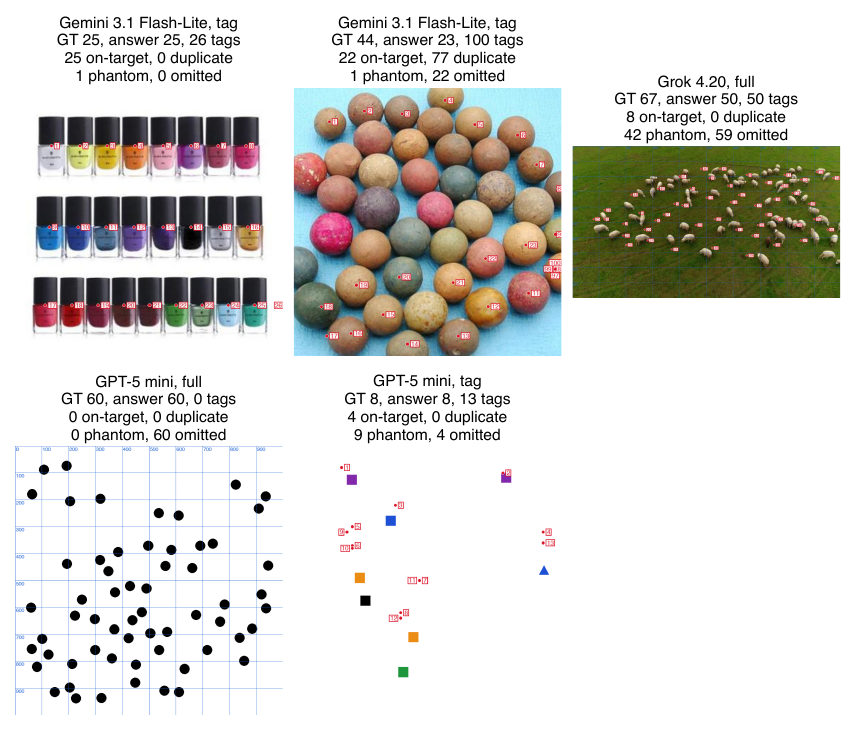}
\caption{Final working images from representative trajectories (re-rendered from stored actions), one per failure category: a case the tag loop fixed, a duplicate-heavy case, a phantom-heavy case, an omission-heavy case, and a cardinality-inconsistent case (announced count far from the number of tags). Titles give model, arm, ground truth, announced count and the audit.}\label{fig:gallery}\end{figure}

\Cref{fig:gallery} shows one representative trajectory for each of five automatically mined categories (final working images re-rendered from the stored actions; the selection script scans all \arm{tag}/\arm{full} trajectories with point ground truth). \emph{Fixed}: the tag loop corrects a wrong \arm{direct} answer. \emph{Duplicates}: several tags pile onto already-counted objects, typically in tight clusters. \emph{Phantoms}: tags land on background or between objects, the signature failure of imprecise pointers, which often follows a degenerate coordinate sequence (evenly spaced points marching across the image). \emph{Omissions}: whole regions are left untagged, the dominant failure in dense scenes. \emph{Cardinality}: the announced count differs from the number of drawn tags, i.e.\ the model tags, then distrusts its tags and estimates. The counts of mined candidates per category (\arm{tag} and \arm{full} over DotArrays and FSC-147) were 325, 81, 1815, 529, and 399 trajectories respectively, so phantoms are by far the most common tag-level pathology in this study.

\end{document}